\documentclass{tlp}

\usepackage{epsfig}
\usepackage{latexsym}

\newtheorem{definition}{Definition}

\newtheorem{theorem}{Theorem}
\newtheorem{lemma}{Lemma}
\newtheorem{corollary}{Corollary}
\newtheorem{example}{Example}

\newcommand{\nneg}[0]{\sim\!\!} 

\newcommand{\cl}[1]{\overline{#1}} 
\renewcommand{\cl}[1]{\neg #1} 
\newcommand{\DLPINH}[0]{\mathrm{DLP}^{<}} 
 
\newcommand{\sn}[1]{#1^\circ} 
\newcommand{\corr}[1]{\widehat{#1}} 
\renewcommand{\corr}[1]{\check{#1}} 
\newcommand{\commadots}[0]{,\ldots ,}

\newcommand{\useq}[1]{{\mathbf #1}}
\renewcommand{\useq}[1]{\mbox{\em\bfseries #1}}
\newcommand{\useqs}[1]{\mbox{\em\bfseries\scriptsize #1}}

\newcommand{\DLV}[0]{\texttt{DLV}}
\newcommand{\XSB}[0]{\texttt{XSB}}
\newcommand{\smodels}[0]{\texttt{smodels}}
\newcommand{\DeRes}[0]{\texttt{DeRes}}
\newcommand{\url}[1]{\texttt{#1}}

\newcommand{\trmlist}[1]{\mathbf{#1}}

\newcommand{\rejfo}[1]{{\mathit{rej}}_#1}
\newcommand{\rejf}[2]{\rejfo{#1}(#2)}
\newcommand{\rejo}{{\mathit{rej}}}
\newcommand{\rej}[1]{\rejo(#1)}
\newcommand{\rejecto}{{\mathit{reject}}}
\newcommand{\reject}[1]{\rejecto(#1)}
\newcommand{\Cn}[1]{\mathit{Cn}(#1)}
\newcommand{\Cni}[2]{\mathit{Cn}_{#1}(#2)}

\newcommand{\iec}[0]{i.e.,\ }
\newcommand{\egc}[0]{e.g.,\ }
\newcommand{\tsim}[0]{\hspace{1.1mm}\rule{.16mm}{2.3mm}\hspace{-.3mm}
\raisebox{.1ex}{$\sim$}\hspace{.8mm}}
\newcommand{\tsimi}[1]{\hspace{1.1mm}\rule{.16mm}{2.3mm}\hspace{-.3mm}
\raisebox{.1ex}{$\sim$}_{\useqs{#1}}\hspace{.8mm}}

\newcommand{\upd} {\lhd}

\newcommand{\head}[1]{H(#1)}

\newcommand{\body}[1]{B(#1)}
\newcommand{\bodyp}[1]{B^+(#1)}
\newcommand{\bodyn}[1]{B^-(#1)}
\newcommand{\KB}{\it KB}
\newcommand{\la}{\leftarrow}
\newcommand{\naf}{{\it not}\,}
\newcommand{\Q}{{\cal Q}}
\renewcommand{\P}{P^<}

\newcommand{\rulems}[1]{\rho(#1)}

\newcommand{\NP}{\mbox{${\rm N\!P}$}}
\renewcommand{\NP}{\mbox{\rm NP}}
\newcommand{\coNP}{\mbox{${\rm coN\!P}$}}
\newcommand{\NPNP}{\NP^{\rm NP}}
\renewcommand{\coNP}{\mbox{\rm coNP}}

\newcommand{\SigmaP}[1]{{\Sigma}_{#1}^{P}}
\newcommand{\PiP}[1]{{\Pi}_{#1}^{P}}
\newcommand{\at}{{\mathit At}}

\renewcommand{\at}{{\mathcal A}}

\newcommand{\extat}{\at_{ext}}
\newcommand{\Lit}{{\mathit Lit}}

\renewcommand{\extat}{{\mathit Lit}_{\at}}

\newcommand{\rs}{Rej}
\renewcommand{\rs}{\mathit{Rej}}

\newcommand{\bel}{\mathit{Bel}}
\newcommand{\lang}{{\cal L}}
\newcommand{\tuple}[1]{\langle#1\rangle}
\newcommand{\SM}{\mathit{SM}}
\newcommand{\SMup}{\mathcal{U}}
\renewcommand{\SM}{\mathcal{S}}
\newcommand{\AS}{\mathit{AS}}
\renewcommand{\AS}{\mathcal{S}}
\newcommand{\ASup}{\mathcal{U}}

\newcommand{\nop}[1]{}

\newcommand{\reductr}[1]{#1^{+}}

\newcommand{\prenot}[1]{{not#1}}

\newcounter{myenumctr}

\begin{document}

\title[
On Properties of Update Sequences]{On Properties of Update Sequences\\
Based on Causal Rejection
}

\author[T.\ Eiter et al.]
{THOMAS EITER, MICHAEL FINK, \and
GIULIANA SABBATINI, and HANS TOMPITS\\
           Technische Universit\"{a}t
           Wien, \\
Institut f{\"u}r Informationssysteme, \\
 Abt.\
           Wissensbasierte Systeme 184/3, \\
Favoritenstra{\ss}e 9-11,  A-1040 Vienna, Austria\\
           \email{[eiter,michael,giuliana,tompits]@kr.tuwien.ac.at}}

\maketitle

\begin{abstract} 
In this paper,
we
consider an approach to update 
nonmonotonic knowledge bases represented as extended logic programs under the answer set semantics. In this approach, new information is incorporated into the current knowledge base subject to a causal rejection principle, which enforces 
that, in case of conflicts between rules, more recent rules are preferred and older rules are overridden. Such a rejection principle is also exploited in other approaches to update logic programs, notably in the method of dynamic logic programming, due to Alferes \emph{et al.}

One of the central issues of this paper is a thorough analysis of
various properties of the current approach, in order to get a better
understanding of the inherent causal rejection principle. For this
purpose, we review postulates and principles for update and revision
operators which have been proposed in the area of theory change and
nonmonotonic reasoning.  Moreover, some new properties for approaches
to updating logic programs are considered as well. Like related update
approaches, the current semantics does not incorporate a notion of
\emph{minimality of change}, so we consider refinements of the
semantics in this direction.  As well, we investigate the relationship
of our approach to others in more detail. In particular, we show that
the current approach is semantically equivalent to inheritance
programs, which have been independently defined by Buccafurri \emph{et
al.}, and that it coincides with certain classes of dynamic logic
programs, for which we provide characterizations in terms of graph
conditions.  In view of this analysis, most of our results about
properties of the causal rejection principle apply to each of these
approaches as well. Finally, we also deal with computational issues. Besides a discussion on the computational complexity of our approach,
we 
outline how the update semantics and its refinements can be directly implemented on top of existing logic programming systems. In the present case, we implemented the update approach using the logic programming system \DLV.
\end{abstract}

\section{Introduction}
\label{sec:intro}

\subsection{Motivation and Context}

Logic programming has been conceived as a computational logic paradigm
for problem solving and offers a number of advantages over
conventional programming languages. In particular, it is a well-suited
tool for declarative knowledge representation and common-sense
reasoning \cite{bara-gelf-94}, and possesses thus a high potential as a key
technology to equip software agents with advanced reasoning
capabilities in order to make those agents behave intelligently (cf., \egc \cite{sadr-toni-99}).

It has been realized, however, that further work is needed on
extending the current methods and techniques to fully support the
needs of agents. In a simple (but, as for currently deployed agent
systems, realistic) setting, an agent's knowledge base, $\KB$, may be
modeled as a logic program, which the agent may evaluate to answer
queries that arise. Given various approaches to semantics, the problem of evaluating a logic program is
 quite well-understood, and
(beside Prolog) provers for semantics with more sophisticated
treatment of negation may be used. Currently available provers include the systems  \DeRes\ \cite{chol-etal-96}, \DLV\
\cite{eite-etal-97a}, \smodels\ \cite{niem-simo-96b}, and \XSB\ \cite{rao-etal-97}.  

An important aspect, however, is that an agent is situated in an
environment which is subject to change. This requests the agent to
adapt over time, and to adjust its decision making.  An agent might be
prompted to adjust its knowledge base $\KB$ after receiving new
information in terms of an {\em update} $U$, given by a clause or a
set of clauses that need to be incorporated into $\KB$.  Simply adding
the rules of $U$ to $\KB$ does not give a satisfactory solution in
practice, even in simple cases. For example, if $\KB$ contains the
rules $a\la b$ and $b\la \ $, and $U$ consists of the rule $\neg a \la
\ $ stating that $a$ is false, then the union $\KB \cup U$ is not
consistent under predominant semantics such as the answer set
semantics \cite{gelf-lifs-91} or the well-founded semantics
\cite{vang-etal-91}. However, by attributing higher priority to the
update $\neg a \la \ ~$, a result is intuitively expected which has a
consistent semantics, where the emerging conflict between old and new
information is resolved.

To address this problem, some approaches for updating
logic programs with (sets of) rules  
have been proposed  recently~\cite{alfe-etal-98,alfe-etal-99b,inou-saka-99,foo-zhan-98}. In this paper, we consider an approach
which is based on a {\em causal rejection principle}. According to
this principle, a rule $r$ is only discarded providing there is a ``reason''
for doing so, in terms of another, more recent rule $r'$ which
contradicts $r$. That is, if both $r$ and $r'$ are applicable (i.e.,
their bodies are satisfied) and have opposite heads, then only $r'$ is
applied while $r$ is discarded. In the example from above, the rule
$r:~a\la b$ in the current knowledge base $\KB$ (whose body is true given rule
$b\la \ $) is rejected by the new rule $r':~\neg a\la \ $ in the update
(whose body is also true), and thus in the updated knowledge base, $r$
is not applied.

The causal rejection principle is not novel---in fact, it constitutes a major
ingredient of the well-known dynamic logic programming approach
\cite{alfe-etal-98,alfe-etal-99a}. Furthermore, it underlies, in
slightly different forms, the related approaches of inheritance logic
programs \cite{bucc-etal-99a-iclp} and ordered logic
programs \cite{LaeSacVer90,bucc-etal-96}. We provide here a simple and
rigorous realization of this principle, in terms of ``founded''
rejection: a rule $r$ may only be rejected by some other rule $r'$
{\em which itself is not rejected}. While this foundedness condition,
as it appears, plays in effect no role in the particular semantics we
consider, it can do so for more involved semantics based on causal
rejection, such as the one by Alferes \emph{et al.}~\shortcite{alfe-etal-98,alfe-etal-99a}.  

Starting from a simple formalization 
of a
semantics for updating logic programs based on causal rejection, which offers the advantage of a clear declarative characterization and of a syntactical realization at the same time, the main goal of this paper is to
investigate properties of this semantics, as well as to analyze the relationship to other semantics for updating logic programs,
in particular to 
dynamic logic
programming. Notice that, although uses and extensions of dynamic logic
programming have been discussed
(cf.~\cite{alfe-etal-99b,alfe-pere-00,leit-etal-00}), its properties and
relations to other approaches and related formalisms have been
less explored so far (but see~\cite{alfe-pere-00}).

\subsection{Main Contributions}

Inspired by ideas in~\cite{alfe-etal-98,alfe-etal-99a}, we consider a
semantics for sequences $\useq{P}=(P_1,\ldots,P_n)$ of extended logic
programs, in terms of a syntactic transformation to an {\em update
program}, which is a single extended logic program in an extended
language. The semantics properly generalizes the answer set
semantics~\cite{gelf-lifs-91} of single logic programs. The readable
syntactic representation of the semantical results---which is useful
from a computational perspective---is complemented, as
in~\cite{alfe-etal-98,alfe-etal-99b}, by an elegant semantical
characterizations in terms of a modified Gelfond-Lifschitz reduction,
resulting from the usual construction by removal of rejected
rules. The transformation we describe is similar to the one by Alferes
\emph{et al.}, but involves only a few types of rules and new
atoms. For capturing the rejection principle, information about rule
rejection is explicitly represented at the object level through
rejection atoms; this is similar to an implementation of the related
inheritance logic program approach proposed by Buccafurri \emph{et
al.}~\shortcite{bucc-etal-99a-iclp}. Though not new in spirit, the
approach we suggest offers a more accessible definition and is
suitable for studying general properties of updates by causal
rejection, providing insight in the mechanism of the rejection
principle itself.

The main contributions of this paper can be summarized as follows.

(1) We extensively investigate, from different points of view,
properties of update programs and answer set semantics for update
sequences. We first analyze them from a belief revision perspective,
and evaluate various (sets of) postulates for revision and iterated
revision from the literature
\cite{alch-etal-85,kats-mend-91,darw-pear-97,lehm-95}. To this end, we
discuss possible interpretations of update programs as change operators
for nonmonotonic logical theories.  As it turns out, update programs
(and thus equivalent approaches) do not satisfy many of the properties
defined in the literature. This is partly explained by the
nonmonotonicity of logic programs and the causal rejection principle
embodied in the semantics, which strongly depends on the syntax of
rules.

Furthermore, we consider properties from a nonmonotonic reasoning
perspective, by naturally interpreting update programs as nonmonotonic
consequence relations, and review postulates and principles which have
been analyzed by Kraus, Lehmann, and Magidor~\shortcite{KrausLehmannMagidor90}, and Makinson~\shortcite{maki-93}.

Finally, we present and discuss some further general properties
relevant for update programs. Among them is an \emph{iterativity
property}, which informally states equivalence of nesting
$((P_1,P_2),P_3)$ and sequences $(P_1,P_2,P_3)$ of updates. A possible
interpretation of this property is that an \emph{immediate update
strategy}, which incorporates new information immediately into the
knowledge base, is equivalent to \emph{demand-driven evaluation},
where the actual knowledge base $\KB$ is built on demand of particular
queries, and full information about $\KB$'s update history is
known. As we shall see, the property does not hold in general, but for
certain classes of programs.

(2) As it appears, update answer sets---like related concepts 
based on causal rejection---do not respect minimality of
change. We thus refine the semantics of update sequences and introduce
\emph{minimal answer sets} and \emph {strictly minimal answer sets}. Informally, in
minimal answer sets, the set of rules that need to be rejected is
minimized.  This means that a largest set of rules should be respected
if an answer set is built; in particular, if all rules can be
satisfied, then no answer sets would be adopted, which request the
rejection of any rule. The notion of strict minimality further refines
minimality by enforcing that rejection of older rules should be preferred to rejection of newer rules,
thus performing hierarchic minimization. 

The refined semantics come at the cost of higher computational
complexity, and increase the complexity of update answer sets for
propositional programs by one level, namely  from the first to the second level
in the polynomial hierarchy. This parallels similar results for the
update semantics by Sakama and Inoue~\shortcite{inou-saka-99}, which
employs a notion of minimality in the basic definition.
 
(3) We conduct a comparison between update programs and alternative
approaches for updating logic programs
\cite{alfe-etal-98,alfe-etal-99a,foo-zhan-98,inou-saka-99,leit-pere-98,leite-97,mare-trus-94}
and related approaches \cite{bucc-etal-99a-iclp,delg-etal-00}. We find
that for some of these formalisms, syntactic subclasses are semantically
equivalent to update programs. Thus, update programs provide a
(different) characterization of these fragments, and by their
simplicity, contribute to better understanding on the
essential working of these formalisms on these fragments.
Furthermore, our results on properties of update answer set semantics
carry over to the equivalent fragments, and establish also semantical
results for these formalisms, which have not been analyzed much in
this respect so far. Finally, equivalent fragments of different
formalisms are identified via update programs.

First, we show that update programs are, on the language we consider,
equivalent to inheritance logic programs.  More precisely, our notion
of an answer set for an update sequence $\useq{P}=(P_1,\ldots,P_n)$
coincides with the notion of an answer set for a corresponding
inheritance program $\P$ in the approach by Buccafurri \emph{et
al.}~\shortcite{bucc-etal-99a-iclp}, where $\P$ results from
$\useq{P}$ by interpreting more recent updates in the sequence
$(P_1\commadots P_n)$ (\iec programs with higher index) as programs
containing more specific information.  Thus, update programs (and
classes of dynamic logic programs) may semantically be regarded as
fragment of the inheritance framework of Buccafurri \emph{et
al.}~\shortcite{bucc-etal-99a-iclp}.  We then compare our update
programs to revision programming by Marek and
Truszczy{\'n}ski~\shortcite{mare-trus-94} and the related approach of
Leite and Pereira~\shortcite{leit-pere-98}, which has been extended to
sequences of programs in \cite{leite-97}. It appears that the fragment
of this formalism where programs merely use weak negation is, apart
from extra conditions on sequences of more than two programs,
semantically equivalent to update programs. Furthermore, we give a
thorough analysis of the dynamic logic programming approach by Alferes
\emph{et al.}~\shortcite{alfe-etal-98,alfe-etal-99a}.  Their notion of
model of an update sequence $\useq{P}$, 
which we refer to as {\em
dynamic answer set}, semantically imposes extra conditions compared to
our update answer set. We give a precise characterization of the case
in which the definitions are equivalent, using graph-theoretic
concepts. From this characterization, we get syntactic conditions
for classes of programs on which dynamic answer sets and update answer
sets coincide. Notice that the examples discussed in
\cite{alfe-etal-98,alfe-etal-99a} satisfy these conditions.%
\footnote{In a preliminary version of this paper \cite{eite-etal-00f},
we erroneously reported, due to misunderstanding notation in
\cite{alfe-etal-98,alfe-etal-99a}, that dynamic logic programs and
update programs are equivalent in general. This view was supported by
the examples mentioned and many others we considered.}  Furthermore,
by this correspondence, some results for update principles and
computational complexity derived for our update programs carry over to
dynamic logic programs as well. Further inspection, which we do not carry
out here, suggests the same results beyond the corresponding
fragments.

To the best of our knowledge, no investigation of approaches to
updating logic programs from the perspectives of belief revision and
nonmonotonic consequences relations has been carried out so far.
In view of our results about the relationship between update
programs and other approaches, in particular to inheritance logic
programs and fragments of dynamic logic programming, our investigations
apply to these formalisms as well.

\subsection{Structure of the Paper}

The  paper is organized as follows. After providing some
necessary preliminaries in the next section, we introduce in
Section~\ref{sec:update-progs}  update programs and
answer sets for such programs, and establish some characterization
results.
In
Section~\ref{sec:properties}, we embark on our study of
general principles of update programs based on causal rejection from various perspectives. The
refinements of answer sets to minimal and strictly minimal answer sets
are considered in Section~\ref{sec:refinements}.
Section~\ref{sec:computation} is devoted to computational issues of
our approach. After an investigation of the computational complexity
of update programs under the semantics introduced, we discuss an
implementation of our approach based on the \DLV\ logic programming
tool~\cite{eite-etal-97a,eite-etal-98a}.  In
Section~\ref{sec:rel-work}, relations to other and related approaches
are investigated. The paper concludes with
Section~\ref{sec:conclusion}, containing a short summary and a
discussion of further work and open issues. Some proofs and further
results, which are omitted here for space reasons, can be found in
\cite{eite-etal-00g}.

\section{Preliminaries}
\label{sec:prelim}

We deal with extended logic programs~\cite{gelf-lifs-91}, which  consist of rules built over a 
set $\at$ of
propositional atoms where both default negation $\naf$ and strong negation $\neg$ is available. A \emph{literal}, $L$, is either an atom $A$ (a \emph{positive 
literal}) or a strongly negated atom $\neg  A$ (a \emph{negative literal}).
For a literal $L$, the \emph{complementary literal}, $\cl{L}$, is $\neg A$ 
if $L=A$, and $A$ if $L=\neg A$, for some atom $A$. 
For a set $S$ of literals, we define $\cl{S}=\{ \cl{L} \mid L \in
S\}$, and  denote by $\extat$ the set $\at \cup \cl{\at}$ of all literals 
over $\at$. 
A literal preceded $\naf$ is called a \emph{weakly negated literal}.

A \emph{rule}, $r$, is an ordered pair \( L_0 \la \body{r}, \) where
$L_0$ is a literal and $\body{r}$ is a finite set of literals or
weakly negated literals. 
We also allow the
case where $L_0$ may be absent. We call $L_0$ the \emph{head} of $r$,
denoted  $\head{r}$, and $\body{r}$ the \emph{body} of $r$.
 For $\body{r}=\{L_1\commadots L_m,\naf
L_{m+1}\commadots \naf L_n\}$, we define \( \bodyp{r} =
\{L_1\commadots L_m\} \) and \( \bodyn{r} = \{L_{m+1}\commadots L_n\}
\).  The elements of $\bodyp{r}$ are referred to as the {\em
prerequisites\/} of $r$.  We employ the usual conventions for writing
rules like $L_0\la B_1\cup B_2$ or $L_0\la B_1\cup \{L\}$ as $L_0\la
B_1,B_2$ and $L_0\la B_1,L$, respectively. Generally, rule $r$ with
$\body{r}$ as above will simply be written as
\[
L_0 \la L_1,\dots,L_m,\naf L_{m+1},\dots,\naf L_n.
\] 
If $r$ has an empty head, then $r$ is a \emph{constraint}; 
if the body of $r$ is empty,
then $r$ is a \emph{fact}; if $n=m$ (\iec if $r$ contains no default negation), then $r$ is a  {\em basic rule\/}. We denote by $\lang_\at$ the 
set of all rules constructible using the 
literals in $\extat$.

An \emph{extended logic program} (ELP), $P$, 
is a (possibly infinite) set   
of rules. If all rules in $P$ are basic, then $P$ is a \emph{basic program}.
Usually, $\at$ will simply be understood as the set of \emph{all} atoms occurring in $P$.

An \emph{interpretation} $I$ is a set of literals which is {\em
consistent\/}, \iec $I$ does not contain complementary literals $A$ and
$\neg A$.  A literal $L$ is \emph{true} in 
$I$ (symbolically $I\models L$) iff $L\in I$, and
\emph{false} otherwise. Given a rule $r$, the body $\body{r}$ of $r$
is true in $I$, denote $I\models \body{r}$, iff (i) each $L\in\bodyp{r}$ is true in $I$ and (ii)
each $L\in\bodyn{p}$ is false in $I$.
Rule $r$ is true in $I$, denoted $I\models r$, iff $\head{r}$ is true in $I$ whenever $\body{r}$ is true in $I$. 
In particular, a constraint $r$ is true in $I$ iff $I\not\models\body{r}$.
For a program $P$,  $I$ is a \emph{model} of $P$, denoted $I\models
P$, if $I\models r$ for all $r\in P$.

Let $r$ be a rule.
Then $\reductr{r}$ denotes the basic rule obtained from $r$ by deleting all
weakly negated literals in the body of $r$, \iec
$\reductr{r}=\head{r}\la\bodyp{r}$.
Furthermore, we say that rule $r$ is {\em defeated\/} by a set of literals $S$ if some literal in $\bodyn{r}$ is true in $S$, \iec if 
$\bodyn{r}\cap S\neq\emptyset$.
As well, each literal in $\bodyn{r}\cap S$ is said to {\em defeat\/} $r$.

The {\em reduct}, $P^S$, of a program $P$ {\em relative to\/} a
set $S$ of literals is defined by
\[
P^S
=
\{\reductr{r} \mid \mbox{$r\in \Pi$ and $r$ is not defeated by $S$}\}.
\]
An interpretation $I$ 
is an \emph{answer set} of a program $P$
iff it is a minimal model of $P^I$.
By $\AS(P)$ we denote the collection of all answer sets of $P$.
If $\SM(P)\neq\emptyset$, then $P$ is said to be  \emph{satisfiable}.

We regard a logic program $P$ 
as the {\em epistemic state\/} of an
agent. The given semantics is used for assigning a \emph{belief set} 
to any epistemic state $P$ as follows.

Let $I\subseteq \extat$ be an interpretation. Define 
\[
\bel_{\at}(I)=\{ r \in \lang_\at \mid I \models r\}.
\]
Furthermore, 
for a class ${\cal I}$ of interpretations, let $\bel_\at({\cal I}) = \bigcap_{I\in
{\cal I}} \bel_\at(I)$.


\begin{definition}
\label{defn:belief-set}
For a logic program $P$, the belief set, $\bel_\at(P)$, of $P$ is
given by $\bel_\at (P) = \bel_\at(\SM(P))$.
\end{definition}
We write $P\models_\at r$ if $r \in \bel_\at(P)$, and for any program $Q$, we write
$P\models_\at Q$ if $P\models_\at q$ for all $q\in Q$. 
Programs $P_1$ and $P_2$ are \emph{equivalent} (modulo 
$\at$), symbolically $P_1 \equiv_\at P_2$, iff $\bel_\at
(P_1)=\bel_\at (P_2)$. It can be seen that if either $P_1$ or $P_2$
involves only finitely many atoms, or if $\at$ is finite, then $P_1
\equiv_\at P_2$ is equivalent to the condition that $P_1$ and $P_2$ have
the same answer sets modulo $\at$.  We will drop the subscript
``$\at\,$'' in $\bel_\at (\cdot)$, $\models_\at$, and $\equiv_\at$ if
no ambiguity can arise.

Belief sets enjoy the following natural properties:

\begin{theorem}  For every logic program $P$, we have that:
\begin{enumerate}
        \item[(i)] $P \subseteq \bel(P)$;
        \item[(ii)] $\bel (\bel (P))= \bel (P)$;
        \item[(iii)] $\{r\mid \mbox{$I\models r$, for every interpretation $I$}\}\subseteq \bel (P)$.
\end{enumerate}
\end{theorem}

\begin{proof}
Properties (i) and (iii) hold trivially. 
Property (ii) can be seen as follows: $\bel(P)\subseteq\bel(\bel(P))$ follows directly from property (i), and $\bel(\bel(P))\subseteq \bel(P)$ holds due to the fact that each answer set of $P$ is also an answer set of $\bel(P)$.
\end{proof}

Clearly, the belief operator $\bel(\cdot)$ is nonmonotonic, i.e., in general, $P_1 \subseteq P_2$
does not imply $\bel (P_1) \subseteq \bel (P_2)$.

\section{Update programs}
\label{sec:update-progs}

We introduce a framework to update logic programs based on a
compilation technique to ELPs. The basic idea is
the following. Given a sequence $(P_1\commadots P_n)$ of ELPs, each $P_i$ is assumed to update the information
expressed by the initial section $(P_1,\ldots,P_{i-1})$. The sequence
$(P_1\commadots P_n)$ is translated into a single ELP $P'$,
respecting the successive update information, such that the answer
sets of $P'$ represent the answer sets of $(P_1\commadots P_n)$. The
translation is realized by introducing new atoms $\rej{\cdot}$ which
control the applicability of rules with respect to the update
information\footnote{This idea can be found elsewhere in the literature, \egc~\cite{kowa-toni-96,inou-00}}. Informally, $\rej{r}$ states that rule $r$ is
``rejected'', in case a more recent rule $r'$ asserts a
conflicting information. This conflict is resolved by enabling
$\rej{r}$ to block the applicability of $r$, and so rule $r'$ is given
precedence over $r$.

In some sense, the proposed update mechanism can be seen as some form
of an \emph{inheritance strategy}, where more recent rules are viewed
as ``more specific'' information, which have to be given preference in
case of a conflict. In Section~\ref{sec:inheritance}, we will discuss
the relationship between our update formalism and the inheritance
framework introduced by Buccafurri \emph{et
al.}~\shortcite{bucc-etal-99a-iclp}.

The general method of expressing update sequences in terms of single
programs has already been discussed by Alferes \emph{et
al.}~\shortcite{alfe-etal-98,alfe-etal-99a}. However, in that
framework, applicability issues are realized in terms of newly
introduced atoms referring to the derivability of \emph{atoms} of the
original programs, and not to the applicability of \emph{rules} as in
the present approach. A detailed comparison between our approach and
the method of Alferes \emph{et
al.}~\shortcite{alfe-etal-98,alfe-etal-99a} is given in
Section~\ref{sec:relations-dynLP}.

\subsection{Basic Approach}

By an \emph{update sequence}, $\useq{P}$, we understand a series
$(P_1,\ldots,P_n)$ of ELPs. 
We say that $\useq{P}$ is an update sequence \emph{over $\at$} iff $\at$ 
represents the set of atoms occurring in the rules of the constituting 
elements $P_i$ of $\useq{P}$ $(1\leq i\leq n)$.

Given an update sequence $\useq{P}=(P_1,\ldots,P_n)$  over $\at$,   
we assume a 
set $\at^\ast$ extending $\at$ by new, pairwise distinct atoms 
$\rej{r}$ and $A_i$, for each $r$ occurring in $\useq{P}$, each atom $A\in\at$, and each $i$, $1\leq i\leq n$.
We further assume an injective {\em naming function} $N(\cdot,\cdot)$,
which assigns to each rule $r$ in a program $P_i$ a distinguished name,
$N(r,P_i)$, obeying the condition $N(r,P_i) \neq N(r',P_j)$ whenever $i\neq j$. 
With a slight abuse of notation we shall 
identify $r$ with $N(r,P_i)$ as usual. 
Finally, for a literal $L$, we write $L_i$ to denote the result of replacing the atomic formula $A$ of $L$ by $A_i$.

\begin{definition}
\label{def:update-program}
Given an update sequence 
$\useq{P}=(P_1, \dots, P_n)$ over a set of atoms $\at$, we define 
the update program $\useq{P}_{\upd}=P_1 \upd \dots \upd P_n$ over $\at^\ast$ consisting 
of the following items:

\begin{enumerate}
\item[(i)] 
all constraints in $P_i$, $1 \leq i \leq n$;

\item[(ii)] 
for each $r \in P_i$, $1 \leq i \leq n$:
\begin{eqnarray*}
L_i & \la & \body{r}, \naf \rej{r} \qquad \quad \mbox{if $\head{r}=L$;} 
\end{eqnarray*}

\item[(iii)] 
for each $r \in P_i$, $1 \leq i < n$:
\begin{eqnarray*}
\rej{r} & \la & \body{r}, \neg L_{i+1} \qquad  \quad \mbox{if $\head{r}=L$;}
\end{eqnarray*}

\item[(iv)] 
for each literal $L$ occurring in $\useq{P}$ $(1\leq i <n)$:
$$L_i \la L_{i+1}; \qquad  
\qquad 
L \la L_1.
$$
\end{enumerate}
\end{definition}

Informally, this program expresses layered derivability of a literal $L$, beginning at the top layer $P_n$ downwards to the bottom
layer $P_1$.  The rule $r$ at layer $P_i$ is only applicable if it is
not refuted by a literal derived at a higher level that is incompatible with $H(r)$. 
Inertia rules propagate a locally
derived value for $L$ downwards to the first level, where the local
value is made global.
The transformation $\useq{P}_\upd$ 
is modular in the
sense that  
for $\useq{P}\,'=(P_1,\ldots,P_n,P_{n+1})$ it augments 
$\useq{P}_\upd=P_1 \upd \dots \upd P_n$ only with rules depending on $n+1$.

We remark that $\useq{P}_\upd$ can obviously be slightly simplified, which is 
relevant
for implementing our approach. 
All weakly negated literals $\naf \rej{r}$ in rules with
heads $L_n$ can be removed: Indeed, since $\rej{r}$ cannot be
derived, each such atom evaluates to false in any answer set of $\useq{P}_\upd$. Thus,
no rule from $P_n$ is rejected in an answer set of $\useq{P}_\upd$, i.e.,
all most recent rules are obeyed.

The intended answer sets of an update sequence $\useq{P}=(P_1,\ldots,P_n)$ are defined
 in terms of the answer sets of $\useq{P}_\upd$.

\begin{definition} 
\label{defn:upd-stable-model}
Let $\useq{P}=(P_1, \dots, P_n)$ be an update sequence over a set of atoms $\at$. 
Then, $S
\subseteq \extat$ is an \emph{update answer set} of $\useq{P}$ iff 
$S = S' \cap \at$ for some answer set $S'$ of $\useq{P}_{\upd}$. The collection of all update answer sets of 
$\useq{P}$ is denoted by $\ASup(\useq{P})$. 
\end{definition}

Following the case of single programs, an update sequence $\useq{P}=(P_1,\ldots, P_n)$ 
is regarded as the epistemic state of an agent, and the belief set $\bel(\useq{P})$ 
is given by $\bel(\ASup(\useq{P}))$. The update sequence $\useq{P}$ is said to be satisfiable 
iff $\ASup(\useq{P})\neq\emptyset$, and $\useq{P} \equiv \useq{P}'$ iff $\bel (\useq{P})=\bel (\useq{P}')$ ($\useq{P}'$ some update sequence). General properties of the belief operator $\bel(\cdot)$ in the context of update sequences will be discussed in Section~\ref{sec:properties}. 

For illustration of Definition~\ref{defn:upd-stable-model}, consider the 
following example, adapted from \cite{alfe-etal-98}.

\begin{example}
\label{example:tv}
 Consider the update of $P_1$ by $P_2$, where
$$
\begin{array}{rll@{}l}
P_1 & = & \big\{ & \ r_1: \ \emph{sleep} \la \naf \emph{tv\_on}, \quad r_2: \ \emph{night} \la \ , \quad r_3: \ \emph{tv\_on} \la \ , \\
&&& \ r_4: \ \emph{watch\_tv} \la \emph{tv\_on} \ \big\};\\[1ex]
P_2 & = & \big\{ & \ r_5: \ \neg \emph{tv\_on} \la \emph{power\_failure}, \quad r_6: \ \emph{power\_failure} \la \ \big\}.
\end{array}
$$
The single answer set of $\useq{P}=(P_1, P_2)$ is, as desired, 
\begin{eqnarray*}
S&  = & \{ \emph{power\_failure}, \neg \emph{tv\_on}, \emph{sleep}, \emph{night}\},
\end{eqnarray*}
since the only answer set of $\useq{P}_{\upd}$ is given by
$$
\begin{array}{rll@{}l}
S' & = & \big\{ & \ \emph{power\_failure}_2, \emph{power\_failure}_1, \emph{power\_failure}, \\
&&& \ \neg \emph{tv\_on}_2, \neg \emph{tv\_on}_1, \neg \emph{tv\_on},  \rej{r_3}, \emph{sleep}_1, \emph{sleep},
\emph{night}_1, \emph{night} \ \big\}.
\end{array}
$$

If new information arrives in form of the program $P_3$:
\begin{eqnarray*}
P_3 & = & \big\{ \ r_7: \ \neg \emph{power\_failure} \la \ \big\},
\end{eqnarray*}
then the update sequence $(P_1,P_2,P_3)$ has the answer set 
\begin{eqnarray*}
T & = & \big\{ \ \neg \emph{power\_failure}, \emph{tv\_on}, \emph{watch\_tv}, \emph{night} \ \big\},
\end{eqnarray*}
generated by the following answer set $T'$ of $P_1 \upd P_2 \upd P_3$:
$$
\begin{array}{rll@{}l}
T' & = & \big\{ & \ \neg \emph{power\_failure}_3, \neg \emph{power\_failure}_2, \neg \emph{power\_failure}_1, \neg \emph{power\_failure},\\
 &&& \ \rej{r_6}, \emph{tv\_on}_1, \emph{tv\_on}, \emph{watch\_tv}_1, \emph{watch\_tv}, \emph{night}_1, \emph{night} \ \big\}.
\end{array}
$$
\end{example}

\subsection{Properties and Characterizations}\label{sec:props:char}

Next, we discuss some properties of our approach.
The first result guarantees that answer sets of $\useq{P}$ are uniquely determined 
by the answer sets of $\useq{P}_\upd$.

\begin{theorem} \label{prop:1-1-corr}
Let $\useq{P}=(P_1,\dots,P_n)$ be an update sequence over a set of atoms $\at$, 
and let $S,T\subseteq\Lit_{\at^\ast}$ be answer sets of $\useq{P}_\upd$. Then, $S\cap \extat = T \cap 
\extat$ only if $S=T$.
\end{theorem}

\begin{proof} See Appendix~\ref{app:proof-prop:1-1-corr}. 
\end{proof}

In view of this result, the following notation is well-defined.

\begin{definition}
Let $\useq{P}$ be an update sequence over $\at$, and let $S$ be an answer set of $\useq{P}$. 
Then, $\corr{S}$ denotes the (uniquely determined) answer set of $\useq{P}_\upd$ 
obeying $S=\corr{S}\cap \extat$.
\end{definition} 

If an update sequence $\useq{P}$ consists of a single program $P_1$, the  
update answer sets of $\useq{P}$ coincide with the regular answer sets of $P_1$.

\begin{theorem}
Let $\useq{P}$ be an update sequence consisting of a single program $P_1$, \iec $\useq{P}=P_1$. 
Then, $\ASup(\useq{P})=\AS(P_1)$.
\end{theorem} 

\begin{proof}
This follows at once from the observation that 
the only difference between $P_1$ and $\useq{P}_\upd$ is that each rule $r=L\la \body{r}$ occurring in $P_1$ is replaced 
by the two rules $L_1\la \body{r},\naf \rej{r}$ and $L\la L_1$. Since there are no rules in $\useq{P}_\upd$ having head literal $\rej{r}$,  it holds that, for each set $S$ of literals, $r$ is defeated by $S$ exactly if $L_1\la \body{r},\naf \rej{r}$ is defeated by $S$.
\end{proof}

Answer sets of update sequences can also be characterized in a purely 
declarative way. To this end, we introduce the concept of a \emph{rejection set}.
Let us call two rules $r_1$ and $r_2$  \emph{conflicting} iff $\head{r_1}=\neg\head{r_2}$.  For an update sequence $\useq{P} = (P_1,\dots,P_n)$ over a set of atoms $\at$ and 
$S\subseteq \extat$, based on the principle of founded rule rejection, we define the rejection
set of $S$ by $\rs(S,\useq{P}) =
\bigcup_{i=1}^n \rs_i(S,\useq{P})$, where $\rs_n(S,\useq{P})=\emptyset$, and, for $n> i \geq 1$,
$$
\begin{array}{rcl@{~}l}
\rs_i(S,\useq{P}) \!\!&=& \!\! \big\{ \ r \in P_i \mid & \exists r' \in
P_j\setminus \rs_j(S,\useq{P}), \textrm{ for some } j\in \{i+1,\ldots, n\},\\
            &&& \textrm{such that $r,  r'$ are conflicting and $S\models\body{r}\cup\body{r'}$} \ \big\}.
\end{array}
$$
That is, $\rs(S,\useq{P})$ contains those rules from $\useq{P}$ which are
rejected on the basis of rules which are not rejected themselves.

The next lemma ensures that the rejection set $\rs(S,\useq{P})$ precisely matches the intended meaning of the control atoms $\rej{\cdot}$. 

\begin{lemma}\label{lemma:rejection-atoms}
Let $\useq{P}=(P_1\commadots P_n)$ be an update sequence over a set of atoms $\at$, let $S$ be an answer set of $\useq{P}$, and let $\corr{S}$ be the corresponding answer set of $\useq{P}_\upd$. Then, $r\in\rs(S,\useq{P})$ iff $\rej{r}\in\corr{S}$.
\end{lemma}

\begin{proof}
We show by induction on $j$ ($0\leq j< n$) that  $r\in\rs_{n-j}(S,\useq{P})$ iff $\rej{r}\in\corr{S}$, whenever $r\in P_{n-j}$.

\medskip
\noindent
{\sc Induction Base.} 
 Assume $j=0$. Then the statement holds trivially because $\rs_n(S,\useq{P})=\emptyset$ and $\rej{r}\notin\corr{S}$ for all $r\in P_n$.

\medskip
\noindent
{\sc Induction Step.} Assume $n>j>0$, and let the statement hold 
for all $k<j$. We show the assertion for $k=j$. Consider some $r\in P_{n-j}$ and suppose $r\in\rs_{n-j}(S,\useq{P})$. We show $\rej{r}\in\corr{S}$. According to the definition of $\rs_{n-j}(S,\useq{P})$, there is some $r'\in P_{n-k}\setminus \rs_{n-k}(S,\useq{P})$, $0\leq k < j$, such that $\head{r'}=\neg\head{r}$, $\bodyp{r}\cup\bodyp{r'}\subseteq S$, and both $r$ and $r'$ are not defeated by $S$. 
The rule $r\in P_{n-j}$ induces the rule $\rej{r}\la\body{r},\neg L_{n-j+1}\in\useq{P}_\upd$, where $L=\head{r}$. From the properties above, we have $\rej{r}\la\bodyp{r},\neg L_{n-j+1}\in(\useq{P}_\upd)^{\corr{S}}$. Now, since $\bodyp{r}\subseteq S\subseteq\corr{S}$, in order to show $\rej{r}\in\corr{S}$ it suffices to show that $\neg L_{n-j+1}\in\corr{S}$. This can be seen as follows. First of all, the rule $r'\in P_{n-k}$ induces the rule $L'_{n-k}\la\body{r},\naf \rej{r'}\in\useq{P}_\upd$, where $L'=\head{r'}$.
Since $\head{r'}=\neg\head{r}$, we actually have  $\neg L_{n-k}\la\body{r},\naf \rej{r'}\in\useq{P}_\upd$.
Now, given that $r'\notin \rs_{n-k}(S,\useq{P})$, and since $k<j$, by induction hypothesis we have $\rej{r}\notin \corr{S}$.
Furthermore, $\bodyn{r'}\cap S=\emptyset$ implies
$\neg L_{n-k}\la\bodyp{r'}\in(\useq{P}_\upd)^{\corr{S}}$. 
Given that $\bodyp{r'}\subseteq S\subseteq\corr{S}$, we obtain $\neg L_{n-k}\in \corr{S}$. By observing that $n-j+1\leq n-k$ (since $k<j$), and given the inertia rules $\neg L_m\la\neg L_{m+1}\in(\useq{P}_\upd)^{\corr{S}}$ ($1\leq m<n$), we eventually obtain $L_{n-j+1}\in\corr{S}$. This proves $\rej{r}\in\corr{S}$.

Conversely, assume $\rej{r}\in\corr{S}$. We show $r\in\rs_{n-j}(S,\useq{P})$. By construction of the update program $\useq{P}_\upd$, the atom $\rej{r}$ can only be derived by means of the rule $\rej{r}\la\body{r},\neg L_{n-j+1}\in \useq{P}_\upd$. So, it must hold that $\bodyp{r}\subseteq S$, $\neg L_{n-j+1}\in \corr{S}$, and $\bodyn{r}\cap S=\emptyset$. Moreover, since $\neg L_{n-j+1}\in \corr{S}$, there must be some $r'\in P_{n-k}$, $k<j$, such that $\neg L_{n-k}\la\body{r'},\naf\rej{r'}\in\useq{P}_\upd$, $\bodyp{r'}\subseteq S$, $\bodyn{r'}\cap S=\emptyset$, and $\rej{r'}\notin \corr{S}$. By induction hypothesis, the latter fact implies $r'\notin\rs_{n-k}(S,\useq{P})$. 
So, we have that there is some $r'\in P_{n-k}\setminus\rs_{n-k}(S,\useq{P})$, $k<j$,  such that  $\head{r'}=\neg \head{r}$, $\bodyp{r}\cup\bodyp{r'}\subseteq S$, and both $r$ and $r'$ are not defeated by $S$. This means that $r\in \rs_{n-j}(S,\useq{P})$.
\end{proof}

It turns out that update answer sets can be characterized in terms of a modified Gelfond-Lifschitz reduction, by taking the elements of the respective rejection sets into account. In what follows, for a given update sequence  $\useq{P}=(P_1\commadots P_n)$, we write $\cup\useq{P}$ to denote the set of all rules occurring in $\useq{P}$, \iec $\cup\useq{P}=\bigcup_{i=1}^n P_i$.

\begin{theorem}
\label {theo:sequence-char-0}
Let $\useq{P}= (P_1,\dots,P_n)$ be an update sequence over a set of atoms
$\at$ and $S \subseteq \extat$ a set of literals. Then, $S$ is an answer set of
$\useq{P}$ iff $S$ is the  minimal model of $(\cup \useq{P}\setminus \rs(S,\useq{P}))^S$.
\end{theorem}

\begin{proof}
See Appendix~\ref{app:proof-theo:sequence-char-0}.
\end{proof}

Update answer sets can also be described using a weaker notion of rejection sets. For $\useq{P}=(P_1\commadots P_n)$ over $\at$ and $S\subseteq\extat$, let us define 
$$
\begin{array}{rcl@{~}l}
\rs'(S,\useq{P}) \!\! &=&  \!\! \bigcup_{i=1}^n \{ \ r \in P_i \mid & \exists r' \in
P_j, \textrm{ for some } j\in \{i+1,\ldots, n\}, \textrm{ such that  }\\
            &&& \textrm{$r$ and $r'$ are conflicting and } S\models 
\body{r}\cup\body{r'} \ \}.
\end{array}
$$
Obviously, $\rs(S,\useq{P})\subseteq\rs'(S,\useq{P})$ always holds. Moreover, we get the following variant of Theorem~\ref{theo:sequence-char-0}: 

\begin{theorem}
\label{theo:sequence-char-0-alt}
Let $\useq{P}= (P_1,\dots,P_n)$ be an update sequence over a set of atoms
$\at$ and $S \subseteq \extat$ a set of literals. Then, $S$ is an answer set of
$\useq{P}$ iff $S$ is a minimal model of $(\cup \useq{P}\setminus \rs'(S,\useq{P}))^S$.
\end{theorem}

\begin{proof}
\emph{Only-if part.} Suppose $S$ is an answer set of $\cup \useq{P}\setminus \rs(S,\useq{P})$. Since $\rs(S,\useq{P})\subseteq\rs'(S,\useq{P})$, it holds that $(\cup \useq{P}\setminus \rs'(S,\useq{P}))\subseteq (\cup \useq{P}\setminus \rs(S,\useq{P}))$. 
Observe that for each $r\in \rs'(S,\useq{P})\setminus\rs(S,\useq{P})$ there is some  $r'\in \rs(S,\useq{P})$ such that $\head{r}=\head{r'}$ and $S\models \body{r}\cup\body{r'}$.
We use the following property: 
Let $S'$ be an answer set of some program $P$, and let $r,r'\in P$ such that $\head{r}=\head{r'}$ and $S\models \body{r}\cup\body{r'}$. Then, $S'$ is an answer set of $P\setminus\{r'\}$.
By repeated applications of this property we get that $S$ is an answer set of $\cup \useq{P}\setminus \rs'(S,\useq{P})$.

\smallskip\noindent
\emph{If part.} Suppose that $S$ is a minimal model of $(\cup \useq{P}\setminus \rs'(S,\useq{P}))^S$, but there is some $r\in (\cup \useq{P}\setminus \rs(S,\useq{P}))\setminus (\cup \useq{P}\setminus \rs'(S,\useq{P}))$ such that $S\models\body{r}$ and $\head{r}\notin S$. It follows that $r\in \rs'(S,\useq{P})\setminus \rs(S,\useq{P})$.

Define $Q_i=\{r'\in P_i\mid \head{r'}\in\{\head{r},
\neg \head{r}\} \textrm{ and } S\models \body{r'}\}$ and let $r'\in Q_k$, where $k=\max\{i\mid r'\in Q_i\}\neq 0$. Then, $r'\notin\rs'(S,\useq{P})$. Since $\head{r}\notin S$, it follows that $\head{r'}=\neg\head{r}$. Furthermore,  $r'\notin\rs'(S,\useq{P})$ implies $r'\notin\rs(S,\useq{P})$. Therefore, it follows that $r\notin\rs(S,\useq{P})$, a contradiction. We obtain that $S$ is a model of $(\cup \useq{P}\setminus \rs(S,\useq{P}))^S$. Moreover, since $(\cup \useq{P}\setminus \rs'(S,\useq{P}))^S\subseteq (\cup \useq{P}\setminus \rs(S,\useq{P}))^S$, $S$ must be a minimal model of $(\cup \useq{P}\setminus \rs(S,\useq{P}))^S$.
 \end{proof}

We make use of this alternative version of $\rs(\cdot,\cdot)$ in Sections~\ref{sec:relations-RevProg} and \ref{sec:relations-dynLP}.

It is important to emphasize that in our approach, the update program
$P_{\upd}$ is not the {\em result} of the update intended to be the
new knowledge state of the agent, but it {\em represents the semantic
result} of the information that a sequence of updates $P_2,\ldots,P_n$
has occurred to a knowledge base $P_1$. Compiling the result of
updates into a single logic program in the original language (having
the desired answer sets) would mean losing history information about
the update sequence.  Instead, the formalism results in a program over
an extended set of atoms, which expresses at the object level
meta-concepts determining applicability of rules and computation of
those intended answer sets. In some sense, the result is therefore a
declarative specification of how rules of the original logic program
and of subsequent updates should be applied, expressed in the language
of logic programs themselves.

\section{Principles of Program Updates}
\label{sec:properties}

In this section, we discuss several kinds of postulates which have been advocated in the literature on belief change and examine to what extent update sequences satisfy these principles. This issue has not been addressed extensively in previous work. We first consider update programs from the
perspective of \emph{belief revision} and assess the relevant postulates from this area. Afterwards, we briefly analyze further properties, like viewing update programs as \emph{nonmonotonic consequence operators} and other general principles. We remark that our analysis applies, in slightly adapted form, to dynamic logic programming as well (cf.\ Section~\ref{sec:relations-dynLP}).

\subsection{Belief Revision}

Following G\"{a}rdenfors and Rott~\shortcite{gard-rott-95}, two different approaches to belief revision 
can be distinguished: (i) {\em immediate revision}, where the new information is simply added to the current stock of beliefs and the belief change is accomplished 
by the semantics of the underlying (often, nonmonotonic) logic; and (ii) {\em logic-constrained revision}, where the new stock of beliefs is determined by a nontrivial operation which adds and retracts
beliefs, respecting logical inference and some constraints.

In the latter approach, it is assumed that beliefs are sentences from
a given logical language $\cal L_B$, 
closed under the standard boolean connectives. A {\em belief set}, $K$, is a subset of $\cal L_B$ which is
closed under a consequence operator $\Cn{\cdot}$ of the underlying logic. A
{\em belief base} for $K$ is a subset $B \subseteq K$ such that
$K=\Cn{B}$. A belief base is a special case of \emph{epistemic state}
\cite{darw-pear-97}, which is a set of sentences $E$ representing an
associated belief set $K$ in terms of a mapping $\bel(\cdot)$ such that $K=\bel(E)$,  where $E$ need not necessarily have the same language as $K$. 

In what follows, we first introduce different classes of postulates, and then we examine them with respect to update sequences.

\subsubsection{AGM Postulates} 

One of the main aims of logic-constrained revision is to characterize
suitable revision operators through postulates. In the AGM approach (after Alchourr\'on, G\"{a}rdenfors, and Makinson \shortcite{alch-etal-85}), three basic operations on a belief set $K$ are considered: 

\begin{itemize}
        \item {\em expansion} $K+\phi$, which is simply adding the new information $\phi\in{\cal L}_B$ to $K$; 
        \item {\em revision} $K\star \phi$, which is sensibly revising $K$
in the light of $\phi$ (in particular, when $K$ contradicts $\phi$); and 
        \item {\em contraction} $K-\phi$, which is removing $\phi$ from $K$. 
\end{itemize}

AGM proposes a set of postulates, K$\star$1--K$\star$8, that any revision operator $\star$ mapping a belief set $K \subseteq {\cal L}_B$ and a
sentence $\phi \in {\cal L}_B$ into the revised belief set $K\star \phi$ should
satisfy. If, following both Darwiche and Pearl~\shortcite{darw-pear-97} and Brewka~\shortcite{brew-00}, we assume that $K$ is
represented by an epistemic state $E$, then the postulates K$\star$1--K$\star$8 can be reformulated as follows:

\begin{description}
        \item[(K1)] $E\star \phi$ represents a belief set.
        \item[(K2)] $\phi \in \bel(E\star \phi)$.
        \item[(K3)] $\bel(E \star \phi) \subseteq \bel(E+\phi)$.
        \item[(K4)] $\neg \phi \notin \bel(E)$ implies $\bel(E+\phi) \subseteq \bel(E\star\phi)$.
        \item[(K5)] $\bot \in \bel(E\star \phi)$ only if $\phi$ is unsatisfiable.
        \item[(K6)] $\phi_1 \equiv \phi_2$ implies $\bel(E\star\phi_1) = \bel(E \star \phi_2)$.
        \item[(K7)] $\bel(E \star(\phi\land\psi)) \subseteq \bel((E \star \phi)+\psi)$.
        \item[(K8)] $\neg\psi \notin \bel(E\star \phi)$ implies $\bel((E\star\phi)+ \psi) \subseteq \bel(E\star(\phi\land \psi))$.
\end{description}

Here, $E\star\phi$ and $E+\phi$ is the revision and expansion
operation, respectively, applied to $E$.  Informally, these postulates
express that the new information should be reflected after the
revision, and that the belief set should change as little as
possible. As has been pointed out, this set of postulates is appropriate
for new information about an \emph{unchanged world}, but not for
incorporation of a change to the actual world. Such a mechanism is addressed by the next set of postulates, expressing \emph{update} operations.

\subsubsection{Update Postulates}

For update operators $B\diamond \phi$ realizing a change $\phi$ to a belief base
$B$, Katsuno and Mendelzon~\shortcite{kats-mend-91} proposed a set of postulates, U$\diamond$1--U$\diamond$8, where both $\phi$ and $B$ are propositional sentences over a finitary language. For epistemic states $E$, these postulates can be reformulated as follows.

\begin{description}
        \item[(U1)] $\phi \in \bel(E \diamond \phi)$.
        \item[(U2)] $\phi \in\bel(E)$ implies $\bel(E\diamond \phi) = \bel(E)$.
        \item[(U3)] If $\bel(E)$ is consistent and $\phi$ is satisfiable,
then $\bel(E \diamond \phi)$ is consistent. 
        \item[(U4)] If $\bel(E) = \bel(E')$ and $\phi \equiv \psi$, then
$\bel(E\diamond \phi) = \bel(E\diamond \psi)$. 
        \item[(U5)] $\bel(E\diamond (\phi \wedge \psi)) \subseteq \bel((E
\diamond \phi)+ \psi)$.
        \item[(U6)] If $\phi \in \bel(E\diamond \psi)$ and $\psi \in
\bel(E\diamond \phi)$, then $\bel(E\diamond \phi) = \bel(E\diamond
\psi)$. 
        \item[(U7)] If $\bel(E)$ is complete, then $\bel(E\diamond (\psi \vee \psi')) \subseteq \bel(E\diamond \psi) \wedge \bel(E \diamond \psi'))$.\footnote{A belief  set K is \emph{complete} iff, for each atom $A$, either $A \in K$ or $\neg A \in K$.}
        \item[(U8)] $\bel((E \lor E') \diamond \psi) = \bel((E\diamond \psi) \lor (E'\diamond \psi)$.
\end{description}

Here, conjunction and disjunction of epistemic states are presumed to be definable  in the given language (e.g., in terms of intersection and union of associated sets of models, respectively).

The most important differences between (K1)--(K8) and (U1)--(U8) are that
revision should yield the same
result as expansion $E + \phi$, providing $\phi$ is compatible with $E$, which is not desirable for update in
general, cf.\ \cite{wins-88}. On the other hand, (U8) says that if $E$
can be decomposed into a disjunction of states (e.g., models), then
each case can be updated separately and the overall result is formed by
taking the disjunction of the emerging states.

\subsubsection{Iterated Revision}

Darwiche and Pearl~\shortcite{darw-pear-97} have proposed postulates for iterated revision, which can be rephrased in our setting as follows (we omit parentheses in sequences $(E\star\phi_1)\star\phi_2$  of revisions):

\begin{description}
        \item[(C1)] If $\psi_2 \in \bel(\psi_1)$, then $\bel (E \star \psi_2 \star \psi_1) = \bel(E\star \psi_1)$. 
        \item[(C2)] If $\neg\psi_2 \in \bel(\psi_1)$, then $\bel(E \star
\psi_1\star \psi_2) = \bel(E \star \psi_2)$. 
        \item[(C3)] If $\psi_2 \in \bel(E \star\psi_1)$, then $\psi_2 \in \bel(E\star \psi_2\star \psi_1)$. 
        \item[(C4)] If $\neg\psi_2 \notin\bel(E\star\psi_1)$, then
$\neg \psi_2 \notin \bel(E\star\psi_2\star\psi_1)$.
        \item[(C5)] If $\neg \psi_2 \in \bel(E\star\psi_1)$ and
$\psi_1\notin \bel(E\star\psi_2)$, then $\psi_1 \notin \bel(E\star\psi_1\star \psi_2)$. 
        \item[(C6)] If $\neg\psi_2\in \bel(E\star\psi_1)$ and
$\neg \psi_1 \in \bel(E\star\psi_2)$, then $\neg\psi_1\in \bel(E\star\psi_1\star
\psi_2)$. 
\end{description}

Another set of postulates for iterated revision, corresponding to a
sequence $E$ of observations, has been formulated by Lehmann~\shortcite{lehm-95}. Here, each observation is a sentence which is assumed
to be consistent (i.e., falsity is not observed), and the epistemic
state $E$ has an associated belief set $\bel(E)$. Lehmann's postulates
read as follows, where $E,E'$ denote sequences of observations and ``,'' stands for concatenation:

\begin{description}
        \item[(I1)] $\bel(E)$ is a consistent belief set. 
        \item[(I2)] $\phi \in \bel(E,\phi)$. 
        \item[(I3)] If $\psi \in \bel(E,\phi)$, then $\phi \Rightarrow \psi \in \bel(E)$.
        \item[(I4)] If $\phi \in\bel(E)$, then $\bel(E,\phi,E') = \bel(E,E)$.
        \item[(I5)] If $\psi \vdash \phi$ then $\bel(E,\phi,\psi,E') = \bel(E,\psi,E')$.
        \item[(I6)] If $\neg\psi\notin \bel(E,\phi)$, then $\bel(E,\phi,\psi,E') = \bel(E,\phi,\psi,E')$.
        \item[(I7)] $\bel(E,\neg\phi,\phi) \subseteq Cn(E+\phi)$.
\end{description}

\subsubsection{Analysis of the Postulates} 

In order to evaluate the different postulates, we need to adapt them for the
setting of update programs. Naturally, the epistemic state $\useq{P}=(P_1,\ldots,P_n)$ of an agent is subject to revision. However, the associated belief set $\bel(\useq{P})$ $(\subseteq \lang_\at)$ does not belong to a logical language closed under boolean connectives. Closing $\lang_\at$ under conjunction does not cause much troubles, as the identification of finite logic programs with finite conjunctions of clauses permits that updates of a logic program $P$ by a program $P'$ can be viewed as the update of $P$ with a single sentence from the underlying belief language. 
Ambiguities arise, however, with the interpretation of expansion, as well 
as with the meaning of negation and disjunction of rules and programs, respectively.
 
Depending on whether the particular structure of the epistemic
state $E$ should be respected, different definitions of expansion are
imaginable in our framework. At the ``extensional'' level of sentences,
represented by a program or sequence of programs $\useq{P}$, $\bel(\useq{P}+P')$ is
defined as $\bel(\bel(\useq{P})\cup P')$. At the ``intensional'' level of
sequences $\useq{P}=(P_1,\ldots,P_n)$, $\bel(\useq{P}+P')$ could be defined as
$\bel(P_1,\ldots,P_n\cup P')$. An intermediate approach would be
defining $\bel(\useq{P}+P') = \bel_\at(\useq{P}_\upd\cup P')$. We adopt the
extensional view here. Note that, in general, adding $P'$ to $\bel(\useq{P})$
does not amount to the semantical intersection of $P'$ and $\bel(\useq{P})$
(nor of $\cup\useq{P}$ and $P'$, respectively).

As for negation, we might interpret the condition $\neg\phi\notin \bel(E)$
(or $\neg\psi \notin \bel(E\star \phi$) in (K4) and (K8)) as satisfiability requirement for $E+\phi$ (or $(E\star\phi) + \psi$, respectively).
 
Disjunction $\vee$ of rules or programs (as epistemic states)
appears to be meaningful only at the semantical level. The union
$\AS(P_1)\cup \AS(P_2)$ of the answer sets of programs $P_1$
and $P_2$ may be represented syntactically through a program $P_3$,
which in general requests an extended set of atoms. We thus do not
consider the postulates involving the operator $\lor$.

Given these considerations, Table~\ref{table:agm} summarizes our interpretation 
of postulates (K1)--(K8) and (U1)--(U6), and includes references
whether the 
respective property holds or fails. We assume that $\useq{P},\useq{P}'$ are  
sequences of ELPs, and $P,P'$ denote single ELPs. Moreover, the notation $(\useq{P},P)$ is an abbreviation for the sequence $(P_1\commadots P_n,P)$ if $\useq{P}=(P_1\commadots P_n)$.  
Demonstrations and counterexamples concerning these properties are given in Appendix~\ref{app:AGM-update}, and can be easily adapted for dynamic logic programming too.


\begin{table}[t!]
\caption{Interpretation of Postulates $($K1\/$)$--$($K8\/$)$ and $($U1\/$)$--$($U6\/$)$.}\label{table:agm}
\begin{minipage}{\textwidth}
\begin{tabular}{clc}
\hline\hline
Postulate & \multicolumn{1}{c}{Interpretation} & Postulate holds \\
\hline
\bf{(K1)} & $(\useq{P},P)$ represents a belief set & yes \\[2ex]

\bf{(K2)}, \bf{(U1)} & $P \subseteq \bel((\useq{P},P))$ & yes \\[2ex]

\bf{(U2)} & $\bel (P) \subseteq \bel (\useq{P})$ implies $\bel ((\useq{P},P)) =
\bel (\useq{P})$  & no \\[2ex]

\bf{(K3)} & $\bel ((\useq{P},P)) \subseteq \bel(\bel(\useq{P}) \cup P)$ &
yes\footnote{If either $\useq{P}$ or $P$ has a finite alphabet.} \\[2ex]

\bf{(U3)} & If $\useq{P}$ and $P$ are satisfiable, then $(\useq{P},P)$ is satisfiable & no \\[2ex]

\bf{(K4)} & If $\bel(\useq{P})\cup P$ has an answer set, then & no \\
& $\bel(\bel(\useq{P})\cup P)\subseteq \bel ((\useq{P},P))$  \\[2ex]

\bf{(K5)} & $(\useq{P},P)$ is unsatisfiable only if $P$ is
unsatisfiable & no \\[2ex]

\bf{(K6)}, \bf{(U4)} & $\useq{P}\equiv \useq{P}'$ and $P\equiv P'$ implies
$(\useq{P},P) \equiv (\useq{P}',P')$ & no \\[2ex]

\bf{(K7)}, \bf{(U5)} & $\bel ((\useq{P},P\cup P')) \subseteq \bel(\bel((\useq{P},P))
\cup P')$ & yes\footnote{If either $(\cup\useq{P})\cup P$ or $P'$ has a finite alphabet.} \\[2ex]

\bf{(U6)} & $\bel (P')\subseteq \bel ((\useq{P},P))$ and  $\bel (P)
\subseteq \bel ((\useq{P},P'))$
 & no \\
& implies $\bel ((\useq{P},P))= \bel ((\useq{P},P'))$  \\[2ex]

\bf{(K8)} & If $\bel((\useq{P},P)) \cup P'$ is satisfiable, then & no \\
& $\bel(\bel((\useq{P},P))\cup P') \subseteq \bel ((\useq{P},P\cup P'))$ \\
\hline\hline
\end{tabular}
\vspace{-2\baselineskip}
\end{minipage}
\vspace{-1.5\baselineskip}
\end{table}


As can be seen from Table~\ref{table:agm}, apart from very simple postulates, the majority of the adapted AGM and update postulates are violated by update programs. This holds even for the case where $\useq{P}$ is a single program. In particular,
$\bel((\useq{P},P))$ violates discriminating postulates such as (U2) for
update and (K4) for revision. In the light of this, update programs
neither have update nor revision flavor.

We remark that the picture does not change if we abandon extensional
expansion and consider the postulates under intensional
expansion. Thus, also under this view, update programs do not satisfy
minimality of change.

The postulates (C1)--(C6) and (I1)--(I7) for iterated revision are treated in 
Table~\ref{table:iter}; proofs of these properties can be found in 
Appendix~\ref{app:iterated-revision}. Observe that Lehmann's postulate (I3) is considered as the pendant to AGM postulate K$\star$3. In a literal
interpretation of (I3), since the belief language associated
with logic programs does not have implication,  we may consider the case where $\psi$ is
a default literal $L_0$ and $\phi = L_1\land\cdots\land L_k$ is a
conjunction of literals $L_i$, such that $\phi \Rightarrow \psi$
corresponds to the rule $L_0 \la L_1,\ldots,L_k$. 
Moreover, since the negation of logic programs is not defined, we do not interpret (I7).


\begin{table}[t!]
\caption{Interpretation of Postulates $($C1\/$)$--$($C6\/$)$ and $($I1\/$)$--$($I6\/$)$.}\label{table:iter}
\begin{tabular}{clc}
\hline\hline
Postulate & \multicolumn{1}{c}{Interpretation} & Postulate holds \\
\hline

\bf{(C1)} & If $P' \subseteq \bel(P)$, then $\bel((\useq{P},P',P)) =
\bel((\useq{P},P))$ & no \\[2ex]

\bf{(C2)} & If $S \not\models P'$, for all $S \in \AS(P)$, then & no \\
& $\bel((\useq{P},P,P')) = \bel((\useq{P},P'))$  \\[2ex]

\bf{(C3)} & If $P' \subseteq \bel((\useq{P},P))$, then $P' \subseteq
\bel((\useq{P},P',P))$  & no \\[2ex]

\bf{(C4)} & If $S \models P'$ for some $S \in \ASup((\useq{P},P))$, then & yes \\
& $S\models P'$ for some $S \in \ASup((\useq{P},P',P))$  \\[2ex]

\bf{(C5)} & If $S\not\models P'$ for all $S\in \ASup((\useq{P},P))$ and $P
\not\subseteq \bel((\useq{P},P'))$,  & no \\
& then $P \not\subseteq
\bel((\useq{P},P,P'))$  \\[2ex]

\bf{(C6)} & If $S \not\models P'$ for all $S \in \ASup((\useq{P},P))$ and
$S\not\models P$ for all & no \\
& $S \in \ASup((\useq{P},P'))$, then $S\not\models
P$ for all $S \in \ASup((\useq{P},P,P'))$  \\[2ex]

\bf{(I1)} & $\bel(\useq{P})$ is a consistent belief set & no \\[2ex]

\bf{(I2)} & $P \subseteq \bel((\useq{P},P))$ & yes \\[2ex]

\bf{(I3)} &  If $L_0 \la \;\in \bel((\useq{P}, \{ L_1\la \ ,\ldots, L_k \la \ \}))$, then & yes \\
& $L_0 \la
L_1,\ldots,L_k \in \bel(\useq{P})$  \\[2ex]

\bf{(I4)} & If $Q_1 \subseteq \bel(\useq{P})$, then & no\\
& $\bel((\useq{P},Q_1,Q_2,\ldots,Q_n)) = \bel((\useq{P},Q_2,\ldots,Q_n))$  \\[2ex]

\bf{(I5)} & If $\bel(Q_2)\subseteq \bel(Q_1)$, then & no \\
& $\bel((\useq{P},Q_1,Q_2,Q_3,\ldots,Q_n))$=$\bel((\useq{P},Q_2,Q_3,\ldots,Q_n))$  \\[2ex]

\bf{(I6)} & If $S \models Q_2$ for some $S \in \ASup((\useq{P},Q_1))$, then & no \\
& \multicolumn{2}{l}{$\bel((\useq{P},Q_1,Q_2,\ldots,Q_n))= \bel((\useq{P},Q_1,Q_1\cup
Q_2,Q_3,$$\ldots,Q_n))$}  \\
\hline\hline
\end{tabular}
\vspace{-2\baselineskip}
\end{table}


Note that, although postulate (C3) fails in general, it holds if $P'$ 
contains a single rule. Thus, all of the above postulates except (C4) fail, and, with the exception of (C3), each change is given by a single rule.

We can view the epistemic state $\useq{P}=(P_1,\ldots,P_n)$ of an agent as a
prioritized belief base in the spirit of \cite{brew-91a,nebe-91,benf-etal-93}.
Revision with a new piece of information $Q$ is accomplished by simply
changing the epistemic state to $\useq{P}'=(P_1,\ldots,P_n,Q)$. The change of
the belief base is then automatically accomplished by the nonmonotonic
semantics of a sequence of logic programs. Under this view, updating
logic programs amounts to an instance of the immediate revision
approach.

On the other hand, referring to the update program, we may view the
belief set of the agent represented through a pair $\tuple{P,\at}$ of
a logic program $P$ and a (fixed) set of atoms $\at$, such that its
belief set is given by $\bel_\at(P)$. Under this view, a new piece of
information $Q$ is incorporated into the belief set by producing a
representation, $\tuple{P',\at}$, of the new belief set, where $P'=P\upd
Q$. Here, (a set of) sentences from an extended belief language is used
to characterize the new belief set, which is constructed by a
non-trivial operation employing the semantics of logic programs. Thus,
update programs enjoy to some extent also a logic-constrained revision
flavor. Nonetheless, as also the failure of postulates shows, they are
more an instance of \emph{immediate} than \emph{logic-constrained} revision.  
What we naturally expect, though, is that the two views described above
amount to the same \emph{at a technical level}. However, as we shall 
demonstrate below, this is not true in general.

\subsection{Update Programs as Nonmonotonic Consequence Relations}

Following G\"{a}rdenfors and Makinson~\shortcite{makinson-gardenfors:1990a,gardenfors-makinson:1994a}, belief revision can be related 
to nonmonotonic reasoning by interpreting it as an abstract consequence relation on
sentences, where the epistemic state is fixed.  In the same way, we can
interpret update programs as abstract consequence relation 
on
programs as follows. For a fixed epistemic state $\useq{P}$ and logic programs $P_1$ and
$P_2$, we define
$$P_1 \tsimi{P} P_2 \textrm{ if and only if } P_2\subseteq \bel(\useq{P},P_1),$$
i.e., if the rules $P_2$ are in the belief set of the agent after update of the epistemic state with $P_1$. 

Various postulates for nonmonotonic inference operations have been identified in the literature. In what follows, we consider some sets of postulates and discuss their interpretations in terms of update programs. First of all, we review principles discussed by Makinson~\shortcite{maki-93}, who considered a set of (desirable) properties for nonmonotonic reasoning, and analyzed the behavior of some reasoning formalisms with respect to these properties. Afterwards, we consider postulates proposed by Lehmann and Magidor~\shortcite{lehm-magi-92}, which deal with properties of so-called \emph{preferential consequence relations}. It is  argued that such properties are necessary but not sufficient for a preferential consequence relation to be meaningful and useful in reasoning. 
As we will see, updates fail also in satisfying the essential properties. 

Although our analysis is based on the specific semantics expressed by the transformation of Definition~\ref{def:update-program}, arguably it holds for other update formalisms as well. In fact, quite the same pattern can be found for dynamic logic programs \cite{alfe-etal-99a}, because, with few exceptions, all proofs and counterexamples hold for this formalism too (dynamic logic programming will be discussed in detail in Section~\ref{sec:relations-dynLP}).
 Thus,  intuitively, the failure of some basic principles of nonmonotonic reasoning in the context of updates stems  from the same nature of update semantics based on rule rejection, and not on the particular transformation chosen.

\subsubsection{General Patterns of Nonmonotonic Inference Relations}

Gabbay~\shortcite{gabbay:1985a} was the first to propose the idea that the output of nonmonotonic systems should be considered as an abstract consequence relation, in order to get a clearer understanding of the diverse nonmonotonic reasoning formalisms. Ensuing research identified several important principles, based on both syntactic and model-theoretic considerations. 
Among the different properties analyzed by Makinson~\shortcite{maki-93}, the following principles are amenable for logic programs under the standard Gelfond-Lifschitz approach, and can thus be formulated for update programs as well:
\begin{tabbing}
{\bf (N1)} \ \= $P_1 \tsimi{P} P_1$. \\[1ex]
{\bf (N2)} \>  If $P_1 \tsimi{P} Q_1 \wedge \dots \wedge Q_m$ and $P_1\wedge 
Q_1 \wedge \dots \wedge Q_m \tsimi{P} P_2$, then $P_1 \tsimi{P} P_2$. \\[1ex]
{\bf (N3)} \> If $P_1 \tsimi{P} Q_1 \wedge \dots \wedge Q_m$  and $P_1 
\tsimi{P} P_2$, then $P_1\wedge Q_1 \wedge \dots \wedge Q_m \tsimi{P} P_2$.  \\[1ex]
{\bf (N4)} \> If $P_1 \tsimi{P} P_2,\; P_2\tsimi{P} P_3,\ldots,\; P_n \tsimi{P}  P_1 \ (n\geq 2)$, then $\{ P' \mid P_i \tsimi{P} P' \}  = $\\[1ex]
\> $\{ P' \mid P_j \tsimi{P} P' \}$, for all  $i,j\leq n$.
  \end{tabbing}
Postulate (N1) is called \emph{Inclusion} and coincides with (K2) and (U1).
Properties (N2) and (N3) are important nonmonotonic inference principles and are respectively called \emph{Cut} and \emph{Cautious Monotony}. 
Inference relations which obey both of these properties are said to be \emph{cumulative}. It is well-known that most nonmonotonic formalisms are not cumulative, and several variants of standard nonmonotonic approaches have been defined in order to satisfy cumulativity (\egc \cite{Brewka91,schaub:1991a}). The last principle, (N4), is called \emph{Loop} and 
was first formulated and studied by Kraus, Lehmann, and Magidor~\shortcite{KrausLehmannMagidor90} as a property of inference relations generated by preferential model structures. 
Roughly speaking, Loop expresses a syntactic counterpart of transitivity on the model structure.

Other properties, additionally studied by Makinson~\shortcite{maki-93}, which cannot be interpreted for logic programs include \emph{Supraclassicality}, \emph{Absorption}, \emph{Distribution}, and \emph{Consistency Preservation}. We refer the reader to \cite{maki-93} for more information on these principles.

\subsubsection{Properties of Updates as a Preferential Relation}

Kraus, Lehmann, and Magidor~\shortcite{KrausLehmannMagidor90} defined \emph{preferential consequence relations} as binary relations $\tsim$ over propositional formulas satisfying the following 
properties (here, ``$\models$'' denotes   validity in classical propositional logic):

\begin{description}
        \item[(P1)] If $\models(\phi\Leftrightarrow \psi)$ and $\phi\tsim \gamma$, then $\psi\tsim \gamma$. 

        \item[(P2)] If $\models(\phi\Rightarrow \psi)$ and $\gamma\tsim \phi$, then $\gamma\tsim \psi$.

        \item[(P3)] $\phi\tsim \phi$.

        \item[(P4)] If $\phi\tsim \psi$ and $\phi\tsim \gamma$, then $\phi\tsim \psi\land\gamma$.

        \item[(P5)] If $\phi\tsim \gamma$ and $\psi\tsim \gamma$, then $(\phi\lor\psi)\tsim \gamma$.

        \item[(P6)] If $\phi\tsim \psi$ and $\phi\tsim \gamma$, then $(\phi\land\psi)\tsim \gamma$.
\end{description}
Rule (P1) is called  \emph{Left Logical Equivalence} and (P2) is the principle of \emph{Right Weakening}. Property (P3) coincides with (N1) and is referred to by Kraus, Lehmann, and Magidor as \emph{Reflexivity}. 
(P4) and (P5) are respectively called \emph{And} and \emph{Or}. The last rule, (P6), is identical with (N3), the principle of Cautious Monotony.

As noted by the above authors, any assertional relation satisfying (P1)--(P6) also satisfies Cut, expressed here as the following rule:
\begin{center}
If $\phi\land\psi\tsim\gamma$ and $\phi\tsim\psi$, then $\phi\tsim\gamma$.
\end{center}

Since not all preferential relations can be considered as reasonable inference procedures, Lehmann and Magidor~\shortcite{lehm-magi-92} subsequently defined a more restricted class of preferential relations, called \emph{rational consequence relations}. 
They show that such rational consequence relations give rise to  logical closure operations which satisfy the principle of cumulativity. 
Since none of the postulates for rational relations can be formulated for logic programs, they are not discussed here.

\subsubsection{Analysis}

 The interpretation of  postulates (N1)--(N4) in terms of update sequences is given in Table~\ref{table:props-Makinson}. The results show that (N1) and (N2) hold, whereas (N3) and (N4) fail. This corresponds to the situation of standard logic programs under the answer set semantics. Hence, in some sense, updates do not represent a loss in properties with respect to standard answer set semantics. 

Table~\ref{table:props-Makinson} contains also the interpretation of postulates (P1)--(P6). As a matter of fact, since (P3) and (P6) coincide with (N1) and (N3), respectively, and (P5) admits no interpretation in terms of logic programs, only postulates (P1), (P2), and (P4) are included in Table~\ref{table:props-Makinson}.   Like the failure of (K6) and (U4), the failure of postulate~(P1) showcases the syntax-dependency of update programs, as equivalent programs do not behave the same way under identical update information.
Proofs and counterexamples for properties (N1)--(N4), (P1), (P2), and (P4) are given in Appendix~\ref{app:nonmon}.


\begin{table}[t!]
\caption{Interpretation of Postulates $($N1\/$)$--$($N4\/$)$, $($P1\/$)$, $($P2\/$)$, and $($P4\/$)$.}\label{table:props-Makinson}
\begin{tabular}{clc}
\hline\hline
Postulate & \multicolumn{1}{c}{Interpretation} & Postulate holds \\
\hline
\bf{(N1)} & $P_1\in\bel((\useq{P},P_1))$ & yes \\[2ex]

\bf{(N2)} & If $\bigcup_{i=1}^m Q_i\subseteq\bel((\useq{P},P_1))$ and   & yes \\[.6ex]
& $P_2\subseteq\bel((\useq{P},P_1\cup \bigcup_{i=1}^m Q_1))$, then $P_2\subseteq\bel((\useq{P},P_1))$ \\[2ex]

\bf{(N3)} & If $\bigcup_{i=1}^m Q_i\subseteq\bel((\useq{P},P_1))$ and $P_2\subseteq\bel((\useq{P},P_1))$, then  & no \\[.6ex]
& $P_2\subseteq\bel((\useq{P},P_1\cup \bigcup_{i=1}^m Q_1))$ \\[2ex]

\bf{(N4)} & If $P_{i+1}\subseteq\bel((\useq{P},P_i))$ ($1\leq i<n$)  and   & no \\[.6ex]
& $P_1\subseteq\bel((\useq{P},P_n)) \ (n\geq 2)$, then \\[.6ex]
&  $\{ P' \mid P'\subseteq\bel((\useq{P},P_i)) \}  = \{ P' \mid P'\subseteq\bel((\useq{P},P_j))\}$, \\[.6ex]
& for all $i,j\leq n$ \\[2ex]

\bf{(P1)} & If $P_1 \equiv P_2$ and $P_3\subseteq\bel((\useq{P},P_1))$, then  & no \\[.6ex] 
& $P_3\subseteq\bel((\useq{P},P_2))$ \\[2ex]

\bf{(P2)} & If $P_1 \models P_2$ and $P_1\subseteq\bel((\useq{P},P_3))$, then & no \\[.6ex]
& $P_2\subseteq\bel((\useq{P},P_3))$  \\[2ex]

\bf{(P4)} & If $P_2\subseteq\bel((\useq{P},P_1))$ and $P_3\subseteq\bel((\useq{P},P_1))$, then  & yes \\[.6ex]
& $P_2 \cup P_3\subseteq\bel((\useq{P},P_1))$ \\
\hline\hline
\end{tabular}
\vspace{-2\baselineskip}
\end{table}


\subsection{Further Properties}
\label{sec:further-prop}

Rounding off our discussion on principles of update sequences, 
we describe some additional general properties which, as we believe,
updates and sequences of updates should satisfy. The given properties are not developed in a systematic manner, though, and they are not meant to 
represent an exhaustive list. Unless stated otherwise,
update programs enjoy these properties.

\begin{description}
\item[\rm\emph{Addition of Tautologies}:] \label {prop:taut} 
 If the program $Q$ contains only tautological rules
(\iec if $Q$ contains only rules of the form $L\la L$), then
$(\useq{P},Q) \equiv \useq{P}$.
\end{description}
This property is violated, which is also the case e.g.\ for dynamic
logic programs. Consider the programs $P_1=\{ a \la \ \}$, $P_2=\{
\neg a \la \ \}$, and $P_3=\{ a \la a \}$. Then $(P_1,P_2)$ has the
single answer set $\{ \neg a\}$. By updating with $P_3$, the
interaction between $P_1$ and $P_3$ generates another answer set for
$(P_1,P_2,P_3)$, namely $\{ a \}$. Note, however, that tautological
rules in updates are, as we believe, rare in practical applications
and can be eliminated easily.

\begin{description}
\item[\rm\emph{Initialization}:] 
$(\emptyset,P) \equiv P$. 
\end{description}
This property states that the update of an initial empty knowledge base yields just  the update itself.

\begin{description}
\item[\rm\emph{Idempotence}:] 
 $(P,P)\equiv P$.
\end{description}
Updating program $P$ by itself has no effect. This property is in fact a special 
case of the following principle:

\begin{description}
\item[\rm\emph{Absorption}:] 
 $(\useq{P},Q,Q)\equiv(\useq{P},Q)$.
\end{description}

The next three properties express conditions involving programs over 
disjoint alphabets. 

\begin{description}
\item[\rm\emph{Update of Disjoint Programs}:] 
 If $P = P_1 \cup P_2$  is the union of programs $P_1,P_2$ on
disjoint alphabets $\at_1$ and $\at_2$, then $P\upd Q \equiv_{\at_1\cup\at_2} (P_1\upd Q) \cup (P_2\upd Q)$. 
\end{description}

\begin{description}
\item[\rm\emph{Parallel Updates}:] 
 If $\useq{P}=(P_1\commadots P_n)$ is an update sequence over $\at$, and $Q_1$ and $Q_2$ are programs defined over disjoint alphabets $\at_1$ and $\at_2$, respectively, then 
$P_1\upd \cdots \upd P_n\upd (Q_1 \cup Q_2) \equiv_{\at\cup\at_1\cup\at_2} (P_1\upd \cdots \upd P_n\upd Q_1) \cup (P_1\upd \cdots \upd P_n\upd Q_2)$.
\end{description}
In other words, the update by non-interfering programs can be done in parallel, by merging the respective results. This property is not satisfied: Consider the case $n=1$, with  $P_1=P$ and $Q_2=\emptyset$. Assuming that the property holds, we would have $\bel(P\upd Q_1) = \bel((P \upd Q_1) \cup P)$, \iec $P$ holds in  $(P,Q_1)$ no matter what. This is quite obviously not the case.

\begin{description}
\item[\rm\emph{Noninterference}:] 
 If $P_1$ and $P_2$ are programs defined over disjoint alphabets, then
$(\useq{P},P_1,P_2) \equiv (\useq{P},P_2,P_1)$.
\end{description}
That is, the order of updates which do not interfere with each other is immaterial.

This property  is an immediate consequence of the following stronger
property: Suppose $Q \subseteq P_2$ is a program such that there are no rules $r
\in Q$ and $r' \in (P_2\setminus Q) \cup P_1$ with $\head{r}=\neg \head{r'}$. 
Then, $(\useq{P},P_1,P_2) \equiv (\useq{P},P_1\cup Q, P_2 \setminus Q)$.

\begin{description}
\item[\rm\emph{Augmented Update}:] 
 If $P_1 \subseteq P_2$, then $(\useq{P},P_1,P_2) \equiv
(\useq{P},P_2)$. 
\end{description}
Updating with additional rules makes the previous update obsolete. This property is a somewhat stronger, syntactic variant of the postulate (C1)
from above, which fails. On the other hand, it includes Absorption as 
a special case. 

Note that $(\useq{P},P_2,P_1) \equiv (\useq{P},P_2)$ does in general 
\emph{not} hold, which may be desired in some cases: Omission of a rule
$r$ in $P_2$ with respect to $P_1$ leaves the possibility to violate
$r$.

As mentioned before, a sequence of updates $\useq{P}=(P_1,\ldots,P_n)$
can be viewed either from the point of view  of ``immediate'' revision, or 
as ``logic-constrained'' revision. The following property, which
deserves particular attention, expresses equivalence of these views 
(the property is formulated for the case $n=3$):

\begin{description}
  
\item[\rm\emph{Iterativity}:] 
 For any epistemic state $P_1$ and ELPs $P_2$ and $P_3$ over $\at$,
it holds that $P_1 \upd P_2 \upd P_3 \equiv_\at (P_1 \upd P_2) \upd P_3$.
\end{description}
However, this property fails. Informally, soundness of this property would 
mean that a sequence of three updates is a shorthand for iterated update of a single program, i.e., the result of $P_1 \upd P_2$ is viewed as a singleton sequence. Stated another way, this  property would
mean that the definition for $P_1\upd P_2\upd P_3$ can be viewed as a
shorthand for the nested case. Vice versa, this property reads
as possibility to forget an update once and for all, by incorporating
it immediately into the current belief set. 

For a concrete counterexample,
consider $P_1 = \emptyset$, $P_2=\{ a\la$, \  $\neg a\la \ \}$, and $P_3 = \{ a
\la \ \}$. The program $\useq{P}_\upd = P_1\upd P_2\upd P_3$ has a unique answer set, in which $a$ is true. On the other hand, $(P_1\upd P_2)\upd P_3$ has no consistent answer set. Informally, while the ``local'' inconsistency of $P_2$ is removed in $P_1\upd P_2\upd P_3$ by rejection of the rule $\neg a\la~$ via $P_3$, a similar rejection in $(P_1\upd P_2)\upd P_3$ is blocked because of a renaming of the predicates in $P_1\upd P_2$. The local inconsistency of $P_2$ is thus not eliminated.

However, under certain conditions, which exclude such possibilities for local
inconsistencies, the iterativity property holds, given by the following result:

\begin{theorem}
\label{theo:iterate}
Let $\useq{P}=(P_1,\ldots,P_m,P_{m+1},\ldots,P_n)$, $n>m \geq 2$ be a sequence
of programs over a set of atoms $\at$. Suppose that for any conflicting rules
$r_1,r_2 \in P_i$, 
$i\leq m$, 
one of the following conditions holds: 

\begin{itemize}
        \item[(i)] There is some rule $r \in P_j$, $i<j\leq m$, such that either $\head{r}=\head{r_1}$ and $\body{r}\subseteq \body{r_1}$, or $\head{r}=\head{r_2}$ and $\body{r}\subseteq \body{r_2}$; 
        \item[(ii)] there are rules $r'_1,r'_2\in P_j$, $m<j\leq n$, such that
$\head{r_k}=\head{r'_k}$ and $\body{r'_k}\subseteq \body{r_k}$, $k\in \{1,2\}$, and no rule $r\in P_{j'}$ exists with $j<j'\leq n$ and $\head{r} = \head{r_1}$ or $\head{r} = \head{r_2}$; or
        \item[(iii)] $\body{r_1}\cup \body{r_2}$ is unsatisfiable. 
\end{itemize}

Then: 
$$P_1\upd \cdots \upd P_n \equiv_{\at}
(P_1\upd\cdots\upd P_m)\upd P_{m+1}\upd \cdots\upd P_n.$$
\end{theorem}

The proof of this theorem is technically involving and is not
presented here; it can be found in~\cite{eite-etal-00g}.
Observe that Conditions~(i)--(iii) of Theorem~\ref{theo:iterate} are simple syntactic criteria, which can be easily checked. 

A weaker version of Theorem~\ref{theo:iterate} may be applied if
updates should be incorporated instantaneously, by only considering
Condition~(iii). This condition can be locally checked on each update,
and is useful, \egc  if $P_1\upd P_2$ has already been constructed.
Since, for any programs $Q_1$ and $Q_2$, the update program $Q_1\upd
Q_2$ does not have rules with opposite heads, we can conclude from
Theorem~\ref{theo:iterate} that incorporating consecutive updates which obey
assertion (iii) is equivalent to the update program for the sequence of updates. 
\begin{theorem}
\label{theo:nested-iterate}
Let $\useq{P}=(P_1,\ldots,P_n)$, $n\geq 2$, be an update sequence on a set
of atoms $\at$.  Suppose that, for any conflicting rules $r_1,r_2 \in P_i$,
$i\leq n$, 
the union $B(r_1)\cup
B(r_2)$ of their bodies is unsatisfiable. Then:
$$(\cdots(P_1\upd P_2)\upd P_3) \cdots
\upd P_{n-1}) \upd P_n \equiv_{\at} P_1\upd P_2\upd P_3\upd\cdots\upd P_n.$$
\end{theorem}

In certain cases, the assertions in Theorem~\ref{theo:iterate} can be
dropped. One such case is a repeated update, i.e., $(P_1, P_2, P_2)$;
see \cite{eite-etal-00g} for more details.

\section{Refined Semantics: Minimal and Strictly Minimal Answer Sets}
\label{sec:refinements}

A property which update programs intuitively do not respect is
\emph{minimality of change}. In general, it is desirable to incorporate a new set of rules
$P_2$ into an existing program
$P_1$ with as little change as possible. This, of course, requests us to 
specify how similarity (or difference) between programs is understood 
and, furthermore, how
proximity 
of programs is measured. In particular, the question is
whether similarity should be model-based, or syntactically defined.

Since the semantics of update programs depends on syntax, a pure
model-based notion of similarity between logic programs seems less
appealing for defining minimality of change. A natural approach for
measuring the change which $P_1$ undergoes by an update with $P_2$ is
by considering those rules in $P_1$ which are abandoned. This leads us
to prefer an answer set $S_1$ of $\useq{P}=(P_1,P_2)$ over another 
answer set $S_2$ if $S_1$ satisfies a larger set of rules from $P_1$  
than $S_2$.

\begin{definition} \label{MIN2}
\label{def:minimal} 
Let $\useq{P}=(P_1,\dots,P_n)$ be an update sequence. 
An answer set $S \in \ASup(\useq{P})$ is \emph{minimal} iff 
there is no $S'\in \ASup(\useq{P})$ such that
$\rs(S',\useq{P}) \subset \rs(S,\useq{P})$.
\end{definition}

\begin{example}
\label{ex:minimal}
Consider the sequence $(P_1, P_2, P_3)$ from Example~\ref{example:tv}. Assume that the following additional update is received, 
describing that a TV can also be turned off:
\begin{eqnarray*}
\setlength{\arraycolsep}{0pt}
P_4 = \big\{&\hspace{-4ex}&r_8: \ 
\emph{switched\_off} \la \naf \emph{tv\_on}, \naf \emph{power\_failure}; \\ 
&\hspace{-4ex}&r_9: \ 
\emph{tv\_on} \la  \naf \emph{switched\_off}, \naf \emph{power\_failure}; \\
&\hspace{-4ex}&r_{10}: \ 
\neg \emph{tv\_on} \la \emph{switched\_off}; \\
&\hspace{-4ex}&r_{11}: \ 
\neg \emph{switched\_off} \la \emph{tv\_on} 
\ \big\}.
\end{eqnarray*}
While $(P_1, P_2, P_3)$ has the single answer set \(S_1=\{ \emph{night}, 
\neg \emph{power\_failure}, \emph{tv\_on}, \linebreak \emph{watch\_tv}\}\), 
the new sequence $\useq{P} = (P_1, P_2, P_3, P_4)$ has two answer sets: 
\(S_1 \cup \linebreak \{ \neg \emph{switched\_off} \}\) and, additionally,  
\(S_2=\{ \emph{night}, \neg \emph{power\_failure}, \emph{switched\_off}, 
\neg \emph{tv\_on}, \linebreak \emph{sleep}\}\). 
Both answer sets reject rule $r_6$,
but $S_2$ rejects 
$r_3$,
too. Thus, $S_1$ is minimal and, corresponding 
to our intuition, should be preferred to $S_2$.
\end{example}

Minimal answer sets put no further emphasis on the temporal order of
updates. Rules in more recent updates may be violated in order 
to satisfy rules from previous
updates. Eliminating this possibility leads us to the following notion:

\begin{definition}
\label{def:strict}
Let $S,S'\in \ASup(\useq{P})$,  for some update sequence $\useq{P}=(P_1,\ldots,P_n)$. Then, $S$ is
\emph{preferred} over $S'$ iff some $i \in \{ 1, \dots,n\}$ exists
such that (i) $\rs_i(S,P) \subset \rs_i(S',P)$, and (ii)~$\rs_j(S,P) = 
\rs_j(S',P)$, for all 
$j=i+1,\ldots, n$. An answer set $S$ of $\useq{P}$ is {\em strictly minimal}, 
if no $S'\in \SMup(\useq{P})$ exists 
which is preferred
over $S$.
\end{definition}

\begin{example}
\label{ex:strict}
Suppose in the previous example we had observed that the TV was off when the 
power returned, \iec replace $P_3$ in $(P_1, P_2, P_3, P_4)$ by:
\begin{eqnarray*}
P^{\prime}_3 &=& \big\{ \ r_7: \ \neg \emph{power\_failure} \la, \quad 
r^{\prime}_7: \ \neg \emph{tv\_on} \la \ \big\}.
\end{eqnarray*}
The modified update sequence $\useq{P}^{\prime}=(P_1, P_2, P^{\prime}_3, P_4)$ 
yields the same answer sets as before: 
\begin{eqnarray*} 
S_1&=&\{ \emph{night}, \neg \emph{power\_failure}, \neg \emph{switched\_off}, 
\emph{tv\_on}, \emph{watch\_tv}\}; \\
S_2&=&\{ \emph{night}, \neg \emph{power\_failure}, \emph{switched\_off}, 
\neg \emph{tv\_on}, \emph{sleep}\}.
\end{eqnarray*}
However, now both answer sets are minimal: We have 
$\rs(S_1,\useq{P}^{\prime})=\{r^{\prime}_7, r_6\}$ and
$\rs(S_2,\useq{P}^{\prime})=\{r_3, r_6\}$. 
Thus, $\rs(S_1,\useq{P}^{\prime})$ and $\rs(S_2,\useq{P}^{\prime})$ are 
incomparable, and hence both $S_1$ and $S_2$ are minimal answer sets.
Since in $S_1$ 
the more recent rule of $P^{\prime}_3$ is 
violated,
$S_2$ is the unique strictly minimal answer set.
\end{example}

We denote by $\bel_{min}(\useq{P})$  the set of
all rules which are true in any minimal answer set  of an update sequence 
$\useq{P}$.
Likewise, $\bel_{str}(\useq{P})$
denotes the set of all rules which are true in all strictly minimal answer sets 
of $\useq{P}$.

Let us consider some further example stressing the difference between regular update answer sets, minimal answer sets, and strictly minimal answer sets.

\begin{example}
\label{ex:concert}
An agent consulting different sources in search of a performance or a 
final rehearsal of a concert on a given weekend may be faced with the 
following situation. First, the agent is notified by one of the sources that 
there is no concert on Friday: 
$$
P_1 \; = \; \big\{\  r_1: \ \neg \emph{concert\_friday} \la \ \big\}.
$$ 
Later on, a second source reports that it is neither aware of a final 
rehearsal on Friday, nor of a concert on Saturday: 
$$
P_2 \; = \; \big\{\ r_2: \ \neg \emph{final\_rehearsal\_friday} \la, \quad 
r_3: \ \neg \emph{concert\_saturday} \la \ \big\}.
$$
Finally, the agent is assured that there is a final rehearsal or a concert on 
Friday and that whenever there is a final rehearsal on Friday, a concert on 
Saturday or Sunday follows:
\begin{eqnarray*}
\setlength{\arraycolsep}{0pt}
P_3 \; = \; \big\{
&\hspace{-4ex}&r_4: \ \emph{concert\_friday} \la \naf 
\emph{final\_rehearsal\_friday}; \\ 
&\hspace{-4ex}&r_5: \ \emph{final\_rehearsal\_friday} \la \naf 
\emph{concert\_friday}; \\ 
&\hspace{-4ex}&r_6: \ \emph{concert\_saturday} \la 
\emph{final\_rehearsal\_friday}, \naf \emph{concert\_sunday}; \\ 
&\hspace{-4ex}&r_7: \ \emph{concert\_sunday} \la 
\emph{final\_rehearsal\_friday}, \naf \emph{concert\_saturday} \ \big\}.
\end{eqnarray*}
The update sequence $\useq{P} = (P_1, P_2, P_3)$ yields three answer sets:
\begin{eqnarray*} 
S_1 \!\!\! & = & \!\!\! \{\emph{final\_rehearsal\_friday}, \neg \emph{concert\_friday}, 
\emph{concert\_saturday}\}; \\
S_2 \!\!\! &=&\!\!\! \{\emph{final\_rehearsal\_friday}, \neg \emph{concert\_friday}, 
\neg \emph{concert\_saturday}, \emph{concert\_sunday}\};\\
S_3\!\!\!  &=& \!\!\! \{\neg \emph{final\_rehearsal\_friday}, \emph{concert\_friday}, 
\neg \emph{concert\_saturday}\}, 
\end{eqnarray*}
The corresponding rejection sets are:
\begin{eqnarray*}
\rs(S_1,\useq{P}) &=& \{r_2, r_3\}; \\
\rs(S_2,\useq{P}) &=& \{r_2\};\\
\rs(S_3,\useq{P}) &=& \{r_1\}.
\end{eqnarray*}
 Thus, $S_2$ and $S_3$ are minimal answer sets, with $S_3$ being the single strictly minimal answer set of $\useq{P}$. 
\end{example}

Clearly, every strictly minimal answer set is minimal, but not vice versa.
It is easily seen that for the case of update sequences involving only two update programs, \iec
for update sequences of the form $\useq{P}=(P_1,P_2)$, the notions of strictly minimal 
answer sets and minimal answer sets coincide.
As for the AGM
postulates, inspection shows that minimal and
strictly minimal answer sets satisfy the same postulates as
regular update answer sets, with the exception that (K3) and (K4) 
hold for the former ones. 

Concerning the implementation of minimal and strictly minimal answer sets, 
in Section~\ref{app:implementation} we will show how 
they can be characterized  in terms of ELPs.

\section{Computational Issues}
\label{sec:computation}

\subsection{Complexity}\label{subs:compl}

In this section, we 
address the computational complexity of
update programs.
We assume that
the reader 
is familiar with the basic concepts of
complexity theory; \egc \cite{john-90} and \cite{papa-94} are good sources (for 
complexity results in logic programming, cf.\ \cite{schl-95a,eite-gott-93b,dant-etal-97}). 
In our analysis, we focus
on the case of finite, propositional update sequences.

We briefly recall the 
definitions of the complexity classes
relevant  in the following analysis. The class \NP\ consists  of all  
 decision problems which are solvable  in polynomial time using a
nondeterministic Turing machine, and
$\SigmaP{2}$ is the class of all decision problems
solvable by a nondeterministic Turing machine in polynomial time with
access to an oracle for problems in $\NP$ ($\SigmaP{2}$ is also written
 as $\NPNP$).  
Furthermore, $\coNP$ refers to the class of problems whose
complementary problems are in \NP, and $\PiP{2}$ contains the 
complements of the problems in $\SigmaP{2}$.\footnote{Two decision 
problems, $D_1$ and $D_2$, are complementary (or, $D_1$ and $D_2$ are complements of each other) if it holds that $I$ 
is a yes-instance of $D_1$ exactly if $I$ is a no-instance of $D_2$.} 
All the mentioned classes belong to the \emph{polynomial hierarchy}: 
\NP\ and \coNP\ are at the first level of the hierarchy, and 
$\SigmaP{2}$ and $\PiP{2}$ are the second level. As well, 
$\NP\subseteq\SigmaP{2}$ and $\coNP\subseteq\PiP{2}$.
It is widely held that these inclusions are proper. 

It is clear that the complexity of normal logic
programs, which resides at the first level of the polynomial hierarchy
\cite{mare-trus-91}, is a lower bound for the complexity of
update programs.
For arbitrary updates, the complexity does not increase, even if we
consider a sequence of updates. 

\begin{theorem} 
\label{theo:comp-1}
Given an update sequence $\useq{P}=(P_1\commadots P_n)$ over a set
of atoms $\at$, then:

\begin{itemize}
\item[(i)]
determining whether
 $\useq{P}$ has an answer set is
$\NP$-complete;
\item[(ii)] determining whether
$L \in \bel (\useq{P})$ for some literal $L$ is $\coNP$-complete.  
\end{itemize}
Hardness holds in both cases
for $n=1$.
\end{theorem}

\begin{proof}
The program $\useq{P}_\upd = P_1\upd P_2 \upd \cdots \upd P_n$ can 
obviously be
generated in polynomial time from $\useq{P}=(P_1\commadots P_n)$.
 Furthermore, deciding consistency of $\useq{P}_\upd$ is in \NP, 
and checking whether $L \in \bel (\useq{P}_\upd)$ is in \coNP. 
This proves membership. 
\NP-hardness and \coNP-hardness of the respective tasks
is
inherited from the complexity of normal logic programs \cite{mare-trus-91}.
\end{proof}

Under minimal updates, the complexity of updates increases by one
level in the polynomial hierarchy. This is no surprise, though, and
parallels analogous results on update logic programs by Sakama and
Inoue~\shortcite{inou-saka-99} as well as previous results on updating logical
theories and iterated circumscription \cite{eite-gott-92e,eite-gott-93b}.

\begin{theorem} 
\label{theo:comp-2}
Given an update sequence $\useq{P}=(P_1\commadots P_n)$  over a set 
of atoms $\at$ and some rule
$r$,  the following two problems are $\PiP{2}$-complete:
\begin{itemize}
\item[(i)] determining whether $r \in \bel_{min}(\useq{P})$; and
\item[(ii)] determining whether $r \in \bel _{str}(\useq{P})$.
\end{itemize}
Hardness holds even if $n=2$. 
\end{theorem}

\begin{proof}
We first show that the two tasks are in $\PiP{2}$. We treat only task (i); 
the case of task (ii) is analogous. In order to show membership of (i) in 
$\PiP{2}$, we show that the complementary problem is in $\SigmaP{2}$. To 
disprove $r \in \bel_{min}(\useq{P})$,
we can construct the update program $\useq{P}_\upd = P_1\upd P_2 
\upd\cdots \upd P_n$ in
polynomial time from $\useq{P}$ and guess an answer set
$\corr{S}\subseteq \at^\ast$ of $\useq{P}_\upd$ such that 
$\corr{S}\not\models r$ and where $S=\corr{S}\cap\at$ is a
 minimal answer set of $\useq{P}$ (recall that $S\in\SMup(\useq{P})$ 
is minimal iff there is no $T\in \SMup(\useq{P})$ such that
$\rs(T,\useq{P}) \subset \rs(S,\useq{P})$).
With the help of an
$\NP$-oracle, the guess for $\corr{S}$ can be verified in
polynomial time. This concludes the proof that checking
 whether $r\not\in \bel_{min}(P)$ 
is in $\SigmaP{2}$.

Hardness of both tasks is shown by a simple reduction 
from the $\PiP{2}$-hard
irrelevance test in abduction from normal logic programs
\cite{eite-etal-97q}, which is the following problem: 
Given a normal logic program $P$, a set of atoms
$H$, a set of literals $M$, and an atom $h_0\in H$, decide whether
$h_0$ is not contained in any
minimal brave explanation of $M$, i.e., decide whether $h_0
\notin E$ holds for each minimal
 $E\subseteq H$ (with respect to inclusion) such
that $P\cup E$ has some stable model $S$ with $S\models L$, for all
$L\in M$. 

The reduction is defined as follows:
For each $h\in H$, let $h'$ and $h''$ be fresh atoms, and 
consider the update sequence $\useq{P}=(P_1,P_2)$, where
\begin{eqnarray*}
P_1 &=& \{
\neg h' \la~ \mid h \in H\}, \\
P_2 &=& P \cup \{ ~\la \naf L \mid L \in M\} \cup \{ h \la h',\, h' \la
\naf h'',\, h'' \la \naf h' \mid h\in H\}.
\end{eqnarray*}
It can be shown that there is a one-to-one correspondence 
between the rejection sets $\rs(S,\useq{P})$ 
of the minimal answer sets
$S$ of $\useq{P}$ and 
the minimal brave explanations
$E$ for $M$. In particular, the rule $\neg h_0'\la\,$ is 
in $Rej(S)$ iff
$h_0$ belongs to the corresponding minimal explanation $E$.
It follows that 
$h_0 \la \ \in \bel_{min}(\useq{P})$ iff $h_0$ is not contained in any
brave explanation for $M$, which establishes
$\PiP{2}$-hardness of (i). Since the notions of minimal and 
strictly minimal answer sets coincide for update sequences 
of length 2, we have that $\bel_{min}(\useq{P})=\bel_{str}(\useq{P})$. 
Thus, $\PiP{2}$-hardness of (ii) holds as well. 
\end{proof}

Similar results hold in the approach of 
Inoue and Sakama~\shortcite{inou-saka-99}. Furthermore, they 
imply that 
minimal and strictly minimal answer sets can be
polynomially translated into disjunctive logic programming, which 
can serve as a basis for implementation purposes.
The next section deals with some algorithmic issues.

\subsection{Implementation}
\label{app:implementation}

The notion of update sequence can be easily extended to the case
where rules may contain variables. As usual, the semantics of a program
$P$ containing variables is defined in terms of the semantics
of its ground instances $P^*$ over the Herbrand base. 
Rules $r$ with variables $\trmlist{X}=X_1,\ldots,X_n$ are denoted $r(\trmlist{X})$, and rejection of $r$ is represented by a
predicate $\rejf{r}{\trmlist{X}}$; further details, can be found in
\cite{eite-etal-00g}. In the rest of this section, we consider function-free
update sequences $\useq{P}$.  

Since answer sets of (first-order) update sequences are defined by
answer sets of (first-order) ELPs, it is relative straightforward to
implement the current update approach.  In fact, an implementation can
be built on top of existing solvers for answer set semantics.  In the
present case, we implemented updates as a front-end for the logic
programming tool \DLV~\cite{eite-etal-97a,eite-etal-98a}.
The latter system is a state-of-the-art solver for \emph{disjunctive
logic programs} (DLPs) under the answer set semantics.  It allows for
non-ground rules and calculates answer sets by performing a reduction
to the stable model semantics.  (Another highly efficient logic
programming implementation, realizing stable model semantics, is the
system \smodels~\cite{niem-simo-96b}, which would similarly fit as
underlying reasoning engine. We chose \DLV\ because of familiarity and
its optimization techniques for grounding, as well as its
expressiveness which would allow an integral solution to compute
strict and strictly minimal answer sets, respectively.) Formally, DLPs
are defined as ELPs where disjunctions may appear in the head of
rules; the answer set semantics for DLPs is due to Gelfond and
Lifschitz~\shortcite{gelf-lifs-91}.

The implemented tasks agree with the decision problems 
discussed in the previous section, \iec they comprise 
the following problems:  
(i) 
checking the existence of an answer set for a given update sequence, 
(ii) 
brave reasoning, 
and (iii) 
skeptical reasoning; as well as the corresponding problems 
for minimal and strictly minimal answer sets.
All of these tasks have been realized for first-order 
update sequences, employing the advanced grounding mechanism of \DLV. 

As regards the implementation for minimal and strictly minimal update 
answer sets, although in principle it is possible to express the 
corresponding reasoning tasks in terms of DLPs 
(which is a consequence of Theorem~\ref{theo:comp-2} and well-known 
expressibility results for the disjunctive answer set 
semantics~\cite{Eiter:1997:DD}),
we chose instead to pursue a two-step evaluation approach for 
our purposes, remaining within the present non-disjunctive 
framework, and, at the same time, adhering more closer to 
the underlying intuitions.
Roughly speaking, this two-step approach can be described as follows:
First, all candidates for 
minimal (strictly minimal) answer sets are calculated, \iec all answer sets 
of the update 
program $\useq{P}_{\upd}$. Afterwards, every candidate is tested for being
 minimal (strictly minimal). 

Testing a candidate, $S$, for minimality (strict minimality) is performed 
by evaluating a test 
program, $\useq{P}^{min}_S$  ($\useq{P}^{strict}_S$), consisting of the rules 
of $\useq{P}_{\upd}$ and a set 
of additional rules. Intuitively, the additional rules constrain 
the answer sets of 
$\useq{P}^{min}_S$ ($\useq{P}^{strict}_S$) to those answer sets of 
$\useq{P}_{\upd}$ having a smaller set of 
rejected rules compared to the rules rejected by $S$ 
(or to those answer sets of 
$\useq{P}_{\upd}$ preferred over $S$, respectively). Hence, the 
candidate $S$ is minimal (strictly minimal) 
if the corresponding test program $\useq{P}^{min}_S$ 
($\useq{P}^{strict}_S$) has no answer set. In 
the following subsections, the implementation approach is described more 
formally.

\subsubsection{Minimal Answer Sets}
\label{imp:min}

\begin{definition} 
\label{def:mintest-program}
Let $\useq{P}_{\upd}=P_1 \upd \dots \upd P_n$ be a 
(first-order)
update program and S an answer set of $\useq{P}_{\upd}$. Let $ok$ be a
new nullary predicate symbol (\iec propositional atom) and, for
each predicate symbol $\rejfo{r}$ occurring in $\useq{P}_{\upd}$, let
$s_r$ be a new predicate symbol of the same arity as $\rejfo{r}$.
Then, the 
minimality-test program with respect to $S$,
$\useq{P}^{min}_S$, consists of all rules and constraints
of $\useq{P}_{\upd}$, together with the following items:

\begin{enumerate}
\item[(i)] \label {minrule:greater} 
for each predicate symbol $\rejfo{r}$ occurring in $\useq{P}_{\upd}$: 
\begin{eqnarray*}
\la & \rejf{r}{\trmlist{X}}, \naf s_r(\trmlist{X});
\end{eqnarray*} 
\item[(ii)] \label {minrule:less} 
for each ground formula $\rejf{r}{\trmlist{t}} \in S$:
\begin{eqnarray*}
ok & \la & \naf \rejf{r}{\trmlist{t}};\\
s_r(\trmlist{t}) & \la & ;
\end{eqnarray*}
\item[(iii)] \label {minrule:equal} 
the constraint 
\begin{eqnarray*}
& \la & \naf ok.
\end{eqnarray*} 
\end{enumerate}
\end{definition} 
Note that in the above definition, only the rules and facts of 
(ii)
manifest the dependence of $\useq{P}^{min}_S$ from $S$. Informally, the constraints (i)
eliminate all answer sets with rejection sets 
which cannot be subsets 
of $\rs(S, \useq{P})$, \iec which reject at least one rule not 
rejected in $S$. 
In the remaining answer sets, if any, either $ok$ is true, 
\iec at least one rule 
which is rejected in $S$ is not rejected in such a set, or $ok$ is false, in which case their rejection sets equal $\rs(S, \useq{P})$, 
and thus these sets are eliminated by Constraint~(iii).
Actually, the following proposition holds:

\begin{figure}[tbp]
\figrule
\programmath
\[ 
\begin{array}{l}
\textbf{Algorithm}~\textrm{Compute\_Minimal\_Models(}\useq{P}\textrm{)} \\[1ex]
\textbf{Input:}~\textrm{A~sequence~of~ELPs}~\useq{P}=(\mathit{P_1\textrm{\commadots}P_n})\textrm{.}\\
\textbf{Output:}~\textrm{All~minimal~answer~sets~of}~\useq{P}\textrm{.} \\[1.5ex]
\textbf{var}~Cands : SetOfAnswerSets;\\
\textbf{var}~MinModels : SetOfAnswerSets;\\
Cands := Compute\_Answer\_Sets(\useq{P}_{\upd});\\
\textbf{for}~\textrm{all}~\mathit{S} \in Cands~\textbf{do}\\
\hspace*{1cm}\textbf{var}~Counter : SetOfAnswerSets;\\
\hspace*{1cm}Counter := Compute\_Answer\_Sets(\useq{P}^{\mathit{min}}_{\mathit{S}});\\
\hspace*{1cm}\textbf{if}~(Counter = \emptyset)~\textbf{then}\\
\hspace*{2cm}MinModels := MinModels \cup \{\mathit{S}\};\\
\hspace*{1cm}\textbf{fi}\\
\textbf{rof}\\
\textbf{return}~MinModels;\\
\end{array}
\]
\unprogrammath
\caption{\label{fig:min-algorithm} Algorithm to calculate minimal 
answer sets.}
\figrule
\end{figure}

\begin{theorem} \label {prop:min}
Let $S$ be an answer set of $\useq{P}_{\upd}$. Then, $S$ is a minimal answer set of 
$\useq{P}_{\upd}$ iff $\useq{P}^{min}_S$ has no answer set.
\end{theorem}

\begin{proof}
\noindent
\emph{Only-if part.} Suppose $\useq{P}^{min}_S$ has an answer set 
$S^{\prime}$. 
Then $ok$ must be true in $S^{\prime}$ due to the constraint 
(iii) of Definition~\ref{def:mintest-program}.
Since rules (ii) of Definition~\ref{def:mintest-program}
are the only 
ones in $\useq{P}^{min}_S$ with head $ok$, there exists a ground term 
$\rejf{r}{\trmlist{t}} \in S$ such that $\rejf{r}{\trmlist{t}}\notin S^{\prime}$. 
Moreover, no ground term $\rejf{{r^{\prime}}}{\trmlist{t}} 
 \in S^{\prime}\setminus S$ can exist due to the constraints (i) of
Definition~\ref{def:mintest-program}.  
(Observe that 
$\rejf{{r^{\prime}}}{\trmlist{t}} \notin S$ implies 
$s_{r^{\prime}}(\trmlist{t}) \notin S^{\prime}$; hence, if 
$\rejf{{r^{\prime}}}{\trmlist{t}} \in S^{\prime}$, then the body of one of the constraints is true in $S^{\prime}$.) This proves $\rs(S^{\prime}, \useq{P}^{min}_S)=\rs(S^{\prime}, \useq{P}) 
\subset \rs(S, \useq{P})$.

Since the predicate symbols $ok$ and $s_r$ do not occur in $\useq{P}_{\upd}$, and 
$\useq{P}^{min}_S$ contains all rules and constraints of 
$\useq{P}_{\upd}$, results on the splitting of logic programs~\cite{lifs-turn-94} imply that 
$\tilde{S}=S^{\prime}\setminus(\{ok\}\cup\{s_r(\trmlist{t})\mid s_r(\trmlist{t})\in S^{\prime}\})$ 
is an answer set of $\useq{P}_{\upd}$. Given that 
$\rs(\tilde{S}, \useq{P})=\rs(S^{\prime}, \useq{P}) \subset \rs(S, \useq{P})$, 
we obtain that $S$ is not minimal.

\smallskip\noindent
\emph{If part.} Suppose $S$ is not minimal. Then there exists an 
answer set 
$\tilde{S}$ of $\useq{P}_{\upd}$ with 
$\rs(\tilde{S}, \useq{P}) \subset \rs(S, \useq{P})$. 
Consider $S^{\prime}=\tilde{S} \cup \{ok\} 
\cup \{s_r(\trmlist{t})\ |\ \rejf{r}{\trmlist{t}} \in S\}$. 
It is easily verified that $\tilde{S}$ is an answer set of $\useq{P}^{min}_S$.
\end{proof}

This result allows us to calculate all minimal answer sets of 
$\useq{P}_{\upd}$ using the 
straightforward algorithm depicted 
in Figure \ref{fig:min-algorithm}, which proceeds as follows: 
compute all answer sets of $\useq{P}_{\upd}$ and check for every 
answer set $S$ if the corresponding minimality-test program 
$\useq{P}^{min}_S$ has at least one answer set. 
If not, then add $S$ to the set of minimal answer sets of $\useq{P}_{\upd}$.

\subsubsection{Strictly Minimal Answer Sets}
\label{imp:strict}

\begin{definition} 
\label{def:stricttest-program}
Let $\useq{P}_{\upd}=P_1 \upd \dots \upd P_n$ be a
(first-order) 
update program  and S an answer set of $\useq{P}_{\upd}$. 
Let $ok$, $ok_i$ ($1\leq i\leq n$), and $eq_i$ ($1\leq i\leq n+1$) 
be new nullary predicate symbols, and, for each predicate symbol
$\rejfo{r}$ occurring in $\useq{P}_{\upd}$, let $s_r$ be a new
predicate symbol of the same arity  as $\rejfo{r}$.
Then, the program
$\useq{P}^{strict}_S$ consists of all rules and constraints of 
$\useq{P}_{\upd}$, together with the following items:  
\begin{enumerate}
\item[(i)] \label {strictrule:greater} 
for each predicate $\rejfo{r}$ occurring in $\useq{P}_{\upd}$, corresponding to $r \in P_i$:
\begin{eqnarray*} 
& \la & \rejf{r}{\trmlist{X}}, \naf s_r(\trmlist{X}), eq_{i+1};
\end{eqnarray*} 
\item[(ii)] \label {strictrule:less} 
for each ground term $\rejf{r}{\trmlist{t}} \in S$, corresponding to $r \in P_i$:
\begin{eqnarray*}
ok_i & \la & \naf \rejf{r}{\trmlist{t}}, eq_{i+1};\\
s_r(\trmlist{t}) & \la & ;
\end{eqnarray*}
\item[(iii)] \label {strictrule:level-equal}
for $1 \leq i \leq n$:
\begin{eqnarray*}
eq_i & \la & eq_{i+1}, \naf ok_i;\\
ok & \la & ok_i;
\end{eqnarray*}
\item[(iv)] \label {strictrule:equal} 
the constraint 
\begin{eqnarray*}
\hphantom{eq_i} & \la & \naf ok\hphantom{_i; eq_{i+1},}
\end{eqnarray*}
and the fact
\begin{eqnarray*}
eq_{n+1} & \la & .\hphantom{\naf ok}
\end{eqnarray*} 
\end{enumerate}
\end{definition} 

Again, program $\useq{P}^{strict}_S$ depends on $S$ only in virtue of Item~(ii). The constraints of Item~(i)
eliminate all answer sets $S^{\prime}$ which 
cannot be preferred over $S$ because at some level $i$ they reject 
a rule not rejected in $S$, and $\rs_j(S, \useq{P}) = \rs_j(S^{\prime}, 
\useq{P})$ holds for $j=i+1 \commadots n$ (expressed by $eq_{i+1}$). 
In the remaining answer sets, if there is any, $ok$ is either true, or false. 
If $ok$ is true in $S^{\prime}$, then $ok_i$ is true in $S^{\prime}$ for 
some level $i$, \iec $S^{\prime}$ does not reject a rule of $P_i$ which 
is rejected in $S$, and $\rs_j(S, \useq{P}) = \rs_j(S^{\prime}, \useq{P})$ 
for $j=i+1 \commadots n$. In this case, $S^{\prime}$ is preferred over $S$. 
If, however, $ok$ is false in $S^{\prime}$, then $\rs_i(S, \useq{P}) = 
\rs_i(S^{\prime}, \useq{P})$, for $i=1 \commadots n$, and $S^{\prime}$ 
is killed by the constraint of Item~(iv).

An equivalent result as for minimality-test programs holds for the above test programs as well. Hence,
the same algorithm using 
$\useq{P}^{strict}_S$ instead of $\useq{P}^{min}_S$ can be used to compute 
all strictly minimal answer sets of $\useq{P}_{\upd}$.
\begin{theorem} 
Let $S$ be an answer set of $\useq{P}_{\upd}$. Then, $S$ is a strictly minimal answer set 
of $\useq{P}_{\upd}$ iff $\useq{P}^{strict}_S$ has no answer set.
\end{theorem}

\begin{proof}
\noindent
\emph{Only-if part.} Suppose $\useq{P}^{strict}_S$ has an answer set 
$S^{\prime}$. Then $ok$ must be true in $S^{\prime}$, due to Constraint 
 (iv) of Definition~\ref{def:stricttest-program}.
Since the rules of Item (iii) of Definition~\ref{def:stricttest-program}
are the only ones in $\useq{P}^{strict}_S$ 
with head $ok$, there exists some $i$, $1 \leq i \leq n$, such that 
$ok_i \in S^{\prime}$. This implies that the body of the corresponding 
rule of (ii)
must be true in $S^{\prime}$. Hence, 
there exists a ground term $\rejf{r}{\trmlist{t}} \in S \setminus S^{\prime}$, where $r \in P_i$ and such that $eq_{i+1} \in S^{\prime}$. 
Moreover, except for the fact $eq_{n+1}\la \ $, the rules of (iii)
are the only ones in $\useq{P}^{strict}_S$ 
with $eq$ predicate symbols in their heads, so that $eq_{i+1}$ 
implies $eq_j \in S^{\prime}$ and $ok_j \notin S^{\prime}$ for 
$j=i+1 \commadots n$ if $i<n$. From this, and the constraints of (i),
it follows that no ground 
term $\rejf{{r^{\prime}}}{\trmlist{t}} S^{\prime}\setminus S$, 
$r^{\prime} \in P_j$, $j=i \commadots n$, can exist 
($\rejf{{r^{\prime}}}{\trmlist{t}} \notin S$ implies 
$s_{r^{\prime}}(\trmlist{t}) \notin S^{\prime}$; hence, having 
$\rejf{{r^{\prime}}}{\trmlist{t}} \in S^{\prime}$ and $eq_{j+1} \in 
S^{\prime}$, the body of one of the constraints is true in $S^{\prime}$). 
It also follows that for every $r^{\prime} \in P_j$, $j=i+1 \commadots n$, 
if $\rejf{{r^{\prime}}}{\trmlist{t}} \in S$, 
then $\rejf{{r^{\prime}}}{\trmlist{t}} \in S^{\prime}$ 
(otherwise the body of one of the rules of (ii)
 would be true in $S^{\prime}$, implying $ok_j \in S^{\prime}$, 
a contradiction). 

Summarizing, we have shown 
 $\rs_i(S^{\prime}, \useq{P}^{strict}_S) = 
\rs_i(S^{\prime}, \useq{P}_{\upd}) \subset 
\rs_i(S, \useq{P}_{\upd})$ and $\rs_j(S^{\prime}, \useq{P}^{strict}_S) = 
\rs_j(S^{\prime}, \useq{P}_{\upd}) = \rs_j(S, \useq{P}_{\upd})$, for 
$j=i+1 \commadots n$. So, $S^{\prime}$ is preferred over $S$.

Since none of the predicate symbols $ok$, $eq$, and $s_r$ occurs in 
$\useq{P}_{\upd}$, and $\useq{P}^{strict}_S$ contains all rules and 
constraints of $\useq{P}_{\upd}$, by a similar argument as in the proof of Theorem~\ref{prop:min} (\iec invoking splitting results from \cite{lifs-turn-94}) it follows that 
$$
\tilde{S}=S^{\prime}\setminus(\{ok, ok_i\}\cup\{eq_j\ |\ j=i+1 \commadots n+1\}\cup
\{s_r(\trmlist{t}) \mid s_r(\trmlist{t}) \in S^{\prime}\})
$$ 
is an answer set of 
$\useq{P}_{\upd}$. Given that $\rs(\tilde{S}, \useq{P}_{\upd})=
\rs(S^{\prime}, \useq{P}_{\upd})$, we obtain that 
$\tilde{S}$ is preferred over $S$. Consequently, $S$ is not 
a strictly minimal answer set.

\smallskip\noindent
\emph{If part.} Suppose $S$ is not a strictly minimal answer set. Then 
there exists an answer set $\tilde{S}$ of $\useq{P}_{\upd}$ which is 
preferred over $S$. In particular, there  exists some $i$, 
$1 \leq i \leq n$, such that 
$\rs_j(\tilde{S}, \useq{P}_{\upd}) = \rs_j(S, \useq{P}_{\upd})$  for 
$j=i+1 \commadots n$ and 
$\rs_i(\tilde{S}, \useq{P}_{\upd}) \subset \rs_i(S, \useq{P}_{\upd})$. 
Consider 
$$
S^{\prime}=
\tilde{S} \cup \{ok, ok_i\} \cup \{eq_j\ |\ j=i+1 \commadots n+1\} \cup 
\{s_r(\trmlist{t})\ |\ \rejf{r}{\trmlist{t}} \in S\}.
$$ 
It is easily verified that $\tilde{S}$ is an answer set of 
$\useq{P}^{strict}_S$.
\end{proof}

\section{Relations to Other Approaches}
\label{sec:rel-work}

In this section, we analyze the relations between the current update framework and other formalisms. First of all, we discuss the connection with inheritance programs by Buccafurri \emph{et al.}~\shortcite{bucc-etal-99a-iclp}, which has not been introduced as a formalism for updates but can be successfully interpreted as such, coming to an equivalence result with our update sequences over the common fragment. 

In a second step, we study the relation on the one hand to the
approach of Leite and Pereira~\shortcite{leit-pere-98}, also modeling
Revision Programming by Marek and
Truszczy{\'n}ski~\shortcite{mare-trus-94}, on the other hand to
dynamic logic programming~\cite{alfe-etal-98,alfe-etal-99a}, which is
close in spirit to the present update method, in the sense that update
sequences are translated to standard logic programs. In particular, we
describe the semantical differences between our update programs and
dynamic programs, and identify
 a wide class of logic programs for 
which they yield the same results.
 
Then, we briefly discuss update approaches for logic programs  based on preference handling~\cite{foo-zhan-98} and abduction~\cite{inou-saka-99}. 
Finally, we mention a method due to Delgrande, Schaub, and Tompits~\shortcite{delg-etal-00} for handling preference information in the context of logic programs, which is also based on an encoding to ELPs.

\subsection{Relation to Inheritance Programs}
\label{sec:inheritance}

The update semantics we suggest resolves conflicts by assigning ``preference'' to the more recent information. As already pointed out earlier, this can also be interpreted as some form of inheritance mechanism, where the more recent information is considered as being more specific. In this section, we discuss this aspect in more detail. To wit, we consider the inheritance approach introduced by Buccafurri \emph{et al.}~\shortcite{bucc-etal-99a-iclp} and we show that update sequences can equivalently be described in terms of inheritance programs.

In what follows, we briefly describe the basic layout of the inheritance approach by Buccafurri \emph{et al.}~\shortcite{bucc-etal-99a-iclp}. Since that method has originally been specified for non-ground DLPs, and we deal here only with non-disjunctive ELPs, we adapt some of the original definitions accordingly. 

A \emph{$\DLPINH$-program}, $\P$, is a finite set $\{\langle o_1,P_1\rangle\commadots\langle o_n, 
P_n\rangle\}$ of object identifiers $o_i$ $(1\leq i\leq n)$ and associated 
ELPs $P_i$, together with a strict partial order ``$<$'' 
between 
object identifiers (
pairs $\langle o_i,P_i\rangle$ are 
called \emph{objects}).\footnote{Strictly speaking, in the current 
context, the term ``$\DLPINH$-program'' (as introduced by 
Buccafurri \emph{et al.}~\shortcite{bucc-etal-99a-iclp}) is a bit of a misnomer, because ``DLP'' 
points to \emph{disjunctive} logic programs; however, we retained 
the original name for reference's sake.}
As well, we say that $\P$ is a $\DLPINH$-program \emph{over} a set of atoms $\at$ 
iff $\at$ denotes the set of all atoms appearing in $\P$. 
Informally, possible conflicts in determining properties
of the objects 
are resolved in favor
of rules which are \emph{more specific}
according to the hierarchy, in the sense that rule $r\in P_{k}$ is considered to represent more specific information than rule $r'\in P_{l}$ whenever $o_{k}<o_{l}$ holds ($1\leq k,l\leq n$ and $k\neq l$).
In the following, $\rulems{\P}$ denotes the multiset of all rules appearing in the programs 
of $\P$. 

Consider some $\DLPINH$-program  $\P$  over a set of atoms $\at$. Let $I\subseteq \extat$ be an interpretation and let
$r_1\in P$ and $r_2\in P'$ be two conflicting rules, where  $\langle o, P\rangle, \langle o',P'\rangle\in\P$. 
Then, $r_1$ \emph{overrides
$r_2$ in} $I$ iff 
(i)
$o < o'$,
(ii) $\head{r_1}$ is true in $I$,
and (iii) $\body{r_2}$ is true in $I$.
A rule $r\in\rulems{\P}$ is \emph{overridden in} $I$ iff there 
exists some  $r'\in\rulems{\P}$ which overrides $r$ in $I$.

An interpretation $I \subseteq \extat$ is a \emph{model} of $\P$ iff 
every rule in $\rulems{\P}$ is either overridden or true in $I$;
moreover, $I$ is \emph{minimal} iff it is the
least model of all these rules. 
The \emph{reduct}, $G_I(\P)$, of the $\DLPINH$-program $\P$ \emph{relative} to $I$ results from $\rulems{\P}$ by
(i)
deleting any rule $r\in\rulems{\P}$ which is either overridden 
in $I$ or defeated by $I$; and
(ii)
deleting all weakly negated literals in the bodies of the 
remaining rules of $\rulems{\P}$.
Then, $I$ is an \emph{answer set} of $\P$ iff it is a minimal 
model of $G_I(\P)$.

This concludes our brief review of the inheritance framework 
of~\cite{bucc-etal-99a-iclp}; we continue with our correspondence 
result. 

\begin{theorem}
\label{theo:elp-upd-inh}
$S \subseteq \extat$ is an answer set of the update sequence $\useq{P}=(P_1,\ldots,P_n)$ over $\at$ iff $S$  is an answer
set of the $\DLPINH$-program  
$\P=\{\langle o_1,P_1\rangle \commadots \langle o_n,P_n\rangle\}$ 
 having inheritance order $o_n<o_{n-1}<\cdots<o_1$.
\end{theorem}

\begin{proof}
We first note the following two properties, which can be verified in a straightforward way. Let  
$I\subseteq\extat$ be some interpretation and $r\in\rulems{\P}$. Then:
\begin{itemize}
\item[(i)] If $r\in\rs(I,\useq{P})$, then $r$ is overridden in $I$.
\item[(ii)] Assume $I$ satisfies for any $r'\in\rulems{\P}$ the condition that $\body{r'}$ is true in $I$ whenever $\head{r'}$ is true in $I$. Then, $r$ is overridden in $I$ only if $r\in\rs(I,\useq{P})$.
\end{itemize}
We proceed with the proof of the main result. Suppose $S$ is an answer set of $\useq{P}=(P_1,\ldots,P_n)$. We show $S$ is an answer set of $\P=\{\langle o_1,P_1\rangle \commadots \langle o_n,P_n\rangle\}$ 
 with inheritance order $o_n<o_{n-1}<\cdots<o_1$.

First, we show that $S$ is a model of $G_S(\P)$. Consider some $r\in G_S(\P)$. Then, there is some rule $\hat{r}\in\rulems{\P}$ such that $r=\reductr{\hat{r}}$ and $\hat{r}$ is neither overridden in $S$ nor defeated by $S$.
Applying Property~(i), we get that $\hat{r}\notin\rs(S,\useq{P})$. Hence, $\hat{r}\in(\cup\useq{P}\setminus\rs(S,\useq{P}))^S$ since $\hat{r}$ is not defeated by $S$. Thus, given that $S$ is an answer set of $\useq{P}$, Theorem~\ref{theo:sequence-char-0} implies that $r=\reductr{\hat{r}}$ is true in $S$. It follows that $S$ is a model of $G_S(\P)$. 

Next, we show that there is no proper subset of $S$ which is also a model of $G_S(\P)$. Suppose there is such a set $S_0\subset S$. Since $S$ is an answer set of $\useq{P}$, Property~(ii) can be applied, and we obtain $r\in G_S(\P)$ whenever $r\in (\cup\useq{P}\setminus\rs(S,\useq{P}))^S$ holds.
As a consequence, $S_0$ is a model of $r\in (\cup\useq{P}\setminus\rs(S,\useq{P}))^S$, which contradicts the fact that $S$ is an answer set of $\useq{P}$.
This concludes the proof that $S$ is an answer set of $\P$ if $S$ is an answer set of $\useq{P}$.

For the converse direction, assume $S$ is an answer set of $\P$. Similar to the argumentation given above, Property~(ii) implies that $S$ is a model of $(\cup\useq{P}\setminus\rs(S,\useq{P}))^S$. As well, $S$ is a minimal model of $(\cup\useq{P}\setminus\rs(S,\useq{P}))^S$, because otherwise Property~(i) would yield a proper subset $S_0$ of $S$ being a model of $G_S(\P)$, contradicting the fact that such a subset $S_0$ cannot exist, because $S$ is an answer of $\P$. 
\end{proof}

Inheritance programs are also related to \emph{ordered logic programs}, due to Laenens \emph{et al.}~\shortcite{LaeSacVer90} and further analyzed by Buccafurri \emph{et al.}~\shortcite{bucc-etal-96}. The difference between inheritance programs and ordered logic programs is that the latter ones have a built-in \emph{contradiction removal} feature, which eliminates local inconsistencies in a given hierarchy of programs. Thus, for linearly ordered programs $P_1<P_2<\cdots < P_n$ where such inconsistencies do not occur, \egc if for any two
conflicting rules in $P_i$ ($1\leq i\leq n$)  their bodies cannot be  simultaneously satisfied,  the above equivalence result holds for ordered logic programs as well.

\subsection{Revision Programs and the Approach of Leite and Pereira}
\label{sec:relations-RevProg}

In the framework of Marek and Truszczy\'nski~\shortcite{mare-trus-94},
a knowledge base is a collection of positive facts, and revision
programs specify conditional insertions or removals of facts under a
semantics very similar to the stable semantics. In discussing this
approach, Leite and Pereira~\shortcite{leit-pere-98} argued that the
approach of revision programs is not adequate if more complex
knowledge is represented in the form of logic programs, because
revision programs compute only ``model-by-model updates'', which do
not capture the additional information encoded by logic
programs. Accordingly, they proposed an extended framework in which a
suitable inertia principle for rules realizes the update independently
of any specific model of the original program.  In the following, we
briefly sketch their approach.

In a first step, Leite and Pereira~\shortcite{leit-pere-98} define their approach for normal logic programs, and afterwards they extend it to handle programs with strong negation as well. Furthermore, \cite{leit-pere-98} deals only with the situation where a given program is updated by a single program; the general case involving an arbitrary number of updates is described in \cite{leite-97}. 
We describe here the latter approach, but, for the sake of simplicity, only the case of normal logic programs.

Following the method of revision programs~\cite{mare-trus-94}, an update program in the sense of \cite{leit-pere-98} is a finite collection of rules of the form
\begin{eqnarray}
\emph{in}(A) & \la & \emph{in}(B_1)\commadots\emph{in}(B_m),\emph{out}(C_1)\commadots\emph{out}(C_n), \quad\mbox{ and}\label{eq:revision:1} \\
\emph{out}(A') & \la & \emph{in}(B_1')\commadots\emph{in}(B_m'),\emph{out}(C_1')\commadots\emph{out}(C_n'),
\label{eq:revision:2}
\end{eqnarray}
where $A,A', B_i, B_i', C_j, C_j'$ are atoms ($1\leq i\leq m, \ 1\leq j\leq n$).
Intuitively, Rule~(\ref{eq:revision:1}) states that $A$ should be true given that $B_1\commadots B_m$ are true and $C_1\commadots C_n$ are false, and a similar meaning holds for Rule~(\ref{eq:revision:2}). Rule~(\ref{eq:revision:1}) is called an \emph{in-rule}, and Rule~(\ref{eq:revision:2}) is an \emph{out-rule}. Semantically, in-rule (\ref{eq:revision:1}) is interpreted as the logic program rule
\begin{eqnarray*}
A & \la & B_1\commadots B_m, \naf C_1\commadots \naf C_n,
\end{eqnarray*}
whilst out-rule (\ref{eq:revision:2}) is interpreted as 
\begin{eqnarray*}
\neg A & \la & B_1\commadots B_m, \naf C_1\commadots \naf C_n.
\end{eqnarray*}

When speaking about update programs, in the following they are always identified with finite sets of rules of the above form. Let us call a sequence $\useq{P}=(P_1\commadots P_n)$ of such programs an \emph{IO-sequence} (for ``sequence of in- and out-rules'').
Consider an IO-sequence $\useq{P}=(P_1\commadots P_n)$ over $\at$, and let $S\subseteq \Lit_\at$ be a set of literals.
Leite~\shortcite{leite-97} introduces the following notion of a rejection set (for $1\leq i,j\leq n$):
$$
\begin{array}{rcl@{~}l}
\mathit{Rejected}(S,i,j) \!\! &=&  \!\! \bigcup_{i<k\leq j} \{ \ r \in P_i \mid & \exists r' \in
P_k \mbox{ such that $r$ and $r'$ are conflicting} \\
            &&& \mbox{and }S\models \body{r} \cup \body{r'} \ \}.
\end{array}
$$
Then, $S\cap\at$ is a \emph{$\useq{P}$-justified update at state j} ($1\leq j\leq n$) iff $S$ is an answer set of 
\[
\bigcup_{i\leq j} (P_i\setminus \mathit{Rejected}(S,i,j)),
\]
provided that each program $\bigcup_{i\leq l} (P_i\setminus
\mathit{Rejected}(S,i,j))$, for $l<j$, possesses an answer
set.\footnote{Strictly speaking, Leite~\shortcite{leite-97} requires
that, for each $l\leq j$, $\bigcup_{i\leq l} (P_i\setminus
\mathit{Rejected}(S,i,j))$ has a consistent answer set. However, in
our setting, answer sets are always consistent.}

It is easily seen that for $\useq{P}=(P_1\commadots P_n)$ and $S$ as
above, $S$ is an answer set of $\bigcup_{i\leq n} (P_i\setminus
\mathit{Rejected}(S,i,j))$ iff it is an answer set of
$\cup\useq{P}\setminus\rs'(S,\useq{P})$, where $\rs'(S,\useq{P})$ is
the weak form of a rejection set, as defined in
Section~\ref{sec:props:char}. Hence, we can state the following
proposition:

\begin{theorem}
Let $\useq{P}=(P_1\commadots P_n)$ be an IO-sequence over $\at$ and $S\subseteq\Lit_\at$ a set of literals. Assume that each subsequence $(P_1\commadots P_j)$ has an answer set, for $1\leq j\leq n$.
Then, $S$ is an answer set of $\useq{P}$ iff it is a $\useq{P}$-justified update at state $n$.
\end{theorem}

Concerning the extended framework of Leite and Pereira in which rules of the form (\ref{eq:revision:1}) and (\ref{eq:revision:2}) may contain literals instead of plain atoms, only a weaker correspondence result holds. Omitting further details, we just mention that in this case our framework corresponds to Leite and Pereira's providing update sequences contain only in-rules.

\subsection{Dynamic Logic Programming}\label{sec:relations-dynLP}

Alferes \emph{et al.}~\shortcite{alfe-etal-98,alfe-etal-99a} introduced the concept of \emph{dynamic logic programs} as a generalization of both
the idea of updating interpretations through revision
programs \cite{mare-trus-94} and of updating programs as defined by Alferes and Pereira~\shortcite{alfe-pere-97} and by Leite and Pereira~\shortcite{leit-pere-98}. Syntactically, dynamic logic programs are based on \emph{generalized logic programs} (GLPs), which allow default negation in the head of rules, but no strong negation whatsoever.

In dynamic logic programming (DynLP in the following), the models of a sequence of updates are defined as the stable models of the program resulting from a syntactic rewriting, similar to the transformation used in our approach. This is called a {\em dynamic update}. Elements of the sequence are GLPs.

Regarding the formalisms discussed in the previous subsection, in 
\cite{alfe-etal-99a} it is demonstrated that revision programs and dynamic updates are equivalent, provided that the original knowledge is \emph{extensional}, \iec the initial program contains only rules of the form $A \la \ $ or $\naf A \la \ $.

Our analysis of dynamic updates can be summarized as follows. First, 
basic definitions and  semantical characterizations of dynamic update programs are given. 
Afterwards, the relation between dynamic updates and updates according to Definition~\ref{def:update-program} is investigated. 
Since the two approaches are defined over different languages, the comparison must include suitable translations to take this distinction into account. 
As a matter of fact, Alferes \emph{et al.}~\shortcite{alfe-etal-99a} already discussed how ELPs can be handled within their framework; likewise, we define a similar translation schema such that GLPs can be treated by our update method.

As it turns out, there is a semantic difference between dynamic updates
and updates according to Definition~\ref{def:update-program}. Although any dynamic update  is an update answer set in the sense of Definition~\ref{def:update-program}, the converse does not hold in general.  Intuitively, this can be explained by the fact that dynamic updates are more restrictive as regards to certain circularities in the given  update information. On the other hand, we provide sufficient conditions under which both approaches yield equivalent results. 
These conditions are formulated by means of suitable graph-theoretical concepts and effectively eliminate the possibility of such circular behavior as mentioned above.
 We also briefly discuss that dynamic logic programs do not eliminate all kinds of circularities.

In view of this equivalence over a wide class of logic programs, the
analysis of update principles we presented in
Section~\ref{sec:properties} applies to dynamic logic programs as
well. Furthermore, similar complexity results for dynamic logic programs
can be concluded, based on the constructions in Section~\ref{subs:compl}.

\subsubsection{Semantics of Dynamic Logic Programs}

Given an update sequence $\useq{P}=(P_1, \ldots, P_n)$ of GLPs  over $\at$, let $\at_{dyn}$ be $\at$ extended by new, pairwise distinct atoms $A^-$, $A_i$, $A_i^-$, $A_{P_i}$, $A_{P_i}^-$, and $\reject{A_i}$, for each $A\in\at$ and each $i\in\{1\commadots n\}$.
The dynamic update program $\useq{P}_\oplus=P_1 \oplus \cdots \oplus P_n$ over $\at_{dyn}$ is defined 
 as the GLP 
consisting of the following items:

\begin{enumerate}
        \item[(i)] 
for each $r \in P_i$, $1 \leq i \leq n$, with $\bodyn{r}=\{C_1\commadots C_n\}$:
\begin{eqnarray*}
A_{P_i} & \la & \bodyp{r}, C_1^-, \dots, C_n^- \quad \quad \mbox{if $\head{r}=A$;}
\\
A_{P_i}^- & \la & \bodyp{r}, C_1^-, \dots, C_n^- \quad \quad \mbox{if $\head{r}=\naf A$;}
\end{eqnarray*}

        \item[(ii)] 
for each atom $A$ occurring in $\useq{P}$ and each  $i\in\{1\commadots n\}$:
$$
\begin{array}{rclrcl}
A_i & \la & A_{P_i}; & \reject{A_{i-1}^-} & \la & A_{P_i}; \\[.6ex]
A_i^- & \la & A_{P_i}^-; \hspace{3em} & \reject{A_{i-1}} & \la &  A_{P_i}^-;
\end{array}
$$
\vspace{-1em}\begin{eqnarray*}
A_i^- & \la & A_{i-1}^-, \naf \reject{A_{i-1}^-}; \\
A_i & \la & A_{i-1}, \naf \reject{A_{i-1}}; 
\end{eqnarray*}

        \item[(iii)]\label{ini-state} for each atom $A$ occurring in $\useq{P}$:
$$A_0^- \la \ ; \qquad A \la A_n; \qquad A^- \la A_n^-; \qquad \naf A \la A_n^-.$$
\end{enumerate} 

\begin{figure*}
\begin{center}
\epsfig{file=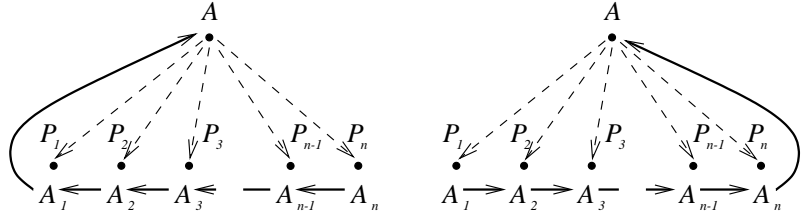,scale=.9}
\end{center}
\caption{``Top-down'' evaluation of update sequences (left diagram) vs.\ ``bottom-up'' evaluation of dynamic logic programs (right diagram).\label{fig:comparison}}
\end{figure*}

One major difference can immediately  be  identified between our update programs and dynamic updates:  
In 
dynamic updates, the value of each atom is determined from the bottom level $P_1$ \emph{upwards} towards $P_n$ (in virtue of  rules $A_i \la A_{i-1}, \naf \reject{A_{i-1}}$ and $A \la A_n$ for positive atoms, and the corresponding ones for dashed atoms), whilst update programs determine such values in a \emph{downward} fashion (cf.\ rules $L_i \la L_{i+1}$ and $L \la L_1$ in Definition~\ref{def:update-program}). This difference is visually depicted in Figure~\ref{fig:comparison}.
More importantly, the different evaluation strategy leads in effect to different semantics, which will be shown later on.

Before we can properly define the semantics of dynamic updates, based on the transformation $\useq{P}_\oplus$ introduced above, we must emphasize that Alferes \emph{et al.}~\shortcite{alfe-etal-98,alfe-etal-99a} use a slightly non-standard concept of stable models. To wit, 
weakly negated literals $\naf A$ ($A$ some atom) are treated like ordinary propositional atoms, so that rules $A_0\la A_1\commadots A_m,\naf A_{m+1}\commadots \naf A_n$ are effectively be viewed as \emph{propositional Horn formulas}. Accordingly, an interpretation $I$ is in this context understood as a 
set consisting of atoms and weakly negated atoms such that for each atom $A$ it holds that $A\in I$ iff $\naf A\notin I$.
To distinguish such interpretations from interpretations in the usual sense, we call them \emph{generalized interpretations}. As usual, a set $B$, comprised of atoms and weakly negated atoms, is true in a generalized interpretation $I$, symbolically $I\models B$, iff $B\subseteq I$.
Towards defining stable models, the following notation is required: 

Let $\at$ be a set of atoms. Then, $\naf \at$ denotes the set $\{\naf A\mid A\in \at\}$. Furthermore, for $M\subseteq\at\cup\naf\at$, we set $M^-=\{\naf A\mid\naf A\in M\}$, and, for $Z\in\at\cup\naf\at$, we define $\naf Z=\naf A$ if $Z=A$ and $\naf Z=A$ if $Z=\naf A$.
For a program $P$ over $\at$, the deductive closure, $\Cni{\at}{P}$, is given by the set 
$$
\{L\mid\mbox{$L\in\at\cup\naf\at$ and $P \vdash L$}\},
$$
where $P$ is interpreted as a propositional Horn theory and ``$\vdash$'' denotes classical derivability. Usually, the subscript ``$\at$'' will be omitted from $\Cni{\at}{P}$. A generalized interpretation $S$ is a stable model of a program $P$ iff $S=\Cn{P\cup S^-}$.

Let $\useq{P}=(P_1\commadots P_n)$ be a sequence of GLPs over $\at$, and let $I$ be a generalized interpretation. 
Alferes \emph{et al.}~\shortcite{alfe-etal-98,alfe-etal-99a} introduce the following concepts:
$$
\begin{array}{rcl@{~}l}
\mathit{Rejected}(I,\useq{P}) \!\! &=&  \!\! \bigcup_{i=1}^n \{ r \in P_i \mid & \exists r' \in
P_j, \textrm{ for some } j\in \{i+1,\ldots, n\}, \textrm{ such }\\
            &&& \textrm{that } \head{r'} = \naf \head{r} \textrm{ and } I\models 
\body{r}\cup\body{r'}\}; \\
\mathit{Defaults}(I,\useq{P}) \!\! &= & \multicolumn{2}{l}{\!\! \{ \naf A \mid \mbox{$\not\!\exists  
r$ in $\useq{P}$ such that $\head{r}=A$ and $I\models\body{r}$}
\}.} 
\end{array}
$$

Then, it holds that $S'\subseteq\at_{dyn}$ is a stable model of $\useq{P}_\oplus$ iff $S=S'\cap\at$ satisfies the following condition:
$$
S=\Cn{(\cup\useq{P} \setminus \mathit{Rejected}(S,\useq{P})) \cup \mathit{Defaults}(S,\useq{P})}.$$
The set $S$ is called a \emph{dynamic stable model} of $\useq{P}$.

Alferes \emph{et al.}~\shortcite{alfe-etal-99a} defined also an extension of their semantics to the
three-valued case: Let
$\useq{P}=(P_1,\ldots,P_n)$ be a sequence of ELPs over $\at$. Then, a consistent set $S
\subseteq \extat$
is a {\em dynamic answer set} of $\useq{P}$ iff $S \cup \{ \naf L \mid L \in \extat\setminus S\}$ is a
dynamic stable model of the sequence  $\useq{P} =
 (P_1,\ldots,P_n\cup \{ \naf A \la \neg A,\, \naf\neg A\la A \mid A
\in \at\})$ of GLPs.  Here, the rules in $\{ \naf A \la \neg A,\, \naf\neg
A\la A \mid A \in \at\}$ serve for emulating classical negation
through weak negation.

\subsubsection{Relating Dynamic Answer Sets and Update Answer Sets}

Let us now define how GLPs are to be rewritten in order to constitute a valid 
input for update programs according to Definition~\ref{def:update-program}. 
For any rule $r$, by $\sn{r}$  we denote  the rule which results from $r$ by
replacing weak negation occurring in the head of $r$ by strong negation, \iec 
$$
\sn{r} = \left\{
\begin{array}{ll}
\neg A \la \body{r} & \textrm{if } \head{r}= \naf A; \\
r & \textrm{otherwise.} 
\end{array}
\right.
$$   
Furthermore, for any GLP $P$, we define $\sn{P} = \{ \sn{r}\mid r\in P
\}$.

\begin{definition}\label{def:glp-simulation}
Let $\useq{P}=(P_1\commadots P_n)$ be a sequence of GLPs  
over $\at$. 
Then, the update sequence $\mathcal{Q}(\useq{P})$ is given by the sequence
$
(\sn{P_1},\ldots,\sn{P_n}\cup \{ \neg A \la
\naf A \mid A \in \at)$. 
\end{definition}

Notice that the part $\{ \neg A \la \naf A \mid A \in \at\}$ serves
for making all answer sets complete. Moreover, no strong
negation occurs in rule bodies of $\cal{Q}(\useq{P})$. Thus,
application of a rule with $\neg A$ in the head can never lead to the
application of further rules; it can only enable that rules with $A$ in
their heads are overridden. As well, the rules in $\{ \neg A \la \naf A
\mid A \in \at\}$ are not able to override any rule in $\Q(\useq{P})$.

\begin{theorem}
\label{lem:dyn-implies-upd}
Let $\useq{P}=(P_1,\ldots,P_n)$ be any sequence of GLPs over a set of atoms
$\at$, and let  $S\subseteq \at \cup \naf \at$ be a dynamic stable model of
$\useq{P}$. Then, $(S \cap \at) \cup \neg(\at\setminus S)$  is an answer set of $\mathcal{Q}(\useq{P})$.
\end{theorem}

\begin{proof}
Given a set $S \subseteq \at \cup \naf \at$ such that 
$$
S\;\; =\;\; \Cn{(\cup\useq{P} \setminus \mathit{Rejected}(S,\useq{P})) \cup \mathit{Defaults}(S,\useq{P})},
$$
 we have to show that $S'=(S \cap \at) \cup \neg (\at \setminus S)$ is an answer set of $\cal{Q}(\useq{P})$, \iec $S'$ is a minimal model of 
$$
\Big(\big(\bigcup_{i=1}^n \sn{P_i}\cup \{ \neg A \la \naf A \mid A \in \at \}\big)\setminus \rs(S',\cal{Q}(\useq{P}))\big)^{S'}.
$$
Let us suppose this is not the case, \iec either the set $S'$ is not a model of the resulting program, or there is a set $S''\subset S'$ such that $S''$ is a model too.

If $S'$ is not a model of 
$$
\Big(\big(\bigcup_{i=1}^n \sn{P_i}\cup \{ \neg A \la \naf A \mid A \in \at \}\big)\setminus \rs(S',\mathcal{Q}(\useq{P}))\Big)^{S'},
$$
 then there is a rule $r\in (\bigcup_{i=1}^n \sn{P_i}\cup \{ \neg A \la \naf A \mid A \in \at \})\setminus \rs(S',\cal{Q}(\useq{P}))$ such that $\body{r}$ is satisfied by $S'$ but $\head{r} \not \in S'$. 
It follows from the construction of $S'$ that such a rule $r$ cannot be of the form $\neg A \la \naf A$, therefore $r \in (\bigcup_{i=1}^n \sn{P_i})
\setminus \rs(S',\cal{Q}(\useq{P}))$ must hold. Let us call $\tilde{r}$ the rule in $\useq{P}$ corresponding to $r$. First of all, invoking Theorem~\ref{theo:sequence-char-0-alt}, we can observe that 
$$
\{ {\tilde{r}}\mid r \in \big(\bigcup_{i=1}^n \sn{P_i}\big)
\setminus \rs(S',\mathcal{Q}(\useq{P})\} \subseteq ((\cup\useq{P}\setminus \mathit{Rejected}(S,\useq{P}))\cup \mathit{Defaults}(S,\useq{P})).
$$
Since the body of $r$ is true in $S'$, we have $S \models \body{\tilde r}$ too, because of the construction of $S'$ and given that $\body{r}=\body{\tilde r}$. Now, if $\head{r}$ is a positive literal,  $\head{r} \not \in S'$ entails $\head{r} \not \in S$,
contradicting the assumption that $S$ is a dynamic stable model of $\useq{P}$. If, on the other hand, $\head{r}$ is a weakly negated atom $\naf A$, from the construction of $S'$ we can deduct that $\naf A$ is not in $S$, contradicting the hypothesis as well.  

Assume now that $S''\subset S'$ is a model of 
$$
\Big(\big(\bigcup_{i=1}^n \sn{P_i}\cup \{ \neg A \la \naf A \mid A \in \at \}\big)\setminus \rs(S',\mathcal{Q}(\useq{P}))\Big)^{S'}.
$$ 
Suppose $S''$ and $S'$ differ on the literal $L$, \iec $L\in S'$ but $L\notin S''$. But this would mean, by construction of $S'$, that a set $\tilde S \subset S$ exists which is also a dynamic stable model of $P$, thus contradicting the hypothesis. Therefore, $S'$ is an answer set of $((\bigcup_{i=1}^n \sn{P_i}\cup \{ \neg A \la \naf A \mid A \in \at \})\setminus \rs(S',\mathcal{Q}(\useq{P})))^{S'}$. 
 \end{proof}

\begin{theorem}
\label{theo:dyn-implies-upd}
Let $\useq{P}=(P_1,\ldots,P_n)$ be a sequence of ELPs over $\at$. Suppose $S\subseteq \extat$
is a dynamic answer set of
$\useq{P}$. Then, $S \in \SMup(\useq{P})$.
\end{theorem}

\begin{proof}
Let us denote classical negation in $\useq{P}$ by $\nneg A$, and rewrite
$S$ accordingly, \iec $S' = (S \cap \at) \cup (\{\nneg A \mid \neg A \in S \})$. Then, by
combining the emulation of classical negation in $\useq{P}$ through
rules $\naf A \la \; \nneg A$ and $\naf\!\nneg A\la A$ ($A \in \at$), and the
transformation $\mathcal{Q}(\cdot)$, we obtain by Theorem~\ref{lem:dyn-implies-upd} that the set
$$
S''\;\;  = \;\;  S' \cup \neg (\{ A, \nneg A \mid A \in \at \} \setminus S')
$$
is an answer set of $\useq{P}'=(P_1,\ldots,P_n\cup
Q)$, where each $\nneg A$ is viewed as a propositional atom and $Q$ contains for each
atom $A \in \at$ the following rules:
\begin{eqnarray*}
\neg \nneg A &\la& A;\\
\neg A &\la& \nneg A; \\
\neg \nneg A &\la& \naf \nneg A;\\
\neg A  &\la&  \naf A.
\end{eqnarray*}
Observe that $\nneg A \in S''$ implies $A \notin
S'$, and $A \in S''$ implies $\neg A \notin S''$. Furthermore, atoms $\nneg A$ or $A$ are included
in $S''$ due to applications of rules $r\in P_i$, $1,\leq i\leq n$, which are not
rejected. 

By  induction on $i$ ($0\leq i<n$) one can show 
that $Rej_{n-i}(S'',\useq{P}')= Rej_{n-i}(S,\useq{P})$ holds. It follows that $S$ is
a minimal model of $((\cup\useq{P})\setminus \rs(S,\useq{P}))^S$, \iec
$S$ is an answer set of $\useq{P}$.
\end{proof}

Theorems~\ref{lem:dyn-implies-upd} and \ref{theo:dyn-implies-upd} do not hold in the converse direction. This can be seen by the following example:
\begin{example}
\label{example:not_if_not}
Consider programs $P_1$, $P_2$, and $P_2'$, where
\begin{eqnarray*}
P_1 & = & \{ \ \emph{it\_is\_raining} \la \ \}; \\
P_2 & = & \{ \ \naf \emph{it\_is\_raining} \la \naf \emph{it\_is\_raining} \ \} ;\\
P_2' & = & \{ \ \neg \emph{it\_is\_raining} \la \naf \emph{it\_is\_raining} \ \}.
\end{eqnarray*}
The sequence $\useq{P}=(P_1,P_2)$ of GLPs has one dynamic stable model, $\{\emph{it\_is\_raining}\}$, but 
$\mathcal{Q}(\useq{P})$ has two answer sets, $\{\emph{it\_is\_raining}\}$ and $\{\neg \emph{it\_is\_raining}\}$. Likewise, the sequence $\useq{P}'=(P_1,P_2')$ of ELPs has also $\{\emph{it\_is\_raining}\}$ as single dynamic stable model, but  $\{\emph{it\_is\_raining}\}$ and $\{\neg \emph{it\_is\_raining}\}$ are answer sets of~$\useq{P}'$.
\end{example}

Intuitively, the syntactic mechanism responsible for the elimination of some stable models in dynamic updates is the renaming of weakly negated literals in the body of rules. This renaming ensures that weakly negated literals are not derived in a cyclic way, \iec the truth value of a weakly negated literal has to be supported by other information besides the literal itself. This, however, is in general not the case with the transformation for update programs based on Definition~\ref{def:update-program}. In the next section we present conditions under which both approaches yield equivalent results. As well, we illustrate that the approach of Alferes \emph{et al.}~\shortcite{alfe-etal-98,alfe-etal-99a} does not eliminate all kinds of cyclic informations.

\subsubsection{Equivalence of Dynamic Stable Models and Update Answer Sets}

In the sequel, we use AND/OR-graphs to identify classes of logic programs for which update answer sets and dynamic stable models 
coincide. Moreover, we present a graph condition which holds precisely in case an update answer set does not correspond to a 
dynamic stable model.
AND/OR-graphs are 
frequently employed to model problem reductions, 
and belong to the category of \emph{hypergraphs}, \iec graphs where nodes are connected by \emph{hypernodes} (also called \emph{connectors}),  represented as a tuple of nodes. Formally, 
an AND/OR-graph is a pair $G=\langle N,C\rangle$, where $N$ is a finite set of AND-nodes or 
OR-nodes,  and $C\subseteq N \times \bigcup^{|N|}_{i=0} N^i$ is a set of connectors  such that $C$ is a function, \iec for each $I\in N$ there is at most one tuple $\langle O_1\commadots O_k\rangle\in \bigcup^{|N|}_{i=0} N^i$ such that $\langle I,O_1\commadots O_k\rangle\in C$.
We call $I$ the \emph{input node} and $O_1\commadots O_k$ the \emph{output nodes} of $\langle I,O_1\commadots O_k\rangle\in C$. As well, $\langle I,O_1\commadots O_k\rangle$ itself is referred to as a \emph{$k$-connector}.

The concept of a path, as defined in ordinary graphs, can be generalized to AND/OR-graphs 
as follows. Let $\langle R, O_1\commadots O_k\rangle$, $k \geq 0$, be the connector for 
a node $R$ in $G$. A tree $p$ is a {\em path rooted at $R$ in $G$} if 
the following conditions hold:
\begin{itemize}
\item[(i)] if $k=0$, then $p=\langle R\rangle$;
\item[(ii)] if $k>0$ and $R$ is an AND-node, then 
$p=\langle R, p_1 \commadots p_k\rangle$, where $p_1 \commadots p_k$ 
are paths rooted at $O_1 \commadots O_k$ in $G$; and

\item[(iii)]  if $k>0$ and $R$ is an OR-node, then $p=\langle R, p_i\rangle$, 
for some $1 \leq i \leq k$,  where $p_i$ is a path rooted at $O_i$ in $G$.
\end{itemize}

 Note that $p$ might 
be an infinite tree. Furthermore, the {\em graph associated with a path $p$}, 
$G(p)$, is the directed graph whose nodes are the nodes of $G$ occurring 
in $p$ and whose edges contain,  for every node $R$ in the recursive 
definition of $p$,  $R\rightarrow O_1\commadots R\rightarrow O_k$ 
if $R$ is an AND-node, and $R\rightarrow O_i$ if $R$ is an OR-node.

Next, we will define how  an AND/OR-graph can be assigned to an update sequence 
$\useq{P}=(P_1 \commadots P_n)$, and how such a graph can be reduced with 
respect to a set of literals $S$.

\begin{definition} 
\label{def:upd-graph} 
Let $\useq{P}=(P_1 \commadots P_n)$ be a sequence of GLPs over $\at$. We associate with 
$\useq{P}$ an AND/OR-graph, $G_{\useqs{P}}=\langle N,C\rangle$, in the following way:
\begin{itemize}
\item[(i)] the set  
$N$ consists of AND-nodes $r$ for every rule $r$ in $\useq{P}$, 
and OR-nodes  $Z$  for every 
$Z\in\head{r'}\cup\body{r'}$ and every $r'$ in $\useq{P}$; 
\item[(ii)] the set  $C$ 
consists of $(k+l)$-connectors 
$\langle r, A_1'\commadots A_k', \naf {A_1'}\commadots \naf {A_l'}\rangle$ for every rule $r$ in $\useq{P}$ of the form 
$H(r) \la A_1\commadots A_k, \naf A_{1}'\commadots \naf A_l'$, $k,l \geq 0$, 
and of $m$-connectors $\langle Z,r_1\commadots ,r_m\rangle$ for all rules $r_1\commadots ,r_m$ in $\useq{P}$ such that $H(r_i)=Z$, where $Z\in\at\cup\naf\at$.
\end{itemize}
\end{definition}

\begin{definition} 
\label{def:red-graph}
Let $\useq{P}=(P_1 \commadots P_n)$ be a sequence of GLPs over $\at$ and let $S\subseteq \at\cup \naf\at$ be a generalized interpretation. The \emph{reduced AND/OR-graph of $\useq{P}$ with respect to $S$}, 
$G^S_{\useqs{P}}$, is the graph resulting from $G_{\useqs{P}}$ by:
\begin{enumerate}
\item[(1)] \label{red-graph:one} removing all AND-nodes $r$ and their connectors 
(as well as removing $r$ from all connectors containing it as output node), 
if  
either $r \in \mathit{Rejected}(S, \useq{P})$ or 
$S \not \models B(r)$ holds; and
\item[(2)]  replacing, for every atom $A$, the connector of $\naf {A}$ by the 
$0$-connector $\langle \naf {A}\rangle$, if $A$ is associated with a $0$-connector after 
Step~(1) and no $r \in \mathit{Rejected}(S, \useq{P})$ exists such that 
$\head{r}=A$.
\end{enumerate}
\end{definition}

Using the definitions above, we are able to state a simple graph 
condition expressing correspondence between update answer sets and dynamic stable 
models. 

\begin{theorem} \label {prop:graph-cond}
Let $\useq{P}=(P_1 \commadots P_n)$ be a sequence of GLPs over $\at$ and let $S\subseteq\extat$ be an 
answer set of $\mathcal{Q}(\useq{P})$. Then, the corresponding generalized interpretation 
$S^{\prime}=\{A\ |\ A \in S\} \cup \{\prenot{A}\ |\ \neg A \in S\}$ 
is a dynamic stable model of $\useq{P}$ iff, for every atom $A\in \at\setminus S$, one of the 
following conditions holds:
\begin{itemize}
\item[(i)] For every rule $r$ in $\useq{P}$ such that $\head{r} = A$, 
$S' \not \models \body{r}$.
\item[(ii)] There exists a path $p$ in $G^{S'}_{\useqs{P}}$, rooted at 
$\prenot{A}$, such that $G(p)$ is acyclic.
\end{itemize}
\end{theorem}

\begin{proof} See \cite{eite-etal-00g}. 
\end{proof}

The next theorem addresses the case of ELPs.

\begin{theorem} \label {theo:elp-graph-cond}
Let $\useq{P}=(P_1 \commadots P_n)$ be a sequence of ELPs over $\at$ and let $S\subseteq\extat$ be an answer 
set of $\useq{P}$. Furthermore, let $\useq{P}^{\prime} = (P_1 \commadots P_n \cup 
\{\naf A \la \neg A, \ \naf \neg A \la A \ |\ A \in \at\})$ and $S' = S \cup \{\naf{L} \mid L \in \extat\setminus S\}$. Then, $S$ is a 
dynamic answer set of $\useq{P}$ iff, for every literal 
$L \in\extat\setminus S$, one of the following conditions holds:
\begin{itemize}
\item[(i)] For every rule $r$ in $\useq{P}'$ such that $H(r) = L$, 
$S' \not \models \body{r}$.
\item[(ii)] There exists a path $p$ in $G^{S'}_{\useqs{P}^{\prime}}$ rooted at 
$\prenot{L}$, such that $G(p)$ is acyclic.
\end{itemize}
\end{theorem}

\begin{proof} See \cite{eite-etal-00g}. 
\end{proof}

Theorem~\ref{prop:graph-cond}
and 
Theorem  \ref{theo:elp-graph-cond} allow us to identify classes of update 
programs for which update answer sets and dynamic stable models coincide. 
If, for example, a cyclic path rooted at $\naf{A}$ depends on $A$, then a 
reduction of the graph with respect to an answer set will break it open since an answer 
set either contains $A$, or not. 
Also, if the graph associated with an update sequence does not contain 
any cyclic paths rooted at weakly negated literals, then the second condition 
of Theorem~\ref{prop:graph-cond} is always fulfilled.

\begin{corollary}
Let $\useq{P}=(P_1 \commadots P_n)$ be a sequence of GLPs over $\at$. If every cyclic path
rooted at $\naf{A}$ in $G_{\useqs{P}}$ also contains node $A$, then 
there is a one-to-one correspondence between answer sets of $\mathcal{Q}(\useq{P})$ and dynamic stable models of 
$\useq{P}$.
In particular, this holds if $G_{\useqs{P}}$ 
does not contain any cyclic path rooted at a $\naf{A}$ node.
\end{corollary}

This corollary can  further be specialized to a condition which can be checked 
efficiently (namely in 
$\mathcal{O}(|\at| \cdot \|\useq{P}\|)$ time, where $\|\useq{P}\|$ denotes the size of $\useq{P}$).

\begin{corollary} \label {coll:eff-graph-cond}
Let $\useq{P}=(P_1\commadots\!P_n)$ be a sequence of GLPs over $\at$ and let 
$\mathcal{G}=\bigcup_{
p} G(p)$ be the union of the graphs for 
all paths $p$ in $G_{\useqs{P}}$. Suppose that, for every node $\naf{\!A}$ in 
$\mathcal{G}$, it holds that every cycle containing $\naf{A}$ also 
contains $A$. Then, there is a one-to-one correspondence between  answer sets of $\mathcal{Q}(\useq{P})$ and dynamic 
stable models of $\useq{P}$.
\end{corollary}

\begin{example}
\label{example:tv_glp}
Consider the following sequence $\useq{P} = (P_1, P_2, P_3)$ of GLPs, taken 
from~\cite{alfe-etal-98}:
\begin{eqnarray*}
P_1 & = & \big\{ \ r_1: \ \emph{sleep} \la \naf \emph{tv\_on}, \quad r_2: \ \emph{tv\_on} \la \ ,
\quad r_3: \ \emph{watch\_tv} \la \emph{tv\_on} \ \big\};\\
P_2 & = & \big\{ \ r_4: \ \naf \emph{tv\_on} \la \emph{power\_failure}, \quad r_5: \ \emph{power\_failure} \la \ \big\};\\
P_3 & = & \big\{ \ r_6: \ \naf \emph{power\_failure} \la \ \big\}.
\end{eqnarray*}

There are only two $\naf{A}$ nodes in $\mathcal{G}$ with outgoing edges, 
namely $\naf{\emph{tv\_on}}$ and $\naf{\emph{power\_failure}}$. Both are 
connected with a single rule node: the former with $r_4$, and the latter with 
$r_6$, which is a terminal node. Node $r_4$ has a single edge leading to 
$\emph{power\_failure}$, which is in turn connected with a single rule node, 
namely $r_5$, a terminal node.
Thus, there does not exist a cycle in $\mathcal{G}$ containing $\naf{A}$. Hence, update answer sets of $\mathcal{Q}(\useq{P})$ and dynamic 
stable models of $\useq{P}$ coincide (modulo the different language). In fact, $\useq{P}$ has the single 
dynamic stable model 
\[
\{\prenot{\emph{power\_failure}}, \emph{tv\_on}, \emph{watch\_tv}, 
\prenot{\emph{sleep}}\},
\] 
which corresponds to the single update answer set of $\mathcal{Q}(\useq{P})$, given by:
\[
\{\neg\emph{power\_failure}, \emph{tv\_on}, 
\emph{watch\_tv}, \neg\emph{sleep}\}.
\]
\end{example}

Having dealt with aspects of equivalence between dynamic stable models and update 
answer sets, let us now discuss their differences. 
Recalling the update sequence $\useq{P}=(P_1,P_2)$ and $\useq{P}'=(P_1,P_2')$ from Example~\ref{example:not_if_not}, the single dynamic stable model $\{\emph{it\_is\_raining}\}$ of $\useq{P}$ seems, in the sense of inertia, more intuitive than the answer set $\{\neg \emph{it\_is\_raining}\}$ of $\useq{P}'=\mathcal{Q}(\useq{P})$ because the tautological update information 
$$
P_3 \;\; = \;\; \{\naf \emph{it\_is\_raining} \la \naf \emph{it\_is\_raining}\}
$$
 is quite irrelevant to the fact that it is raining, as given by 
$$
P_1\;\; = \; \; \{ \emph{it\_is\_raining} \la \}.
$$  
So, in some sense, the semantics of Alferes \emph{et al.}~\shortcite{alfe-etal-98,alfe-etal-99a} eliminates unintended stable models, as it does not allow for cyclic derivations of negative information.   
However, the rewritten rule  of 
$$
P_2' \;\; = \;\; \{\neg \emph{it\_is\_raining} \la \naf \emph{it\_is\_raining}\}
$$
 differs in that it allows to conclude that 
it is not raining given that there is no information whether it is raining. In this 
sense, both answer sets $\{\emph{it\_is\_raining}\}$ and $\{\neg \emph{it\_is\_raining}\}$ are, in principle, reasonable. Observing that the 
more intuitive answer set of $\useq{P}'$ is minimal while the other is not, 
one can use the notion of minimality to filter out the unintended answer set. 
But, in general, there exist dynamic stable models such that the corresponding 
answer sets are not minimal (or strictly minimal) and vice versa. Also, acyclic 
derivations of negative information do not always capture the intuition of 
inertia as shown by the following example:
\begin{example}
\label{example:nota_if_notb}
Let us consider a slight modification of Example \ref{example:not_if_not}, where 
the knowledge base 
$$
P \;\; = \;\; \{ \ \emph{it\_is\_raining} \la \ ,\  \emph{it\_is\_cloudy} \la \emph{it\_is\_raining} \ \}
$$ 
is updated by the information 
$$
U \;\; = \;\; \{ \naf \emph{it\_is\_raining} \la \naf \emph{it\_is\_cloudy} \},
$$
 which, by the same 
intuition of inertia, is also irrelevant to the fact that it is now actually 
raining and thus cloudy. However, this yields two dynamic stable models
\begin{eqnarray*} 
S_1^{\prime} & = & \{ \emph{it\_is\_raining}, \emph{it\_is\_cloudy} \}; \\ 
S_2^{\prime} & = & \{ \naf \emph{it\_is\_raining}, \naf \emph{it\_is\_cloudy} \},
\end{eqnarray*} 
corresponding to the answer sets 
\begin{eqnarray*}
S_1 & = & S_1^{\prime}; \\
S_2 & = & \{ \neg \emph{it\_is\_raining}, \neg \emph{it\_is\_cloudy} \}
\end{eqnarray*}
 of the rewritten ELP\footnote{Similar to Example~\ref{example:not_if_not}, here, the intuitively preferred answer set is also 
a minimal answer set, while the other is not.}, showing that also the mechanisms enacted in DynLP do not completely avoid cyclic derivations.
\end{example}

Despite their differences, the general properties of program updates, as investigated 
in Section~\ref{sec:properties}, hold for dynamic logic programs also. By virtue of Theorem~\ref{theo:elp-graph-cond} and Corollary~\ref{coll:eff-graph-cond}, one can easily 
verify that every counterexample for an invalid property belongs to a class 
where update answer sets and dynamic stable models coincide. As well, 
arguments similar to those used for the demonstrations of the valid properties of Section~\ref{sec:properties} can be found in 
order to show that these properties also hold for dynamic logic programs.

\subsection{Program Updates Through Abduction}

The use of abduction for solving update problems in logic programming
and data\-bases goes back to
\cite{kaka-manc-90}.  Taking advantage of their framework of {\it extended
abduction} \cite{inou-saka-95b}, Inoue and Sakama \shortcite{inou-saka-99} integrated
three different types of updates into a single framework, namely \emph{view update}, \emph{theory update}, and \emph{inconsistency removal}. In
particular, view update deals with the problem of changing \emph{extensional
facts} (which do not occur in the heads of rules), whilst theory update
covers the general case in which (a set of) rules should be
incorporated into a knowledge base. We discuss the latter problem here. 

Informally, for ELPs $P_1$ and $P_2$, an
update of $P_1$ by $P_2$ is a largest program $P'$ such that $P_1
\subseteq P' \subseteq P_1\cup P_2$ holds and where $P'$ is consistent (\iec $P'$ has a consistent answer set). This intuition is formally
captured 
 by reducing the problem of updating $P_1$ with
$P_2$ to computing a minimal set of abducible rules $Q \subseteq
P_1\setminus P_2$ such that $(P_1\cup P_2)\setminus Q$ is consistent.
In technical terms of \cite{inou-saka-95b}, the program $P_1
\cup P_2$ is considered for abduction where the rules in $P_1
\setminus P_2$ are abducible, and the intended update is realized via
a minimal {\em anti-explanation} for falsity, which removes abducible
rules to restore consistency.

While this looks similar to our minimal updates, there is, however, a
salient difference: abductive update does not
respect {\em causal rejection}. That is, a rule $r$
from $P_1\setminus P_2$ may be rejected even if no rule
$r'$ in $P_2$ fires whose head contradicts the application of $r$. For
example, consider $P_1 = \{ q \la\, ,\neg q \la a \}$
and $P_2 = \{ a\la \, \}$. Both $P_1$ and $P_2$ have consistent
answer sets, but $(P_1,P_2)$ has no (consistent) answer set because no rule in $P_1$ is rejected and thus both rules must fire.
On the other hand, in Inoue and Sakama's approach, one of the two rules in $P_1$ will be
removed. Furthermore, inconsistency removal in a program $P$ occurs in this framework as
special case of updating (take, \egc $P_1=P$ and $P_2=\emptyset$). 

From a computational point of view, abductive updates are---due to
 inherent minimality criteria---harder than update programs; in
 particular, some abductive reasoning problems are shown to be
 $\SigmaP{2}$-complete \cite{inou-saka-99}.

\subsection{Updates Through Priorities}

Zhang and Foo~\shortcite{foo-zhan-98} described an approach for updating logic
programs based on their preference-handling framework for logic programs introduced in~\cite{foo-zhan-97a}. 
The general approach is rather involved and
proceeds in two stages, roughly described as follows. For updating $P_1$ with
$P_2$, in the first stage, each answer set $S$ of $P_1$ is updated to a ``closest''
answer set $S'$ of $P_2$, where distance is measured in terms of the
set of atoms for which $S$ and $S'$ have different truth values, and
closeness is set inclusion. Then, a maximal set of rules
$Q\subseteq P_1$ is chosen in such a way that $P_3= P_2\cup Q$ has an answer
set containing $S'$. In the second stage, $P_3$ is viewed as a prioritized logic program  in which rules from $P_2$ have
higher priority than rules from $Q$, and its answer sets are computed.
The resulting answer sets are identified as the answer sets of the 
update of $P_1$ with $P_2$.

This approach is apparently different from our update framework.
 In fact, it is in the spirit of Winslett's~\shortcite{wins-88} \emph{possible models approach}, where the models of a propositional theory are
updated separately and which satisfies update postulate (U8).
More specifically, the two stages in Zhang and Foo's approach respectively aim at removing contradictory rules from $P_1$ and resolving
conflicts between the remaining rules of $P_2$.
However, like in Inoue and Sakama's approach, rules are not removed on
the basis of causal rejection. In particular, on the example considered
in~\cite{foo-zhan-98}, both approaches yield the same result. The second stage of the
procedure indicates a strong update flavor of the approach, since
rules are unnecessarily abandoned.  For example, the update of $P_1=\{ p
\la \naf q\}$ with $P_2 = \{ q \la \naf p\}$ results in $P_2$, even
though $P_1 \cup P_2$ is consistent. Since, in general, the result of
an update is given by a set of programs, naive handling of sequences
of updates consumes exponential space in general.

\subsection{Compiled Preferences}

Since the underlying conflict-resolution strategy of many update
formalisms, including the current one, is to associate, in some sense,
``higher preference'' to new pieces of information, as final
installment of our discussion on related work, we briefly review the
approach of Delgrande \emph{et al.} \shortcite{delg-etal-00} to
preference handling in logic programming, which is also based on a
transformational principle.

To begin with, Delgrande \emph{et al.} \shortcite{delg-etal-00}
define an \emph{ordered logic program} as an ELP in
which rules are named by unique terms and in which preferences among
rules are given by a new set of atoms of the form $s \prec t$, where
$s$ and $t$ are names.  Thus, preferences among rules are encoded at
the {\em object-level}.  An ordered logic program is transformed into
a second, regular ELP wherein the preferences are
respected, in the sense that the answer sets obtained in the
transformed theory correspond to the preferred answer sets of the
original theory.  The approach is sufficiently general to allow the
specification of preferences among preferences, preferences holding in
a particular context, and preferences holding by default.

The encoding of ordered logic programs into standard ELPs is realized by means of dedicated atoms, which control the applicability of rules with respect to the intended order. More specifically, if rule $r$ has preference over rule $r'$, the control elements ensure that $r$ is
considered before $r'$, in the sense that, for a given answer set $S$, rule
$r$ is known to be applied or defeated \emph{ahead of} $r'$. 

This control mechanism is more strict than the rejection principle realized in Definition~\ref{def:update-program}. For instance, in the preference approach, it may happen that no answer set exists because the applicability of a higher-ranked rule depends on the applicability of a lower-ranked rule, effectively resulting in a circular situation which cannot be resolved in a consistent manner. On the other hand, this is not necessarily the case in the current update framework, where newer rules may only be applicable given older pieces of information. So, in order to simulate updates within the framework of  \cite{delg-etal-00}, under the proviso that newer information has preference over older information, 
it is necessary to relax the 
conditions which enable successive rule applications.

\section{Conclusion}
\label{sec:conclusion}

In this paper, we considered 
a formalization of 
an approach to
sequences of logic program updates based on a causal rejection principle for rules, which
is inherent to other approaches as well. We 
provided, in the
spirit of dynamic logic programming~\cite{alfe-etal-98,alfe-etal-99a},
a definition of the semantics of sequences $\useq{P}$ of ELPs in terms of a simple transformation to update programs,
$\useq{P}_\upd$, which are ordinary ELPs, and described a declarative semantical characterization as well. Then, 
as a main novel contribution, we investigated the properties of this
approach and of similar ones from the perspective of belief revision and nonmonotonic
reasoning, based on the given characterization. For this purpose, we 
considered different
possibilities of interpreting update programs as theory change
operators and abstract nonmonotonic consequence operators,
respectively. Our main findings on this aspect  were that many of the
postulates and principles from these areas are not
satisfied by update programs. We then have introduced further
properties, including an iterativity property, and evaluated them on
update programs.

Motivated by an apparent lack of minimality of change, we 
then considered refinements of the semantics in terms of minimal and
strictly minimal answer sets, and 
discussed their complexity
and implementation. Furthermore, we 
compared the current proposal to
other related approaches, and found that it is
semantically equivalent to a fragment of inheritance
logic programs as defined by  Buccafurri \emph{et al.}~\shortcite{bucc-etal-99a-iclp}. Moreover, our approach
coincides on certain classes with dynamic logic programming, which has been introduced by  Alferes \emph{et al.}~\shortcite{alfe-etal-98,alfe-etal-99a}. For the latter correspondence results, we provided exact
characterizations in terms of graph-theoretical conditions. In view of these results, our discussion on general principles of update sequences based on causal rejection applies for these formalisms as well.

Several issues remain for further work. An interesting point
concerns the formulation of postulates for update operators on
logic programs and, more generally, on nonmonotonic theories. As we
have seen, several postulates from the area of logical theory change
fail for update programs (cf.\ \cite{brew-00} for related
observations on this topic). This may partly be explained by the nonmonotonicity of
answer sets semantics, and by the dominant role of syntax for update
embodied by causal rejection of rules. However, similar features are not
exceptional in the context of logic programming. Therefore, it would be
interesting to consider further postulates and desiderata for updating logic programs besides the ones we analyzed here, as well as an AGM style
characterization of update operators compliant with them. This issue
seems to be rather demanding, though, and we might speculate---without further
evidence---that it will be difficult to find a general acceptable set of
 postulates which go beyond ``obvious'' properties.

A natural issue for update logic programs is the inverse of addition,
\iec retraction of rules from a logic program.  Dynamic logic
programming evolved into LUPS~\cite{alfe-etal-99a}, which is a
language for specifying update behavior in terms of conditional addition and
retraction of sets of rules to a logic program. LUPS is generic,
however, as in principle different approaches to updating logic
programs could provide the underlying semantical basis for the single update
steps. Exploring properties of the general framework, as well as of
particular instantiations, 
and
reasoning about update programs describing the behavior of agents
programmed in LUPS or in other similar languages is 
topic of ongoing research.

Finally, building real-life applications, like intelligent
information agents 
whose rational component is modeled by a knowledge base, which is in turn maintained
using update logic programs, is an interesting issue for further
research. The integration of reasoning components into agent
architectures amenable to logic programming methods, such as the one
of the IMPACT agent platform~\cite{subr-etal-98}, is an important next
step in order to make the techniques available to agent
developers. This is also part of our current research.

\subsubsection*{Acknowledgments.}

The authors would like to thank the anonymous referees and Michael
Gelfond for their comments, which helped improving the paper, as well
as Jos{\`e} Julio Alferes, Jo{\~a}o Alexandre Leite, and Lu{\'\i}s
Moniz Pereira for their useful remarks and corrections. This work was
supported by the Austrian Science Fund (FWF) under grants
P13871-INF and N~Z29-INF.

\appendix
\label{appendix}

\section{Proofs
}\label{app:proofs}

\subsection{Proof of Theorem \ref{prop:1-1-corr}}
\label{app:proof-prop:1-1-corr}

For any set $U\subseteq\Lit_{\at^\ast}$, define $U_0=U\cap\extat$, and, for  $1\leq i\leq n$, let $U_i=\{ L_i \mid L_i\in U\}$ and $U_i^\rejo=\{\rej{r} \mid \rej{r}\in U, r\in P_i\}$. Clearly, it holds that 
\(
U=U_0\cup\bigcup_{i=1}^n(U_i\cup U_i^\rejo).
\) 

Consider the answer sets $S,T$ of $\useq{P}_\upd$ and assume that $S\cap\extat=T\cap\extat$.  
We show by induction on $j$ ($0\leq j\leq n-1$) that $S_{n-j}=T_{n-j}$ and $S_{n-j}^\rejo=T_{n-j}^\rejo$. From this, and given the relation $S_0=T_0$ (by the assumption $S\cap\extat=T\cap\extat$), it follows that $S=T$.

\medskip
\noindent
{\sc Induction Base.}
 Assume $j=0$. First of all, it is quite obvious that $S_{n}^\rejo=T_{n}^\rejo=\emptyset$. Consider now some $L_n\in\Lit_{\at^\ast}$.  According to the construction of the transformation $\useq{P}_\upd$, the literal $L_n$ can only be derived by some rule $L_n\la\body{r}, \naf \rej{r} \in \useq{P}_\upd$, where $r\in P_n$.
Since $S_{n}^\rejo=T_{n}^\rejo=\emptyset$, it follows that 
$L_n\la\bodyp{r}$ must be a member of both $\useq{P}_\upd^S$ and
$\useq{P}_\upd^T$. Since $\bodyp{r}\subseteq\extat$ and $S_0=T_0$,  we have $\bodyp{r}\subseteq S$ iff $\bodyp{r}\subseteq T$. Thus, $L_n\in S_n$ iff $L_n\in T_n$. This implies $S_n=T_n$.

\medskip
\noindent
{\sc Induction Step.} Assume $n-1\geq j>0$, and let the assertions $S_{n-k}=T_{n-k}$ and $S_{n-k}^\rejo=T_{n-k}^\rejo$ hold for all $k<j$. We show that they hold for $k=j$ as well. Consider some atom $\rej{r}$ where  $r\in P_{n-j}$. Given the transformation $\useq{P}_\upd$, the atom $\rej{r}$ can only be derived by means of rule $\rej{r}\la \body{r},\cl{L}_{n-j+1}\in\useq{P}_\upd$. Since $\bodyn{r}\subseteq\extat$ and $S_0=T_0$, it holds that $\bodyn{r}\cap S=\bodyn{r}\cap T$. Hence, $\rej{r}\la \bodyp{r},\cl{L}_{n-j+1}$ is in $\useq{P}_\upd^S$ iff it is in
$\useq{P}_\upd^T$. By induction hypothesis,  $\cl{L}_{n-j+1}\in S$ iff $\cl{L}_{n-j+1}\in T$.
Since we also have that $\bodyp{r}\subseteq S$ iff $\bodyp{r}\subseteq T$, it follows that  $\rej{r}\in S$ iff $\rej{r}\in T$, and so $S_{n-j}^\rejo=T_{n-j}^\rejo$. 

Consider now some literal $L_{n-j}\in \Lit_{\at^\ast}$. This literal can only be derived by means of rule $L_{n-j}\la L_{n-j+1}$, or by a rule of the form $L_{n-j}\la \body{r},\naf \rej{r}$, for some $r\in P_{n-j}$.
If $L_{n-j}$ is derived by $L_{n-j}\la L_{n-j+1}$, it follows immediately from the induction hypothesis that $L_{n-j}\in S$ iff $L_{n-j}\in T$.
So assume now that the second case applies. 
Since we already know that $S_{n-j}^\rejo=T_{n-j}^\rejo$, and since $\bodyn{r}\cap S=\bodyn{r}\cap T$, we have that $L_{n-j}\la \bodyp{r}$ lies in $\useq{P}_\upd^S$ iff it lies in 
$\useq{P}_\upd^T$. Again using the property that  $\bodyp{r}\subseteq S$ iff $\bodyp{r}\subseteq T$, we obtain that  $L_{n-j}\in S$ iff $L_{n-j}\in T$. 
Combining the two cases, and since the literal $L_n$ was arbitrarily chosen, it follows that $S_{n-j}=T_{n-j}$.

\subsection{Proof of Theorem \ref{theo:sequence-char-0}}\label{app:proof-theo:sequence-char-0}

\emph{Only-if part.} 
Suppose $S$ is an answer set of
$\useq{P}=(P_1\commadots P_n)$. We show that $S$ is a minimal model of $(\cup \useq{P}\setminus \rs(S,\useq{P}))^S$.
First, we show that $S$ is a model of $(\cup \useq{P}\setminus \rs(S,\useq{P}))^S$. 

Let $\corr{S}$ be the uniquely determined answer set of $\useq{P}_{\upd}$ such that $S=\corr{S}\cap\extat$.
Consider some $\reductr{r}\in (\cup \useq{P}\setminus \rs(S,\useq{P}))^S$.
We first assume that $r$ is a constraint. So, $r\in \useq{P}_\upd$. 
Since $\bodyn{r}\cap S=\emptyset$, $\bodyn{r}\subseteq\extat$, and $S\subseteq\corr{S}$, we have $\bodyn{r}\cap\corr{S}=\emptyset$. Hence, $\bodyp{r}\not\subseteq\corr{S}$, since $\corr{S}$ is an answer set of $\useq{P}_\upd$ and $\reductr{r}\in(\useq{P}_\upd)^{\corr{S}}$.  It follows that $\reductr{r}$ is true in $S$. Let us now consider the case when $r$ is not a constraint. Then,  
there is some $i$, $1\leq i\leq n$, such that $r\in P_i$  and 
$r\not\in \rs(S,\useq{P})$. 
We must show that $\head{r}\in S$ whenever $\bodyp{r}\subseteq S$.
By construction of the update program $\useq{P}_\upd$, $r$ induces some rule $L_i \la \body{r},\naf \rej{r}\in\useq{P}_\upd$, where $L=\head{r}$.
We claim that $L_i \la \body{r},\naf \rej{r}$ is not defeated by $\corr{S}$.
First of all, since $\bodyn{r}\cap S=\emptyset$, it follows that $\bodyn{r}\cap \corr{S}=\emptyset$, as argued above. Furthermore, since
$r\not\in \rs(S,\useq{P})$, Lemma~\ref{lemma:rejection-atoms} implies $\rej{r}\notin\corr{S}$.
This proves the claim. 
Thus, $L_i\la\bodyp{r}\in(\useq{P}_\upd)^{\corr{S}}$. 
Consequently, assuming $\bodyp{r}\subseteq S$, it holds that $L_i\in\corr{S}$, since $\corr{S}$ is an answer set of $\useq{P}_\upd$ and $S\subseteq\corr{S}$.
Moreover, since $(\useq{P}_\upd)^{\corr{S}}$ contains the inertia rules $L_i\la L_{i+1}$  ($1\leq i<n$) and $L\la L_1$, it follows that $L\in \corr{S}$. By observing that $L\in\extat$,  $L\in S$ follows, which implies that $\reductr{r}$ is true in $S$. This concludes the proof that $S$ is a model of $(\cup \useq{P}\setminus \rs(S,\useq{P}))^S$. It remains to show that $S$ is a minimal model of  $(\cup \useq{P}\setminus \rs(S,\useq{P}))^S$.

Assume that $S_0\subset S$ is a model of  $(\cup \useq{P}\setminus \rs(S,\useq{P}))^S$. Consider the set 
\begin{eqnarray*}
\tilde{S}_0 & = & \corr{S}\setminus(\{L\mid L\in S\setminus S_0\}\cup \{L_i\mid L\in S\setminus S_0, 1\leq i\leq n\}).
\end{eqnarray*}
It is easy to show that $\tilde{S}_0$ is a model of $(\useq{P}_\upd)^{\corr{S}}$. Moreover, $\tilde{S}_0\subset\corr{S}$. We arrive at a contradiction, because $\corr{S}$ is assumed to be an answer set of $\useq{P}_\upd$. As a consequence, $S$ must be a minimal model of $(\cup \useq{P}\setminus \rs(S,\useq{P}))^S$. This concludes the proof that $S$ is a minimal model of $(\cup \useq{P}\setminus \rs(S,\useq{P}))^S$ whenever $S$ is an answer set of~$\useq{P}$.

\smallskip\noindent
\emph{If part.}
Assume that $S$ is a minimal model of $(\cup \useq{P}\setminus \rs(S,\useq{P}))^S$.
Define $\tilde{S}\subseteq\Lit_{\at^\ast}$ as follows:
\begin{eqnarray*}
\tilde{S}\!\! &=& \!\! S \cup \{
rej(r) \mid r \in \rs(S,\useq{P})\} \cup \\
 && \bigcup_{i=1}^n \{ L_j \mid 1\leq j\leq i, \exists r\in P_i
        \setminus \rs(S,\useq{P}) \mbox{ such that $\head{r}=L$ and $S\models\body{r}$}\}.
\end{eqnarray*}
We show that $\tilde{S}$ is an answer set of $\useq{P}_{\upd}$. Since $\tilde{S}\cap\extat=S$, this will imply that $S$ is an answer set of $\useq{P}$.

We first show that $\tilde{S}$ is a model of $(\useq{P}_\upd)^{\tilde{S}}$.
Consider some $\reductr{r}\in (\useq{P}_\upd)^{\tilde{S}}$. Depending on the construction of $\useq{P}_\upd$, there are several cases to distinguish.

(i) $r$ is a constraint. 
Then, $\bodyp{r}\not\subseteq\tilde{S}$. 
Otherwise, we would have $\bodyp{r}\subseteq S$ and $\reductr{r}\in(\cup \useq{P}\setminus \rs(S,\useq{P}))^S$ (since $S\subseteq\tilde{S}$, $\bodyp{r}\subseteq \extat$, and $\bodyn{r}\cap\tilde{S}=\emptyset$), violating the condition that $S$ is a model of $(\cup \useq{P}\setminus \rs(S,\useq{P}))^S$. 
Thus, $\reductr{r}$ is true in $\tilde{S}$.

(ii) $r$ is a rule of form $L_i\la\body{r'},\naf\rej{r'}$, where $L=\head{r'}$.
Since  $\reductr{r}\in (\useq{P}_\upd)^{\tilde{S}}$, $\reductr{r}$ is not defeated by $\tilde{S}$ and $\rej{r'}\not\in \tilde{S}$. According to the definition of $\tilde{S}$, the latter condition implies that $r'\notin\rs(S,\useq{P})$.
Since $\head{r}=L_i$, it holds that $r'\in P_i$, so $r'\in P_i\setminus\rs(S,\useq{P})$. Assume $\bodyp{r'}\subseteq \tilde{S}$.
Since $S\subseteq\tilde{S}$ and $\bodyp{r'}\subseteq\extat$, we get $\bodyp{r'}\subseteq S$. Moreover, since $r$ is not defeated by $\tilde{S}$, the definition of $\tilde{S}$ implies that $L_i\in\tilde{S}$. This shows that $\reductr{r}$ is true in $\tilde{S}$.

(iii) $r$ is a rule of form $\rej{r'}\la \body{r'},\neg L_{i+1}$, where $r'\in P_i$ and $L=\head{r'}$. Assume $\bodyp{r}\subseteq\tilde{S}$. Hence, $\neg L_{i+1}\in\tilde{S}$. By definition of $\tilde{S}$, this implies that there is some rule $r''\in P_j\setminus\rs(S,\useq{P})$, $i+1\leq j\leq n$, such that $\head{r''}=\neg L$, $\bodyp{r''}\subseteq S$, and $r''$ is not defeated by $S$.
Since $\rs_j(S,\useq{P})\subseteq\rs(S,\useq{P})$, it follows immediately that $r'\in\rs_i(S,\useq{P})\subseteq\rs(S,\useq{P})$, which in turn implies $\rej{r'}\in\tilde{S}$, by definition of $\tilde{S}$, proving that $\reductr{r}$ is true in $\tilde{S}$.

(iv) $r$ is a rule of form $L_i\la L_{i+1}$ ($1\leq i<n$). Then $r$ is trivially true in $\tilde{S}$, by construction of $\tilde{S}$.

(v) $r$ is a rule of form $L\la L_1$. If $L_1\in \tilde{S}$, then there is some $r'\in P_1\setminus\rs(S,\useq{P})$ such that $\head{r'}=L$, $\bodyp{r'}\subseteq {S}$, and $r'$ is not defeated by $S$.
Since $S$ is a model of $(\cup\useq{P}\setminus\rs(S,\useq{P}))^S$, it follows that $L\in S\subseteq\tilde{S}$. Thus, $r$ is true in $\tilde{S}$.

This concludes the proof that $\tilde{S}$ is a model of $(\useq{P}_\upd)^{\tilde{S}}$. We proceed by showing that $\tilde{S}$ is a minimal model of $(\useq{P}_\upd)^{\tilde{S}}$.
Suppose $\tilde{S}_0$ is a model of $(\useq{P}_\upd)^{\tilde{S}}$ such that $\tilde{S}_0\subset\tilde{S}$. We show that this implies $\tilde{S}\subseteq\tilde{S}_0$, a contradiction. Hence, $\tilde{S}$ must be minimal.

Let us first assume that $\tilde{S}_0 \cap \extat \subset
\tilde{S}\cap \extat$, i.e., $\tilde{S}_0$ is smaller on the literals in $\extat$. Then, for some $L \in \extat$, no rule $\reductr{r}\in P_{\upd}^{\tilde{S}}$ with $\head{r}=L_i$
fires in $\tilde{S}_0$, i.e., $\bodyp{r}\not\subseteq
\tilde{S}_0$. Hence, by definition of $\tilde{S}$ and $P_{\upd}^{\tilde{S}}$, there is no $r' \in (\cup\useq{P}\setminus \rs(S,\useq{P}))^S$ such that $\head{r'}= L$ and $\bodyp{r'}\subseteq S$.  Consequently, $S\setminus\{L\}$ satisfies all rules in
$(\cup\useq{P}\setminus \rs(S,\useq{P}))^S$. This, however, contradicts the fact that $S$ is a minimal model of $(\cup\useq{P}\setminus
\rs(S,\useq{P}))^S$. It follows that $\tilde{S}_0\cap \extat = \tilde{S}\cap \extat$ holds.

Now consider any $L_i \in \tilde{S}$. Then, there is a rule $r\in
P_j\setminus \rs(S,\useq{P})$, $i\leq j\leq n$, such that $\bodyp{r}\subseteq S$ and $r$ is not defeated by $S$. According to the definition of $\useq{P}_\upd$, and by Lemma~\ref{lemma:rejection-atoms}, rule $r$ introduces a rule
$A_j \la \bodyp{r} \in (\useq{P}_{\upd})^{\tilde{S}}$. Since $\bodyp{r} \subseteq S=\tilde{S}\cap \extat$ and 
$\tilde{S}\cap \extat = \tilde{S}_0\cap \extat$, 
it follows that $A_j \in \tilde{S}_0$, by the assumption that $\tilde{S}_0$ is a model of $(\useq{P}_{\upd})^{\tilde{S}}$. Moreover, since $i\leq j$, the inertia rules $L_k\la L_{k+1}\in (\useq{P}_{\upd})^{\tilde{S}}$ ($1\leq k<n$) imply $L_i\in \tilde{S}_0$.

Finally, consider $\rej{r} \in \tilde{S}$, where $r \in P_i$. By the definitions
of $\rs(S,\useq{P})$ and $\tilde{S}$, it follows that $\bodyp{r}\subseteq S$, $\bodyn{r}\cap S=\emptyset$, and $\neg L_{i+1} \in
\tilde{S}$. From the above considerations, $\neg L_{i+1} \in
\tilde{S}$ implies $\neg L_{i+1} \in
\tilde{S}_0$. Moreover, $\bodyn{r}\cap \tilde{S}=\emptyset$. So, $\rej{r}\la \bodyp{r}, \neg L_{i+1}\in \useq{P}_{\upd}^{\tilde{S}}$. Since $\tilde{S}_0$ is a model of $(\useq{P}_\upd)^{\tilde{S}}$, and given the fact that $\tilde{S}_0\cap\extat=S$, we obtain $\rej{r}\in\tilde{S}_0$. This concludes the proof that $\tilde{S}\subseteq\tilde{S}_0$.

\subsection{Proofs of the Revision and Update Postulates}\label{app:AGM-update}

In
what follows, we assume that $\useq{P}$ is a nonempty sequence 
$(P_{1},\ldots,P_{n})$ of ELPs. 

\begin{description}

\item [(K1)] $(\useq{P},P)$ represents a belief set.  
\end{description}
This holds by convention.

\begin{description}

\item[(K2) \& (U1)] $P \subseteq \bel((\useq{P},P))$. 
\end{description}
This
is clearly satisfied, as the rules of $P$ cannot be rejected in the
updated program.

\begin{description}
\item[(U2)] $\bel (P) \subseteq \bel (\useq{P})$ implies $\bel ((\useq{P},P)) =
\bel (\useq{P})$. 
\end{description}
This 
postulate states that
no change occurs if the update is already entailed. 
This
means that inconsistency is preserved under updates and
contradictions cannot be removed by updates. This is clearly not
the case 
: Updating $\useq{P}=\{a\la~, \neg a\la \}$
with $P = \{ a\la\}$ removes inconsistency. Also for a consistent
$\useq{P}$, update by a logically implied rule may lead to a change in
semantics. Consider, e.g., $\useq{P} = \{ a \la \naf b\}$ and $P = \{ b
\la \naf a\}$. Then $\useq{P}$ has the unique answer set $S=\{a\}$, and 
$S\models b \la\naf a$. However, $(\useq{P},P)$ has, besides $S$, another
answer set $S' = \{ b\}$.

\begin{description}
\item[(K3)] $\bel ((\useq{P},P)) \subseteq \bel(\bel(\useq{P}) \cup P)$.
\end{description}
 This
property fails in general, if programs have infinite alphabets.
This can be seen by the following example. Let $\useq{P} = P_1$ and $P=P_2$, where 
\begin{eqnarray*}
P_1 & = & \{ a_i \la \naf
b_i,\, b_i \la \naf a_i,\, c\la a_i \mid i\geq 1\} \cup \{ ~\la \naf c
\}; \\
P_2 & = & \{ \ \la b_i \mid i \geq 1\}.
\end{eqnarray*}
 It is easy to see that
every answer set $S$ of $P_1$ must contain $c$, and that either $a_i$
or $b_i$ (but not both) are contained in $S$. Therefore, $c \in
\bel(\useq{P})$ holds.  Furthermore, $S' = \{ a_i \mid i \geq 1\} \cup \{
c\}$ is an answer set of $\bel(\useq{P})$. Since $S' \models P_2$, it follows that $S'$ is an answer set of $\bel(\useq{P})\cup P$. This implies
$\bel(\bel(\useq{P})\cup P) \subset \bel(\{ \la \})$, i.e.,
$\bel(\bel(\useq{P})\cup P)$ does not contain all possible rules.

On the other hand, $\bel((\useq{P},P)) = \bel(\{ \la \})$: Since negation
does not occur in rule heads of $P_1$ and $P_2$, we have
$\rs(S,(\useq{P},P)) = \emptyset$, and thus $\SMup((\useq{P},P)) = \SM(P_1\cup P_2)$ holds. However, $P_1\cup P_2$ clearly has no answer set, which
implies $\bel((\useq{P},P))=\bel(P_1\cup P_2) = \bel(\{ \la\})$. 
It follows that $\bel((\useq{P},P)) \not\subseteq \bel(\bel(\useq{P})\cup P)$,
which proves our claim. 

That property (K3) holds if either $\useq{P}$  or $P$ has a finite alphabet
follows from (K7), which subsumes (K3) by choosing $P =
\emptyset$ in (K7), and by virtue of $\bel((\useq{P},\emptyset)) = \bel(\useq{P})$. 

\begin{description}
\item[(U3)] If both $\useq{P}$ and $P$ are satisfiable, then 
$(\useq{P},P)$ is satisfiable. 
\end{description}
This is clearly violated. Consider, \egc $\useq{P}=P_1$ and $P=P_2$, where
\begin{eqnarray*}
P_1 & =& \{ a \la b, \naf a\}; \\
P_2 & = & \{ b\la~\}.
\end{eqnarray*}

\begin{description}
\item[(K4)] If $\bel(\useq{P})\cup P$ has an answer set, then
$\bel(\bel(\useq{P})\cup P)\subseteq \bel ((\useq{P},P))$.
\end{description}
 The property
is violated. Consider 
$P_1 = \{ a\la~$, $b\la \naf c$,
$c\la\naf b\}$ and $P_2 = \{ \neg a\la b\}$.  As easily seen, the
sequence $(P_1,P_2)$ has two answer sets, $S = \{ b, \neg a\}$ and $S' =
\{ a, c\}$. On the other hand, since $P_1 \subseteq
\bel(P_1)$, $S$ cannot be an answer set of
$\bel(P_1)\cup P_2$; in fact, $S'$ is its unique answer set. Since, \egc $S' \models c\la \naf a, b$ whilst $S \not\models
c\la \naf a, b$, it follows that $\bel(\bel(P_1)\cup P_2)\not\subseteq
\bel ((P_1,P_2))$.

\begin{description}
\item[(K5)] $(\useq{P},P)$ is unsatisfiable iff $P$ is
unsatisfiable. 
\end{description}
This is violated, since contradictory rules in
$\useq{P}$ are not affected unless they are rejected by rules in
$P$. For instance, if $\useq{P}$ consists of the single program $\{ a\la~$, $\neg a \la ~\}$,  then the update of $\useq{P}$ by $P=\{
b\la ~\}$ does not have an answer set. 

\begin{description}
\item[(K6) \& (U4)] $\useq{P}\equiv \useq{P}'$ and $P\equiv P'$ implies
$(\useq{P},P) \equiv (\useq{P}',P')$. 
\end{description}
This expresses {\em irrelevance of
syntax} which is clearly not satisfied, since rejection of rules
depends on their syntactical form. For instances, take $\useq{P}=\useq{P}' = P_1$, $P=P_2$, and $P'=P'_2$, where
\begin{eqnarray*}
P_1 &=& \{ a \la~, b\la~\};\\
P_2 &=& \{ \neg a \la b\};\\
P'_2 &=& \{ \neg b \la a\}.
\end{eqnarray*}
 Then clearly $\useq{P}\equiv\useq{P}'$ and $P_2 \equiv P'_2$, but the resulting updates have
different answer sets: $\{ \neg a,b \}$ is an answer set of $(P_1,P_2)$ but not of $(P_1,P'_2)$.

\begin{description}
\item[(K7) \& (U5)] $\bel ((\useq{P},P\cup P')) \subseteq \bel(\bel((\useq{P},P))
\cup P')$. 
\end{description}
The property does not hold if both $\useq{P}$ and $P'$ (or $P$
and $P'$) have infinite alphabets, which follows from the example
showing the failure of (K3) (set 
$P=\emptyset$,
 and exploit the relation $\bel((\useq{P},\emptyset))=\bel(\useq{P})$).

Property (K7) holds
if $(\cup\useq{P})\cup P$ or $P'$ has
a finite alphabet. Towards a contradiction, suppose it fails. Then, there exists 
$r\in \bel ((\useq{P},P\cup P')) \setminus \bel(\bel((\useq{P},P))
\cup P')$, and hence an answer set $S \in \AS(\bel((\useq{P},P))\cup P')$
such that $S\not\models r$. 

Consider $\useq{P}'=(\useq{P},P)$, and let $\at'$ denote the atoms in
$\useq{P}'$. Then, for every finite set of atoms $\at_0 \subseteq
\at'$, there must exist some answer set $S_{\at_0}$ of $\useq{P}'$ such that $S$
and $S_{\at_0}$ coincide with respect to $\at_0$. Indeed, $\bel(\useq{P}')$
must contain, for each interpretation $M$ which does not coincide with
any answer set of $\useq{P}'$ with respect to $\at_0$, the constraint $\la
L_1,\ldots,L_m,\naf L_{m+1},\ldots,\naf L_n$, where $\{L_1,\ldots,L_m\} = \Lit_{\at_0}\cap M$ and  
$\{L_{m+1},\ldots,L_n\} = \Lit_{\at_0}\setminus M$,
respectively. Furthermore, all answer sets of $\bel(\useq{P}')\cup P'$
must coincide on the atoms in $\at\setminus\at'$. 
Thus, assuming that either $\useq{P}'$ 
or $P'$ has a finite alphabet, it follows that  $S$ is an answer set of $\useq{P}'$ or $P'$. Without loss of generality, we assume that $S$ is an answer set of $\useq{P}'$.

Hence, Theorem~\ref{theo:sequence-char-0} implies that $S$ is a minimal model
of $((\cup\useq{P}')\setminus \rs(S,\useq{P}'))^S$. Since $S \models P'$, 
we conclude that $S$ is also a minimal model of $(((\cup\useq{P}')\cup P')\setminus \rs(S,\useq{P}))^S$. Furthermore, for the
update sequence $\useq{P}'' = (\useq{P},P\cup P')$,  it holds
 that $\rs(S,\useq{P}'') = \rs(\useq{P}')$. Indeed, $S\models r'$ for all $r'\in P\cup P'$, thus $r'\notin
\rs(S,\useq{P}')$ and $r\notin \rs(S,\useq{P}'')$. Equivalence for the
rules in $\useq{P}$ can be shown by induction on the length of $\useq{P}$.
 Hence, we obtain that $S$ is a minimal model of 
$((\cup\useq{P}'')\setminus \rs(S,\useq{P}''))^S$. From
Theorem~\ref{theo:sequence-char-0}, we obtain that $S$ is an answer set of
$\useq{P}''$. Since $S\not\models r$, it follows $r\notin
\bel(\useq{P},P\cup P')$, a contradiction. 

\begin{tabbing}
{\bf (U6)} \ \= Given $\bel (P')\subseteq \bel ((\useq{P},P))$ and  $\bel (P)
\subseteq \bel ((\useq{P},P'))$, then \\
\>  $\bel ((\useq{P},P))= \bel ((\useq{P},P'))$.
\end{tabbing}
 This postulate fails. Consider, \egc $\useq{P} = P_1$, $P=P_2$, and $P'=P_3$, where
\begin{eqnarray*}
P_1 &=& \{ \ b \la \ , d\la \ \};\\
P_2 &=& \{ \ \neg a \la \ , \neg e \la d , \neg d \la e \ \};\\
P_3 &=& \{ \ \neg a \la \ , \neg c \la b, \neg b\la b \ \}. 
\end{eqnarray*}
Then, $\{\neg a,b,d,\neg e\}$ is the unique answer set of $(P_1,P_2)$,
and $\{\neg a,b,d,\neg c\}$ is the unique answer set of
$(P_1,P_3)$. Moreover, it is easily verified that $\neg a \in S\cap
S'$, for any answer set $S$ of $P_2$ and any answer set $S'$ of
$P_3$. Hence, $\bel(P_3)\subseteq\bel((P_1,P_2))$ and
$\bel(P_2)\subseteq\bel((P_1,P_3))$.  However, $\bel ((P_1,P_2)) \neq
\bel ((P_1,P_3))$.

\begin{tabbing}
{\bf (K8)} \ \= If $\bel((\useq{P},P)) \cup P'$ is satisfiable, then \\ 
\> $\bel(\bel(\useq{P},P)\cup P') \subseteq \bel ((\useq{P},P\cup P'))$. 
\end{tabbing}
This postulate fails. Setting $P=\emptyset$, the property reduces to (K4)
since $\bel(\useq{P},\emptyset)=\bel(\useq{P})$. The failure follows from the
failure of (K4).

\subsection{Proofs of the Postulates for Iterated Revision}\label{app:iterated-revision}

\begin{description}
  
\item[(C1)] If $P' \subseteq \bel(P)$, then $\bel((\useq{P},P',P)) =
\bel((\useq{P},P))$. 
\end{description}
Adding rules which are implied after the
previous update does not change the epistemic state.  This is not
satisfied: take, \egc $\useq{P} = \emptyset$, $P =\{b\la \naf a\}$, and
$P'=\{a \la\naf b\}$. Then $(\useq{P},P',P)$ has two answer sets, while
$(\useq{P},P)$ has a single answer set. The associated belief sets are
thus different.
  
\begin{description}
\item[(C2)] If $S \not\models P'$, for all $S \in \AS(P)$, then
$\bel((\useq{P},P,P')) = \bel((\useq{P},P'))$. 
\end{description}
This property is not satisfied. For a counterexample,
consider $P_1 = \{ a\la b\}$, $P_2=\{b\la \ \}$, and
 $P_3=\{\neg b \la\naf a\}$. Then $(P_1,P_2,P_3)$ has two answer sets,
$\{a,b\}$ and $\{ \neg b\}$, whilst $(P_1,P_3)$ possesses the single
answer set $\{ \neg b\}$.
  
\begin{description}
\item[(C3)] If $P' \subseteq \bel((\useq{P},P))$, then $P' \subseteq
\bel((\useq{P},P',P))$.
\end{description}
 Implied rules can be added before the update.
This property fails in general. For example, let
$\useq{P}=P_1$, $P=P_2$, and $P'=P_3$, where
\begin{eqnarray*}
P_1 &=& \emptyset;\\
P_2 &=& \{ \ a \la \naf b, \ b\la \naf a, \ g \la a, \ g
\la\naf g, \  c\la \ \};\\
P_3 &=& \{ \ g\la, \ \neg c \la \naf a \ \}.
\end{eqnarray*}
 Note
that $P_2$ has a single answer set, $S=\{a, g, c\}$, and clearly $S
\models P_3$. However, $(P_1,P_3,P_2)$ has among its answer sets $S'=\{ b, g,
c\}$, and $S' \not\models \neg c\la \naf a$. 

The property holds, however, providing $P'$ contains a single rule.
Suppose $P' \subseteq \bel((\useq{P},P))$ but $r \notin \bel((\useq{P},P',P))$,
for $P'=\{r\}$. Then, $r \in Rej(S,(\useq{P},P',P))$ for some answer set $S$ of $(\useq{P},P',P)$. This means, however, that $S$ is an answer set of $(\useq{P},P)$ (as $r$ cannot reject any rule in
$\useq{P}$). Thus, $r\notin \bel((\useq{P},P))$.

\begin{description}
\item[(C4)] If $S \models P'$ for some $S \in \ASup((\useq{P},P))$, then
$S\models P'$ for some $S \in \ASup((\useq{P},P',P))$.
\end{description}
 This property
holds.  By hypothesis, there exists some $S \in \ASup((\useq{P},P))$ 
such that $S\models P'$. By Theorem~\ref{theo:sequence-char-0}, $S$ is a minimal model of 
$$
\big(((\cup\useq{P})\cup  P)\setminus \rs(S,(\useq{P},P))\big)^S.
$$
 Since $S\models P'$ and 
$S\models P$ (due to $S \in \ASup((\useq{P},P))$), no rule $r' \in P'$ can be rejected 
by a rule $r$ of $P$. Also, $r'$ can reject a rule $r''$ in $P$ only if $r''$ 
is rejected within $P$. Thus, 
$\rs(S,(\useq{P},P)) = \rs(S,(\useq{P},P',P))$, and $S$ is a minimal model of 
$$
\big(((\cup\useq{P})\cup P'\cup  P)\setminus \rs(S,(\useq{P},P',P))\big)^S.$$
 This means, by
Theorem~\ref{theo:sequence-char-0}, that $S$ is an answer set of
$(\useq{P},P',P)$.

\begin{tabbing}
{\bf (C5)} \ \= If $S\not\models P'$ for all $S\in \ASup((\useq{P},P))$ and $P
\not\subseteq \bel((\useq{P},P'))$, then \\
\> $P \not\subseteq
\bel((\useq{P},P,P'))$.
\end{tabbing}
 This property fails: just consider $\useq{P} =
\emptyset$, $P=\{ a\la \ \}$, and $P' = \{b\la \ \}$.  

\begin{tabbing}
{\bf (C6)} \ \=  If $S \not\models P'$ for all $S \in \ASup((\useq{P},P))$ and
$S\not\models P$ for all $S \in \ASup((\useq{P},P'))$, then \\
\>  $S\not\models P$ for all $S \in \ASup((\useq{P},P,P'))$. 
\end{tabbing}

This property fails as well, which can be seen by the counterexample for (C5), setting $\useq{P}=\emptyset$. Another counterexample for (C6)---which does not exploit minimization of answer sets---is $\useq{P} =\{ \neg b\la \ , \ \neg a\la b\}$, $P=\{a\la \ \}$, and $P' = \{b\la \ \}$.

\begin{description}
\item[(I1)] $\bel(\useq{P})$ is a consistent belief set. 
\end{description}
This is clearly violated in general.
  
\begin{description}
\item[(I2)] $P \subseteq \bel((\useq{P},P))$.
\end{description}
The postulate is easily seen to be satisfied (cf.\ (K2) and (U1)).

\begin{tabbing}
{\bf (I3)} \ \=  If $L_0 \la \ \in \bel((\useq{P}, \{ L_1 \la \ ,\ldots, L_k\la \ \}))$, then \\
\> $L_0 \la
L_1,\ldots,L_k \in \bel(\useq{P})$. 
\end{tabbing}
This property holds. Suppose there is some $S \in \ASup(\useq{P})$ such that
$\{ L_1,\ldots,L_k\}\subseteq S$ but $L_0\notin S$. 
 Let $\useq{P}'= 
(\useq{P},
\{ L_1\la \ ,\ldots, L_k \la \ \})$.
 Then, the following holds: For every rule $r$ in $\useq{P}'$, $r \in \rs(S,\useq{P}')$ iff $r\in \rs(S,\useq{P})$. Indeed, each $L_i$ ($1\leq i\leq n$) is
neither in $\rs(S,\useq{P}')$ nor in $\rs(S,\useq{P})$.
By
Theorem~\ref{theo:sequence-char-0}, $S$ is a minimal model of $((\cup\useq{P})\setminus
\rs(S,\useq{P}))^S$. It follows that $S$ is a minimal model of $((\cup\useq{P}')\setminus
\rs(S,\useq{P}'))^S$, which in turn implies, by using Theorem~\ref{theo:sequence-char-0} again, that $S\in \SMup(\useq{P})$. Since $L_0\notin S$, we obtain $L_0 \la \ \notin
\bel(\useq{P}')$.

\begin{description}
\item[(I4)] If $Q_1 \subseteq \bel(\useq{P})$, then
$\bel((\useq{P},Q_1,Q_2,\ldots,Q_n)) = \bel((\useq{P},Q_2,\ldots,Q_n))$.
\end{description}
 This
property fails. Consider
$\useq{P} = \{ a \la \naf b\}$ and $Q_1 =
\{b\la \naf a\}$ for $n=1$.

\begin{tabbing}
{\bf (I5)} \ \= If $\bel(Q_2)\subseteq \bel(Q_1)$, then \\
\> $\bel((\useq{P},Q_1,Q_2,Q_3,\ldots,Q_n))=\bel((\useq{P},Q_2,Q_3,\ldots,Q_n))$.
\end{tabbing}
This property fails, because it generalizes (C1), which fails.

\begin{tabbing}
{\bf (I6)} \ \= If $S \models Q_2$ for some $S \in \ASup((\useq{P},Q_1))$, then\\
\> $\bel((\useq{P},Q_1,Q_2,Q_3,\ldots,Q_n))= \bel((\useq{P},Q_1,Q_1\cup
                Q_2,Q_3,\ldots,Q_n))$. 
\end{tabbing}
The property fails: Let $\useq{P} =\{ a\la \naf b, \ b\la
\naf a\}$, $Q_1 = \{ c\la \ \}$, and $Q_2 = \{ \neg c\la a \}$. Then,
$S = \{ c,b\}$  is an answer set of $(\useq{P},Q_1)$ such that $S\models
Q_2$.  However, $(\useq{P},Q_1,Q_2)$ has two answer sets, $S_1 = \{ a, \neg c \}$ and $S_2 = \{ c,
b\}$, whilst $(\useq{P},Q_1,Q_1\cup Q_2)$ has the single answer set $\{ c,b\}$.

\subsection{Proofs of the Postulates of Updates as Nonmonotonic Consequence Relations}\label{app:nonmon}

\begin{description}
        \item[(N1)] $P_1\in\bel((\useq{P},P_1))$.
\end{description}
This is clearly satisfied (cf.\ (K2), (U1), and (I2)).

\begin{tabbing}
{\bf (N2)} \ \=   If $\bigcup_{i=1}^m Q_i\subseteq\bel((\useq{P},P_1))$ and $P_2\subseteq\bel((\useq{P},P_1\cup \bigcup_{i=1}^m Q_1))$, then \\ 
\> $P_2\subseteq\bel((\useq{P},P_1))$.
\end{tabbing}
The property holds. Let $Q=\bigcup_{i=1}^m Q_i$ and $\useq{P}'=(\useq{P},P_1)$. Assume $Q\subseteq\bel(\useq{P}')$ and $P_2\subseteq\bel((\useq{P},P_1\cup Q))$, and consider some answer set $S$ of $\useq{P}'$. Then, $S\models Q$. Moreover, $\corr{S}$ is an answer set of $\useq{P}'_\upd\cup Q$. Since $Q\subseteq\bel(\useq{P}')$, it follows that for each rule $s \in Q$ rejecting a rule $r$ from $\useq{P}$, there exists a rule $r_1 \in P_1$ also rejecting $r$. Hence, no further rule in $\useq{P}$ can be rejected using $Q$. Let  $\useq{P}''=(\useq{P},P_1 \cup Q)$. Then, $\corr{S}$ is an answer set of $\useq{P}''_\upd$, so $S$ is an answer set of $(\useq{P},P_1\cup Q)$. Since $P_2\subseteq\bel((\useq{P},P_1 \cup Q))$, we obtain $S\models P_2$. This proves the property.

\begin{tabbing}
{\bf (N3)} \ \= If $\bigcup_{i=1}^m Q_i\subseteq\bel((\useq{P},P_1))$ and $P_2\subseteq\bel((\useq{P},P_1))$, then  \\
\> $P_2\subseteq\bel((\useq{P},P_1\cup \bigcup_{i=1}^m Q_1))$.
\end{tabbing}
The property fails: Consider the counterexample $\useq{P}= \emptyset$, $P_1 = \{ a \la \naf b\}$, $P_2= \{ a \la \ \}$, and, for $m=1$,  $Q_1= \{ b \la \naf a\}$.

\begin{tabbing}
{\bf (N4)} \ \= If $P_{i+1}\subseteq\bel((\useq{P},P_i))$ ($1\leq i<n$)  and   $P_1\subseteq\bel((\useq{P},P_n)) \ (n\geq 2)$, then \\
\> $\{ P' \mid P'\subseteq\bel((\useq{P},P_i)) \}  = \{ P' \mid P'\subseteq\bel((\useq{P},P_j))\}$, for all $i,j\leq n$.
\end{tabbing}
The property does not hold, because it includes (U6) as a special case, which fails. 

\begin{description}
        \item[(P1)] If $P_1 \equiv P_2$ and $P_3\subseteq\bel((\useq{P},P_1))$, then  $P_3\subseteq\bel((\useq{P},P_2))$.
\end{description}
The property fails, due to the following counterexample: $\useq{P} = \{ a \la \ , b \la \ \}$, $P_1 = \{ \neg a \la b\}$, $P_2 = \{ \neg b \la a\}$, and $P_3 = \{ b \la \ \}$.

\begin{description}
        \item[(P2)] If $P_1 \models P_2$ and $P_1\subseteq\bel((\useq{P},P_3))$, then  $P_2\subseteq\bel((\useq{P},P_3))$.
\end{description} 
This property does not hold. For a counterexample, consider $\useq{P} = \emptyset$, $P_1 = \{ a \la \naf b\}$, $P_2 = \{ a \la \ \}$, and $P_3 = \{ b \la \ , \neg a \la \ \}$.

\begin{tabbing}
{\bf (P4)} \ \= If $P_2\subseteq\bel((\useq{P},P_1))$ and $P_3\subseteq\bel((\useq{P},P_1))$, then $P_2 \cup P_3\subseteq\bel((\useq{P},P_1))$.
\end{tabbing}
The property is trivially satisfied.


\bibliographystyle{tlp}


\begin{thebibliography}{}

\bibitem[\protect\citename{Alchourr{\'o}n {\em et~al.}\relax,
  }1985]{alch-etal-85}
Alchourr{\'o}n, C.E., G{\"a}rdenfors, P., \& Makinson, D. (1985).
\newblock {On the Logic of Theory Change: Partial Meet Functions for
  Contraction and Revision}.
\newblock {\em {Journal of Symbolic Logic}}, {\bf 50}, 510--530.

\bibitem[\protect\citename{Alferes \& Pereira, }1997]{alfe-pere-97}
Alferes, J.J., \& Pereira, L.M. (1997).
\newblock {Update-programs can Update Programs}.
\newblock {\em Pages  110--131 of:} Dix, J., Pereira, L.M., \& Przymusinski, T.C.
  (eds), {\em {Selected Papers from the ICLP'96 Workshop on Nonmonotonic Extensions of Logic Programming $($NMELP'96\/$)$}}.
\newblock {LNAI}, vol. 1216.
\newblock {Springer}.

\bibitem[\protect\citename{Alferes \& Pereira, }2000]{alfe-pere-00}
Alferes, J.J., \& Pereira, L.M. (2000).
\newblock {Updates plus Preferences}.
\newblock {\em Pages 345--360 of:} Aciego, M.O., de~Guzm{\'a}n, I.P., Brewka, G., \& Pereira, L.M.
  (eds), {\em {Proc.\ Seventh Europ.\ Workshop on Logic in
  Artificial Intelligence $($JELIA 2000\/$)$}}.
\newblock {LNAI}, vol. 1919.
\newblock {Springer}.

\bibitem[\protect\citename{Alferes {\em et~al.}\relax, }1998]{alfe-etal-98}
Alferes, J.J., Leite, J.A., Pereira, L.M., Przymusinska, H., \& Przymusinski,
  T.C. (1998).
\newblock {Dynamic Logic Programming}.
\newblock {\em Pages  98--111 of:} Cohn, A., Schubert, L., \& Shapiro, C. (eds), {\em {Proc.\ Sixth Int.\ Conf.\ on Principles
  of Knowledge Representation and Reasoning $($KR'98\/$)$}}.
\newblock {Morgan Kaufmann}.

\bibitem[\protect\citename{Alferes {\em et~al.}\relax, }1999]{alfe-etal-99b}
Alferes, J.J., Pereira, L.M., Przymusinska, H., \& Przymusinski, T.C. (1999).
\newblock {LUPS - A Language for Updating Logic Programs}.
\newblock {\em Pages  162--176 of:} Gelfond, M., Leone, N., \& Pfeifer, G.
  (eds), {\em {Proc.\ Fifth Int.\ Conf.\ on Logic
  Programming and Nonmonotonic Reasoning $($LPNMR'99\/$)$}}.
\newblock {LNAI}, vol. 1730.
\newblock {Springer}.

\bibitem[\protect\citename{Alferes {\em et~al.}\relax, }2000]{alfe-etal-99a}
Alferes, J.J., Leite, J.A., Pereira, L.M., Przymusinska, H., \& Przymusinski,
  T.~C. (2000).
\newblock {Dynamic Updates of Non-Monotonic Knowledge Bases}.
\newblock {\em {Journal of Logic Programming}}, {\bf 45}(1--3), 43--70.

\bibitem[\protect\citename{Baral \& Gelfond, }1994]{bara-gelf-94}
Baral, C., \& Gelfond, M. (1994).
\newblock {Logic Programming and Knowledge Representation}.
\newblock {\em {Journal of Logic Programming}}, {\bf 19/20}, 73--148.

\bibitem[\protect\citename{Benferhat {\em et~al.}\relax, }1993]{benf-etal-93}
Benferhat, S., Cayrol, C., Dubois, D., Lang, J., \& Prade, H. (1993).
\newblock {Inconsistency Management and Prioritized Syntax-Based Entailment}.
\newblock {\em Pages  640--645 of:} Bajcsy, R. (ed), {\em {Proc.
  Thirteenth Int.\ Joint Conf.\ on Artificial Intelligence
  $($IJCAI'93\/$)$}}.
\newblock {Morgan Kaufmann}.

\bibitem[\protect\citename{Brewka, }1991a]{Brewka91}
Brewka, G. (1991a).
\newblock Cumulative {D}efault {L}ogic: {I}n {D}efense of {N}onmonotonic
  {I}nference {R}ules.
\newblock {\em {Artificial Intelligence}}, {\bf 50}(2), 183--205.

\bibitem[\protect\citename{Brewka, }1991b]{brew-91a}
Brewka, G. (1991b).
\newblock {\em {Nonmonotonic Reasoning: Logical Foundations of Commonsense}}.
\newblock {Cambridge Tracts in Theoretical Computer Science}, vol.~12.
\newblock {Cambridge Univ.\ Press}.

\bibitem[\protect\citename{Brewka, }2000]{brew-00} Brewka, G. (2000).
\newblock {Declarative Representation of Revision Strategies}.
\newblock {\em Pages 18--22 of:} Horn, W. (ed), {\em {Proc.
Fourteenth Europ.\ Conf.\ on Artificial Intelligence $($ECAI
2000\/$)$}}.  \newblock {IOS Press}. Preliminary version presented at
Intl.\ Workshop on Nonmonotonic Reasoning, Breckenridge,
2000. \newblock URL: http://arXiv.org/abs/cs.AI/0003036.

\bibitem[\protect\citename{Buccafurri {\em et~al.}\relax, }1996]{bucc-etal-96}
Buccafurri, F., Leone, N., \& Rullo, P. (1996).
\newblock {Stable Models and their Computation for Logic Programming with
  Inheritance and True Negation}.
\newblock {\em {Journal of Logic Programming}}, {\bf 27}(1), 5--43.

\bibitem[\protect\citename{Buccafurri {\em et~al.}\relax,
  }1999a]{bucc-etal-99a-iclp}
Buccafurri, F., Faber, W., \& Leone, N. (1999a).
\newblock {Disjunctive Logic Programs with Inheritance}.
\newblock {\em Pages  79--93 of:} De Schreye, D. (ed), {\em {Proc.
  Sixteenth Int.\ Conf.\ on Logic Programming $($ICLP'99\/$)$}}.
\newblock {MIT Press}.

\bibitem[\protect\citename{Cholewi{\'n}ski {\em et~al.}\relax,
  }1996]{chol-etal-96}
Cholewi{\'n}ski, P., Marek, V.W., \& Truszczy{\'n}ski, M.
  (1996).
\newblock {Default Reasoning System DeReS}.
\newblock {\em Pages 518--528 of:} Carlucci Aiello, L., Doyle, J., \& Shapiro, S. (eds) {\em Proc.\ Fifth Int.\ Conf.\ on Principles of
  Knowledge Representation and Reasoning (KR '96)}.
\newblock {Morgan Kaufmann}.

\bibitem[\protect\citename{Dantsin {\em et~al.}\relax, }1997]{dant-etal-97}
Dantsin, E., Eiter, T., Gottlob, G., \& Voronkov, A. (1997).
\newblock {Complexity and Expressive Power of Logic Programming}.
\newblock {\em Pages  82--101 of:} {\em {Proc.\ Twelfth IEEE
  Int.\ Conf.\ on Computational Complexity $($CCC'97\/$)$}}.
\newblock {IEEE Computer Society Press}.

\bibitem[\protect\citename{Darwiche \& Pearl, }1997]{darw-pear-97}
Darwiche, A., \& Pearl, J. (1997).
\newblock {On the Logic of Iterated Belief Revision}.
\newblock {\em {Artificial Intelligence}}, {\bf 89}(1--2), 1--29.

\bibitem[\protect\citename{Delgrande {\em et~al.}\relax, }2000]{delg-etal-00}
Delgrande, J., Schaub, T., \& Tompits, H. (2000).
\newblock {Logic Programs with Compiled Preference}.
\newblock {\em Pages  392--398 of:} Horn, W. (ed), {\em {Proc.
  Fourteenth Europ.\ Conf.\ on Artificial Intelligence $($ECAI
  2000\/$)$}}.
\newblock {IOS Press}.

\bibitem[\protect\citename{Eiter \& Gottlob, }1992]{eite-gott-92e}
Eiter, T., \& Gottlob, G. (1992).
\newblock {On the Complexity of Propositional Knowledge Base Revision, Updates
  and Counterfactuals}.
\newblock {\em {Artificial Intelligence}}, {\bf 57}(2--3), 227--270.

\bibitem[\protect\citename{Eiter \& Gottlob, }1995]{eite-gott-93b}
Eiter, T., \& Gottlob, G. (1995).
\newblock {On the Computational Cost of Disjunctive Logic Programming:
  Propositional Case}.
\newblock {\em {Annals of Mathematics and Artificial Intelligence}}, {\bf
  15}(3--4), 289--323.

\bibitem[\protect\citename{Eiter {\em et~al.}\relax, }1997a]{eite-etal-97a}
Eiter, T., Leone, N., Mateis, C., Pfeifer, G., \& Scarecello, F. (1997a).
\newblock {A Deductive System for Nonmonotonic Reasoning}.
\newblock {\em Pages  363--374 of:} Dix, J., Furbach, U., \& Nerode, A. (eds),
  {\em {Proc.\ Fourth Int.\ Conf.\ on Logic Programming
  and Nonmonotonic Reasoning $($LPNMR'97\/$)$}}.
\newblock {LNAI}, vol. 1265.
\newblock {Springer}.

\bibitem[\protect\citename{Eiter {\em et~al.}\relax, }1997b]{eite-etal-97q}
Eiter, T., Gottlob, G., \& Leone, N. (1997b).
\newblock {Abduction From Logic Programs: Semantics and Complexity}.
\newblock {\em {Theoretical Computer Science}}, {\bf 189}(1--2), 129--177.

\bibitem[\protect\citename{Eiter {\em et~al.}\relax, }1997c]{Eiter:1997:DD}
Eiter, T., Gottlob, G., \& Mannila, H. (1997c).
\newblock Disjunctive {Datalog}.
\newblock {\em {ACM} {T}ransactions on {D}atabase {S}ystems}, {\bf 22}(3),
  364--418.

\bibitem[\protect\citename{Eiter {\em et~al.}\relax, }2000a]{eite-etal-00f}
Eiter, T., Fink, M., Sabbatini, G., \& Tompits, H. (2000).
\newblock {Considerations on Updates of Logic Programs}.
\newblock {\em Pages 2--20 of:} Aciego, M.O., de~Guzm{\`a}n, I.P., Brewka, G., \& Pereira, L.M.
  (eds), {\em {Proc.\ Seventh Europ.\ Workshop on Logic in
  Artificial Intelligence $($JELIA 2000\/$)$}}.
\newblock {LNAI}, vol. 1919.
\newblock {Springer}.

\bibitem[\protect\citename{Eiter {\em et~al.}\relax, }2000b]{eite-etal-00g}
Eiter, T., Fink, M., Sabbatini, G., \& Tompits, H. (2000).
\newblock {On Updates of Logic Programs: Semantics and Properties}.
\newblock {Technical Report INFSYS RR-1843-00-08, Technical Univ.\ of Vienna, Knowledge-Based Systems Group}.

\bibitem[\protect\citename{Eiter {\em et~al.}\relax, }1998]{eite-etal-98a}
Eiter, T., Leone, N., Mateis, C., Pfeifer, G., \& Scarcello,
  F. (1998).
\newblock {The KR System {\tt dlv}: Progress Report, Comparisons and
  Benchmarks.}
\newblock {\em Pages  406--417 of:} Cohn, A., Schubert, L., \& Shapiro, S. (eds), {\em Proc. Sixth Int.\ Conf.\ on
  Principles of Knowledge Representation and Reasoning (KR'98)}.
\newblock {Morgan Kaufmann}.

\bibitem[\protect\citename{Gabbay, }1985]{gabbay:1985a}
Gabbay, {D.M.} (1985).
\newblock Theoretical {F}oundations for {N}on-{M}onotonic {R}easoning in
  {E}xpert {S}ystems.
\newblock {\em Pages  439--457 of:} Apt, K.~R. (ed), {\em Logics and {M}odels
  of {C}oncurrent {S}ystems}.
\newblock {Springer}.

\bibitem[\protect\citename{G{\"a}rdenfors \& Makinson,
  }1994]{gardenfors-makinson:1994a}
G{\"a}rdenfors, P., \& Makinson, D. (1994).
\newblock Nonmonotonic {I}nferences {B}ased on {E}xpectations.
\newblock {\em {Artificial Intelligence}}, {\bf 65}(2), 197--245.

\bibitem[\protect\citename{G{\"a}rdenfors \& Rott, }1995]{gard-rott-95}
G{\"a}rdenfors, P., \& Rott, H. (1995).
\newblock {Belief Revision}.
\newblock {\em Pages  35--132 of:} Gabbay, D.M., Hogger, C.J., \& Robinson,
  J.A. (eds), {\em {Handbook of Logic in AI and Logic
  Programming, Volume IV}}.
\newblock {Oxford Science Publications}.

\bibitem[\protect\citename{Gelfond \& Lifschitz, }1988]{gelf-lifs-88}
Gelfond, M., \& Lifschitz, V. (1988).
\newblock {The Stable Model Semantics for Logic Programming}.
\newblock {\em Pages  1070--1080 of:} Kowalski, R., \& Bowen, K.A. (eds), {\em
  {Proc.\ Fifth Int.\ Conf.\ and Symposium on Logic
  Programming}}.
\newblock {MIT Press}.

\bibitem[\protect\citename{Gelfond \& Lifschitz, }1991]{gelf-lifs-91}
Gelfond, M., \& Lifschitz, V. (1991).
\newblock {Classical Negation in Logic Programs and Disjunctive Databases}.
\newblock {\em {New Generation Computing}}, {\bf 9}(3--4), 365--385.

\bibitem[\protect\citename{Inoue, }2000]{inou-00}
Inoue, K. (2000).
\newblock {A Simple Characterization of Extended Abduction}.
\newblock {\em Pages  718--732 of:} Lloyd, J.W., Dahl, V., Furbach, U., Kerber, M., Lau, K.K., Palmidessi, C., Pereira, L.M., Sagiv, Y., \& Stuckey, P.J. (eds), {\em {Proc.\ First Int.\ Conf.\ on Computational Logic $($CL 2000\/$)$}}.
\newblock {LNAI}, vol. 1861.
\newblock {Springer}.

\bibitem[\protect\citename{Inoue \& Sakama, }1995]{inou-saka-95b}
Inoue, K., \& Sakama, C. (1995).
\newblock {Abductive Framework for Nonomonotonic Theory Change}.
\newblock {\em Pages  204--210 of:} {\em {Proc.\ Fourteenth
  Int.\ Joint Conf.\ on Artificial Intelligence $($IJCAI'95\/$)$}}.
\newblock {Morgan Kaufmann}.

\bibitem[\protect\citename{Inoue \& Sakama, }1999]{inou-saka-99}
Inoue, K., \& Sakama, C. (1999).
\newblock {Updating Extended Logic Programs through Abduction}.
\newblock {\em Pages  147--161 of:} Gelfond, M., Leone, N., \& Pfeifer, G.
  (eds), {\em {Proc.\ Fifth Int.\ Conf.\ on Logic
  Programming and Nonmonotonic Reasoning $($LPNMR'99\/$)$}}.
\newblock {LNAI}, vol. 1730.
\newblock {Springer}.

\bibitem[\protect\citename{Johnson, }1990]{john-90}
Johnson, D.S. (1990).
\newblock {A Catalog of Complexity Classes}.
\newblock {\em Pages  67--161 of:} Van Leeuwen, J. (ed), {\em {Handbook of
  Theoretical Computer Science}},  vol. {A}.
\newblock {Elsevier}.

\bibitem[\protect\citename{Kakas \& Mancarella, }1990]{kaka-manc-90}
Kakas, A.C., \& Mancarella, P. (1990).
\newblock {Database Updates Through Abduction}.
\newblock {\em Pages  650--661 of:} McLeod, D., Sacks-Davis, R., \& Schek, H.J. (eds), {\em {Proc.\ Sixteenth
  Conf.\ on Very Large Databases $($VLDB'90\/$)$}}.
\newblock {Morgan Kaufmann}.

\bibitem[\protect\citename{Katsuno \& Mendelzon, }1991]{kats-mend-91}
Katsuno, H., \& Mendelzon, A.O. (1991).
\newblock {On the Difference between Updating a Knowledge Database and Revising
  It}.
\newblock {\em Pages  387--394 of:} Allen, J.F., Fikes, R., \& Sandewall, E.
  (eds), {\em {Proc.\ Second Int.\ Conf.\ on Principles
  of Knowledge Representation and Reasoning $($KR'91\/$)$}}.
\newblock {Morgan Kaufmann}.

\bibitem[\protect\citename{Kowalski \& Toni, }1996]{kowa-toni-96}
Kowalski, R.A., \& Toni, F. (1996).
\newblock {Abstract Argumentation}.
\newblock {\em {Artificial Intelligence and Law}}, {\bf 4}(3-4), 275--296.

\bibitem[\protect\citename{Kraus {\em et~al.}\relax,
  }1990]{KrausLehmannMagidor90}
Kraus, S., Lehmann, D., \& Magidor, M. (1990).
\newblock Nonmonotonic {R}easoning, {P}referential {M}odels and {C}umulative
  {L}ogics.
\newblock {\em {Artificial Intelligence}}, {\bf 44}(1-2), 167--207.

\bibitem[\protect\citename{Laenens {\em et~al.}\relax, }1990]{LaeSacVer90}
Laenens, E., Sacc{\`a}, D., \& Vermeir, D. (1990).
\newblock Extending {L}ogic {P}rogramming.
\newblock {\em Pages  184--193 of:} Garcia-Molina, H., \& Jagadish, H.V. (eds) {\em Proc.\ Nineteenth {ACM} {SIGMOD} {C}onf.\ on the
  {M}anagement of {D}ata}.

\bibitem[\protect\citename{Lehmann, }1995]{lehm-95}
Lehmann, D. (1995).
\newblock {Belief Revision, Revised}.
\newblock {\em Pages  1534--1540 of:} {\em {Proc.\ Fourteenth Int.\
  Joint Conf.\ on Artificial Intelligence $($IJCAI'95\/$)$}}.
\newblock {Morgan Kaufmann}.

\bibitem[\protect\citename{Lehmann \& Magidor, }1992]{lehm-magi-92}
Lehmann, D., \& Magidor, M. (1992).
\newblock {What Does a Conditional Knowledge Base Entail?}
\newblock {\em {Artificial Intelligence}}, {\bf 55}, 1--60.

\bibitem[\protect\citename{Leite, }1997]{leite-97}
Leite, J.A. (1997).
\newblock {Logic Program Updates}.
\newblock {M.Sc.\ Thesis, Universidade Nova de Lisboa}.

\bibitem[\protect\citename{Leite {\em et~al.}\relax, }2000]{leit-etal-00}
Leite, J.A., Alferes, J.J., \& Pereira, L.M. (2000).
\newblock {Multi-Dimensional Dynamic Logic Programming}.
\newblock {\em Pages 17--26 of:} Satoh, K., \& Sadri, F. (ed) {\em {Proc.\ CL-2000 Workshop on Computational Logic in
  Multi-Agent Systems $($CLIMA 2000\/$)$}}.

\bibitem[\protect\citename{Leite \& Pereira, }1997]{leit-pere-98}
Leite, J.A., \& Pereira, L.M. (1997).
\newblock {Generalizing Updates: From Models to Programs}.
\newblock {\em Pages  224--246 of:} Dix, J., Pereira, L.M., \& Przymusinski, T.C.
  (eds), {\em {Logic Programming and Knowledge Representation, Selected Papers
  of the Third Int.\ Workshop $($LPKR'97\/$)$}}.
\newblock {LNAI}, vol. 1471.
\newblock {Springer}.

\bibitem[\protect\citename{Lifschitz \& Turner, }1994]{lifs-turn-94}
Lifschitz, V., \& Turner, H. (1994).
\newblock {Splitting a Logic Program}.
\newblock {\em Pages  23--37 of:} Van Hentenryck, P. (ed) {\em Proc.\ Eleventh Int.\ Conf.\ on Logic Programming $($ICLP'94\/$)$}.
\newblock {MIT Press}.

\bibitem[\protect\citename{Makinson, }1993]{maki-93}
Makinson, D. (1993).
\newblock {General Patterns in Nonmonotonic Reasoning}.
\newblock {\em Pages  35--110 of:} Gabbay, D.M., Hogger, C.J., \& Robinson,
  J.A. (eds), {\em {Handbook of Logic in Artificial Intelligence and Logic
  Programming, Volume III}}.
\newblock {Oxford Science Publications}.

\bibitem[\protect\citename{Makinson \& G{\"a}rdenfors,
  }1991]{makinson-gardenfors:1990a}
Makinson, D., \& G{\"a}rdenfors, P. (1991).
\newblock Relations {B}etween the {L}ogic of {T}heory {C}hange and
  {N}onmonotonic {L}ogic.
\newblock {\em Pages  185--205 of:} Fuhrmann, A., \& Morreau, M. (eds), {\em
  {T}he {L}ogic of {T}heory {C}hange}.
\newblock {Springer}.

\bibitem[\protect\citename{Marek \& Truszczy{\'n}ski, }1994]{mare-trus-94}
Marek, {V.W.}, \& Truszczy{\'n}ski, M. (1994).
\newblock {Revision Specifications by Means of Programs}.
\newblock {\em Pages  122--136 of:} MacNish, C., Pearce, D., \& Pereira, L.M.
  (eds), {\em {Proc.\ Europ.\ Workshop on Logics in Artificial
  Intelligence $($JELIA'94\/$)$}}.
\newblock {LNAI}, vol. 838.
\newblock {Springer}.

\bibitem[\protect\citename{Marek \& Truszczy{\'n}ski, }1991]{mare-trus-91}
Marek, {V.W.}, \& Truszczy{\'n}ski, M. (1991).
\newblock {Autoepistemic Logic}.
\newblock {\em {Journal of the ACM}}, {\bf 38}(3), 588--619.

\bibitem[\protect\citename{Nebel, }1991]{nebe-91}
Nebel, B. (1991).
\newblock {Belief Revision and Default Reasoning: Syntax-Based Approaches}.
\newblock {\em Pages  417--428 of:} Allen, J.F., Fikes, R., \& Sandewall, E. (eds) {\em {Proc.\ Second
  Int.\ Conf.\ on Principles of Knowledge Representation and
  Reasoning $($KR'91\/$)$}}.
\newblock {Morgan Kaufmann}.

\bibitem[\protect\citename{Niemel{\"a} \& Simons, }1996]{niem-simo-96b}
Niemel{\"a}, I., \& Simons, P. (1996).
\newblock {Efficient Implementation of the Well-founded and Stable Model
  Semantics}.
\newblock {\em Pages  289--303 of:} Maher, M.J. (ed) {\em {Proc.\ Thirteenth Joint
  Int.\ Conf.\ and Symposium on Logic Programming}}.
\newblock {MIT Press}.

\bibitem[\protect\citename{Papadimitriou, }1994]{papa-94}
Papadimitriou, C.H. (1994).
\newblock {\em {Computational Complexity}}.
\newblock {Addison Wesley}.

\bibitem[\protect\citename{Rao {\em et~al.}\relax, }1997]{rao-etal-97}
Rao, P., Sagonas, K.F., Swift, T., Warren, D.S., \&
  Freire, J. (1997).
\newblock {XSB: A System for Efficiently Computing Well-Founded Semantics}.
\newblock {\em Pages  431--441 of:} Dix, J., Furbach, U., \& Nerode, A. (eds) {\em Proc.\ Fourth Int.\
  Conf.\ on Logic Programming and Non-monotonic Reasoning (LPNMR'97)}.
\newblock {LNAI}, vol.~1265.
\newblock {Springer}.

\bibitem[\protect\citename{Sadri \& Toni, }2000]{sadr-toni-99}
Sadri, F., \& Toni, F. (2000).
\newblock {Computational Logic and Multi-Agent Systems: a Roadmap}.
\newblock {\em {C}omputational {L}ogic, {S}pecial {I}ssue on the {F}uture
  {T}echnological {R}oadmap of {C}ompulog-{N}et}.
\newblock Pages 1--31.
\newblock URL: http://www.compulog.org/net/Forum/Supportdocs.html.

\bibitem[\protect\citename{Schaub, }1991]{schaub:1991a}
Schaub, T. (1991).
\newblock On {C}ommitment and {C}umulativity in {D}efault {L}ogics.
\newblock {\em Pages  305--309 of:} Kruse, R., \& Siegel, P. (eds), {\em
  Symbolic and {Q}uantitative {A}pproaches for {U}ncertainty: {P}roc.\ of
  the {E}urop.\ {C}onf.\ {ECSQAU}}.
\newblock {Springer}.

\bibitem[\protect\citename{Schlipf, }1995]{schl-95a}
Schlipf, {J.S.} (1995).
\newblock {Complexity and Undecidability Results in Logic Programming}.
\newblock {\em {Annals of Mathematics and Artificial Intelligence}}, {\bf
  15}(3--4), 257--288.

\bibitem[\protect\citename{Subrahmanian {\em et~al.}\relax,
  }2000]{subr-etal-98}
Subrahmanian, {V.S.}, Bonatti, P., Dix, J., Eiter, T., Kraus, S., Ozcan,
  F., \& Ross, R. (2000).
\newblock {\em {Heterogeneous Agent Systems: Theory and Implementation}}.
\newblock {MIT Press}.

\bibitem[\protect\citename{{Van Gelder} {\em et~al.}\relax,
  }1991]{vang-etal-91}
{Van Gelder}, A., Ross, K.A., \& Schlipf, J.S. (1991).
\newblock {The Well-Founded Semantics for General Logic Programs}.
\newblock {\em {Journal of the ACM}}, {\bf 38}(3), 620--650.

\bibitem[\protect\citename{Winslett, }1988]{wins-88}
Winslett, M. (1988).
\newblock {Reasoning about Action Using a Possible Models Approach}.
\newblock {\em Pages  89--93 of:} {\em {Proc.\ Seventh Nat.\ Conf. on Artificial Intelligence $($AAAI'88\/$)$}}.
\newblock {AAAI Press/MIT Press}.

\bibitem[\protect\citename{Zhang \& Foo, }1997]{foo-zhan-97a}
Zhang, Y., \& Foo, N.Y. (1997).
\newblock {Towards Generalized Rule-based Updates}.
\newblock {\em Pages  82--88 of:} {\em {Proc.\ Fifteenth
  Int.\ Joint Conf.\ on Artificial Intelligence $($IJCAI'97\/$)$}},
   vol. 1.
\newblock {Morgan Kaufmann}.

\bibitem[\protect\citename{Zhang \& Foo, }1998]{foo-zhan-98}
Zhang, Y., \& Foo, N.Y. (1998).
\newblock {Updating Logic Programs}.
\newblock {\em Pages  403--407 of:} Prade, H. (ed), {\em {Proc.
  Thirteenth Europ.\ Conf.\ on Artificial Intelligence $($ECAI'98\/$)$}}.
\newblock {Wiley}. 

\end{thebibliography}

\end{document}